\documentclass[letterpaper]{article} 
\usepackage{aaai2026}  

\usepackage{times}  
\usepackage{helvet}  
\usepackage{courier}  
\usepackage[hyphens]{url}  
\usepackage{graphicx} 
\usepackage{natbib}  
\usepackage{caption} 
\usepackage{algorithm}

\usepackage{newfloat}
\usepackage{listings}
\DeclareCaptionStyle{ruled}{labelfont=normalfont,labelsep=colon,strut=off} 
\floatstyle{ruled}
\newfloat{listing}{tb}{lst}{}
\floatname{listing}{Listing}
\title{Selection of LLM Fine-Tuning Data based on Orthogonal Rules}

\title{Selection of LLM Fine-Tuning Data based on Orthogonal Rules}

\author{
    Xiaomin Li\textsuperscript{\rm 1}\thanks{Correspondence to: Xiaomin Li (xiaominli@g.harvard.edu).},
    Mingye Gao\textsuperscript{\rm 2},
    Zhiwei Zhang\textsuperscript{\rm 3},
    Chang Yue\textsuperscript{\rm 4},
    Hong Hu\textsuperscript{\rm 5}
}
\affiliations{
    \textsuperscript{\rm 1}Harvard University\\
    \textsuperscript{\rm 2}Massachusetts Institute of Technology\\
    \textsuperscript{\rm 3}Pennsylvania State University\\
    \textsuperscript{\rm 4}Princeton University\\
    \textsuperscript{\rm 5}Washington University in St.~Louis
}

\usepackage{bibentry}

\usepackage{amsmath,amsfonts,bm}

\def\vv{{\bm{v}}}

\def\vx{{\bm{x}}}

\def\mC{{\bm{C}}}

\def\mI{{\bm{I}}}

\def\mK{{\bm{K}}}

\def\mS{{\bm{S}}}

\def\mSigma{{\bm{\Sigma}}}

\newcommand{\calB}{\mathcal{B}}

\newcommand{\bydef}{\stackrel{\text{def}}{=}}
\newcommand{\bmS}{\bar{\mS}}

\newcommand{\R}{\mathbb{R}}
\newcommand{\E}{\mathbb{E}}
\renewcommand{\P}{\mathbb{P}}

\usepackage[utf8]{inputenc} 
\usepackage[T1]{fontenc}    
\usepackage{url}            
\usepackage{booktabs}       
\usepackage{amsfonts}       
\usepackage{nicefrac}       
\usepackage{microtype}      
\usepackage{xcolor}         

\usepackage{multirow}
\usepackage{enumitem}
\usepackage[most]{tcolorbox}
\usepackage{float}
\usepackage{caption}
\usepackage{subcaption}
\usepackage{longtable}
\usepackage{listings}
\usepackage{amssymb}
\usepackage{mathtools}
\usepackage{amsthm}
\usepackage{multicol}
\usepackage{titletoc}
\usepackage{algpseudocode}
\usepackage{bm}   
\usepackage{makecell}
\usepackage{tabularx}
\usepackage{adjustbox}

\newtheorem{theorem}{Theorem}

\newtheorem{lemma}{Lemma}

\newcommand{\mytiny}{\fontsize{9pt}{10pt}\selectfont}

\begin{document}

\maketitle

\begin{abstract}
High-quality training data is critical to the performance of large language models (LLMs). Recent work has explored using LLMs to rate and select data based on a small set of human-designed criteria (rules), but these approaches often rely heavily on heuristics, lack principled metrics for rule evaluation, and generalize poorly to new tasks. We propose a novel rule-based data selection framework that introduces a metric based on the orthogonality of rule score vectors to evaluate and select complementary rules. Our automated pipeline first uses LLMs to generate diverse rules covering multiple aspects of data quality, then rates samples according to these rules and applies the determinantal point process (DPP) to select the most independent rules. These rules are then used to score the full dataset, and high-scoring samples are selected for downstream tasks such as LLM fine-tuning. We evaluate our framework in two experiment setups: (1) alignment with ground-truth ratings and (2) performance of LLMs fine-tuned on the selected data. Experiments across IMDB, Medical, Math, and Code domains demonstrate that our DPP-based rule selection consistently improves both rating accuracy and downstream model performance over strong baselines.
\end{abstract}

\begin{links}
    \link{Code \& Data}{https://github.com/XiaominLi1998/Submission-OrthoRules/}
\end{links}

\section{Introduction}\label{sec:Introduction}
Large language models (LLMs) have been widely adopted across a diverse range of applications. Training these models—both during pre-training and fine-tuning—typically requires large and varied datasets. Prior work has shown that data quality plays a critical role in the effectiveness of LLM training \citep{brown2020language, chowdhery2023palm, du2022glam, dubey2024llama, wenzek2019ccnet}. For example, Meta’s LIMA paper \citep{zhou2024lima} demonstrated that just 1K carefully curated samples can outperform a much larger set of 50K original samples. Similar findings have emerged from other studies, where selecting high-quality data subsets improves both training efficiency and model performance \citep{cao2023instruction, hsieh2023distilling, xie2024data, sachdeva2024train, zhang2023opt}.

A recent trend involves using LLM-as-a-judge to rate data quality based on a set of human-designed metrics, referred to here as ``rules'' \citep{yuan2024self, wettig2024qurating, bai2022constitutional, mu2024rule}. For instance, \citet{wettig2024qurating} used LLMs to score pre-training data according to four predefined rules. RedPajama \citep{RedPajama2023} built a rule set with over 40 criteria for evaluating LLM training data. In specialized domains like safety, Constitutional AI \citep{bai2022constitutional} defined a set of ``constitutions'' for generating safe synthetic data; \citet{huang2024collective} later expanded this to 133 rules. OpenAI’s Rule-based Rewarding \citep{mu2024rule} similarly introduced 21 general safety rules as part of the reinforcement learning from human feedback (RLHF) pipeline. These rule-based approaches provide better interpretability by assigning granular, rule-specific scores rather than a single opaque quality label, and studies have shown this fine-grained strategy leads to more accurate assessments \citep{yuan2024self, wettig2024qurating, bai2022constitutional, mu2024rule}.

Despite this progress, several challenges remain. First, designing an effective rule set is difficult, as acknowledged in the prior work mentioned above. Most current rules rely heavily on human heuristics and are often too broad to yield useful signal. Second, there is a lack of principled metrics for evaluating rules, and little understanding of how rule quality and quantity affect downstream outcomes. Prior work \citep{bai2022constitutional, wettig2024qurating, RedPajama2023} typically randomly selects a subset of rules for scoring, a decision that can substantially impact the resulting data quality. Additionally, many rules are correlated, introducing redundancy and potential bias in the ratings \citep{wettig2024qurating}. This raises an important question: with a ``constitution'' (a pool of rules) in hand, exactly which ``laws'' (a subset of task-related rules) should be applied to a specific task? Random selection as in \citet{bai2022constitutional} may not be the optimal strategy. A third major drawback is the limited flexibility of these rules; they are often designed for specific settings, such as safety tasks, and lack general applicability across diverse settings.

In our work, we aim to address these challenges. We first leverage GPT-4 \citep{achiam2023gpt} to automatically generate candidate rules, prompting it with descriptions of both the target task and the source dataset. The generated rules are then manually reviewed to ensure clarity and validity. At this stage, some generated rules are found to be repetitive or redundant. Our strategy is to first generate a rule set that can comprehensively cover various data quality aspects, which may include many correlated and redundant rules. The second step is to filter out repetitive ones, and select a subset of rules that are relatively uncorrelated/independent. This is achieved by first using the rules to rate a batch of data, creating one score vector for each rule, and then assessing the independence of rule subsets through the overall orthogonality of their corresponding score vectors. We propose the formula in Section~\ref{sec:Methodology} to measure the orthogonality and use determinant point process (DPP) sampling \citep{macchi1975coincidence, borodin2000distributions} to identify a subset of independent rules. Once the rules are determined, the third step is to apply them to rate the full dataset and select the high-quality subset. Combining rule generation, rule-based rating, rule selection by DPP, and data selection, our method establishes a fully automated framework for rule-based data selection (illustrated in Figure~\ref{fig:pipeline}). To our knowledge, we are the first to introduce a mathematical evaluation metric for rules based on score vector orthogonality. Moreover, this pipeline can be applied to new tasks with minimal manual effort, addressing the shortcomings of existing methods.

We validate our approach across four domains: IMDB, Medical, Math, and Code, by fine-tuning LLMs on data selected by our framework.
We show that our rule-based method consistently improves data rating accuracy and leads to better model performance. Below is a list of main contributions of our work:
\begin{enumerate}[left=0pt,nosep]
    \item \textbf{Rule-free vs. Rule-based Rating.} We provide the first systematic comparison showing that fine-grained, rule-based evaluation outperforms rule-free methods in data quality assessment.

    \item \textbf{Rule Evaluation Metric:} We introduce a novel rule evaluation metric designed to promote low correlation and high diversity among rules, the first rule evaluation metric for rule-based LLM-as-a-judge to our knowledge. We also propose the method of using DPP on task-aware score vectors to select a subset of independent rules.

    \item \textbf{Automated Rule-based Selection Pipeline.} We confirm that LLMs are effective rule generators, eliminating the need for manual rule crafting. Our automated pipeline generates the rules, selects the rules, and then uses them to identify high-quality data.

    \item \textbf{Cross-Domain and Cross-Model.} We validate our method both by comparing against ground-truth ratings and by benchmarking LLMs trained on selected data across multiple domains (IMDB, Medical, Math, Code) and model families (Pythia-1B and Llama3-8B with LoRA), confirming the generality of our approach.
\end{enumerate}

\section{Related Work}\label{sec:RelatedWork}

\textbf{Rule-based rating.} Some studies adopt a more fine-grained approach to data quality by distilling it into a finite set of metrics, which we refer to as ``rules''. For instance, RedPajama \citep{RedPajama2023} provides over 40 quality rules as basic filtering criteria. More relevant to our research, several studies apply this rule-based idea to rate LLM data. \citet{yuan2024self} rates each sample from 1 to 5 based on how many of five predefined criteria it satisfies. \citet{wettig2024qurating} uses four general rules to rate and select pre-training data, while \citet{sun2024principle} proposes 16 handcrafted rules to assess response quality. Rule-based evaluation is also used in safety applications. Constitutional AI \citep{bai2022constitutional} applies a random subset of 16 safety critique rules (called ``constitutions'') to iteratively revise synthetic data. In \citet{mu2024rule}, rule-based scores from 21 safety rules are directly integrated into the RLHF reward, while \citet{wang2024helpsteer2} trains a composite reward model using rule-based ratings. As noted earlier, existing rule designs often rely heavily on human heuristics, lack principled evaluation metrics and exploration of rule sizes, and are not easily adaptable. Our framework addresses these limitations.

\textbf{LLM data selection.} There are different genres of data selection approaches for LLMs. Basic filtering, such as setting thresholds on word length, is used in many studies to eliminate low-quality data \citep{soldaini2024dolma, wenzek2019ccnet, raffel2020exploring, penedo2023refinedweb}. Fuzzy deduplication removes repetitive or similar samples \citep{allamanis2019adverse, lee2021deduplicating, gao2020pile, jiang2022fuzzydedup}. Another method is \textit{heuristic classification}, selecting data based on similarity to high-quality sources such as Wikipedia \citep{brown2020language, touvron2023llama, chowdhery2023palm, du2022glam, gao2020pile, wenzek2019ccnet}. More recently, querying LLMs to rate data directly has become a standard practice \citep{li2023quantity, chen2023alpagasus, bai2022constitutional, wettig2024qurating, yuan2024self, dubois2024alpacafarm}. Other methods include coreset and optimization-based subset selection \citep{xia2022moderate, xiao2024feature, borsos2020coresets}.

\section{Methodology}\label{sec:Methodology}

\subsection{Definitions and Notations}\label{subsec:DefinitionsNotations}
We introduce the definitions of the primary objects considered in our method:
\begin{itemize}
\item $R$: the total number of available rules.
\item $r$: the number of selected rules, using a specified rule selection method.
\item $\mathcal{D}$: the set of all data samples, with its size denoted by $N \bydef |\mathcal{D}|$.
\item $\mathcal{B} \subseteq \mathcal{D}$: a batch of data samples, randomly selected for evaluating the correlation of rules during the rule selection step, with its size denoted by $n \bydef |\mathcal{B}|$ (we use $n=10^4$ in experiments in Section~\ref{sec:EvaluationB}).
\item $\mS \in \R^{n \times R}$: the rating matrix $\mS$ where each entry $S_{i,j}$ represents the score of the $i$-th data sample according to the $j$-th rule and is constrained to the interval $[0,1]$.
\item $\bmS \in \R^{n \times r}$: a submatrix of $\mS$ consisting of the $r$ selected columns from $\mS$, corresponding to the $r$ selected rules. 
\end{itemize}

\textbf{Measure orthogonality:}
To guide rule selection, we propose a metric based on the orthogonality of score vectors. To achieve this, we introduce a mathematical framework to quantify the degree of orthogonality or correlation among a given set of score vectors. Given $\bmS \in \mathbb{R}^{n \times r}$, define its sample mean as $\mu_i \bydef \frac{1}{n} \sum_{k=1}^{n} S_{k,i}$,
and sample covariance matrix $\widehat{\mSigma}(\bmS) \in \mathbb{R}^{r \times r}$ by
\[
\widehat{\mSigma}_{i,j}(\bmS)
\bydef
\frac{1}{n} \sum_{k=1}^{n}
\left( S_{k,i} - \mu_i \right) \left( S_{k,j} - \mu_j \right).
\]
Then define the \emph{sample correlation matrix} $\widehat{\mC}(\bmS) \in \mathbb{R}^{r \times r}$ as
\[
\widehat{\mC}_{i,j}(\bmS) \bydef
\frac{
\widehat{\mSigma}_{i,j}(\bmS)
}{
\sqrt{
\widehat{\mSigma}_{i,i}(\bmS) \;
\widehat{\mSigma}_{j,j}(\bmS)
}
},
\quad
1 \leq i,j \leq r.
\]
To quantify the degree of correlation/dependence for a given rating submatrix $\bmS$, we introduce the quantity called \textit{rule correlation}:
\begin{equation}\label{eq:rule_correlation}
\rho(\bmS) \bydef \frac{1}{r} \left\| \widehat{\mC}(\bmS) - \mI_r \right\|_F
=
\frac{1}{r} \sqrt{ \sum_{i \neq j} \left( \widehat{\mC}_{i,j}(\bmS) \right)^2}.
\end{equation}
Here, $\mI_r$ is the identity matrix and $\|\cdot\|_F$ is the Frobenius norm. Intuitively, $\widehat{\mC}(\bmS) \approx \mI_r$ means the columns of $\bmS$ have low pairwise correlations, making $\rho(\bmS)$ small. This metric quantifies how much the columns of $\bmS$ deviate from orthogonality, by measuring the deviation of its correlation matrix from the identity matrix. The second equality in \eqref{eq:rule_correlation} provides another intuitive understanding: $\rho(\bmS)$ essentially aggregates the correlations of all pairwise correlations of rules $(i, j)$ for $i\neq j$. 

To validate our approach of using $\calB$, batch of random $n$ samples, to generate rating scores and evaluate rule correlations based on these score vectors, we present the following theorem, which characterizes the concentration of the sample rule correlation around the true rule correlation.

\begin{theorem}\label{thm:RuleCorrelation}
Let $\mC \in \mathbb{R}^{r \times r}$ be the \emph{true} correlation matrix among the $r$ rules, i.e.,
\begin{equation}
    \mC_{j,\ell} = \frac{\Sigma_{j,\ell}}{\sqrt{\Sigma_{j,j} \Sigma_{\ell,\ell}}},
\end{equation}
and assume each candidate rule has nontrivial variance ($\Sigma_{j,j} \geq \sigma_{\min}^2 > 0$ for some constant $\sigma_{\min}$). We draw $n$ i.i.d.\ samples $\{\vx^{(k)}\}_{1\leq k \leq n}$ where each $\vx^{(k)} \in [0,1]^r$ represents ratings for the $k$-th sample based on $r$ rules. From these we form the empirical correlation matrix $\widehat{\mC}$. Then there exists a universal constant $c > 0$ such that for any $\delta > 0$ and sufficiently large $n$,
\[
\P \left(
\left|
\rho(\bmS)
-
\frac{1}{r} \left\|
\mC - \mathbf{I}_r
\right\|_F
\right|
\leq
c \sqrt{\frac{\ln\left(\frac{r^2}{\delta}\right)}{n}}
\right)
> 1-\delta.
\]
In particular, if $\mC$ is close to $\mathbf{I}_r$ in Frobenius norm (i.e., if the rules are nearly uncorrelated ``in truth''), then the observed rule correlation $\rho(\bmS)$ also remains close to zero for sufficiently large $n$.
\end{theorem}

\begin{proof}
    See Appendix~\ref{sec:Appendix-ProofTheorm}.
\end{proof}

\subsection{Determinantal point process (DPP)}
The optimal solution to this mathematical problem of selecting the most orthogonal subset of a set of vectors is NP-hard \citep{civril2007finding, kulesza2012determinantal} but we use DPP sampling to provide a relatively good solution. The determinant point process (DPP) is a probabilistic model that describes the likelihood of selecting diverse subsets from a larger set \citep{macchi1975coincidence, borodin2000distributions}. Mathematically, a DPP is defined by a kernel matrix that describes the similarities between elements in a set. The probability of selecting a particular subset is proportional to the determinant of the corresponding submatrix of this kernel matrix. Intuitively, subsets with highly similar items (leading to higher correlation in the submatrix) have smaller determinants and are thus less likely to be chosen.

\textbf{DPP Definitions.} Given a discrete ground set $\mathcal{Y}$, without loss of generality we let $\mathcal{Y} = \{1, 2, \dots, R\}$, a (discrete) DPP defines a probability measure over $2^{\mathcal{Y}}$, the power set of $\mathcal{Y}$. Let $Y$ be a randomly chosen subset. Then for any subset $A \subseteq \mathcal{Y}$, the probability of $A$ being chosen by a DPP is given by:
\[
\mathbb{P}(A \subseteq Y) = \det(\mK_A)
\]
where $\mK \in \R^{R\times R}$ is a real positive-semidefinite matrix called the \textit{kernel matrix} and $\mK_A \bydef [\mK]_{i,j \in A}$ is the submatrix of $\mK$ indexed by elements in $A$.

\textbf{Kernel Matrix.} Each entry $K_{ij}$ in the kernel matrix $\mK$ describes the similarity between elements $i$ and $j$ in $\mathcal{Y}$. For our purpose of selecting orthogonal rules, we will define $\mK$ as the Gram matrix of the score vectors: $\mK \bydef\mS^\top \mS$.

\textbf{DPP Sampling.} To sample a diverse subset using DPP, there are several existing algorithms \citep{hough2006determinantal, kulesza2012determinantal, tremblay2018optimized} and the Python library \texttt{DPPy} \citep{GPBV19} implements some of them. The computation of the DPP sampling primarily hinges on the overhead of computing the inner product kernel matrix $\mK$ and its eigendecomposition. In our case, $\mK \in \R^{R \times R}$ and hence it requires $O(R^3)$ time, where $R$ is the number of all rules. Nonetheless, we set $R=50$ in our experiments, therefore our DPP rule selection algorithm is extremely fast (typically within 0.1 seconds). Further details about DPP sampling algorithms and their time complexities can be found in Appendix~\ref{sec:Appendix-DPPSampling}.

\subsection{DPP rule-based rating algorithm}\label{subsec:Algo}
\begin{figure*}[t]
    \centering
    \includegraphics[width=1.0\textwidth]{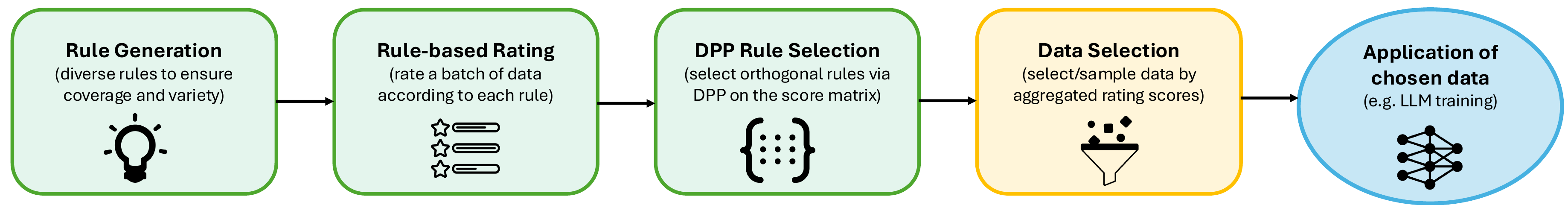}
    \caption{Pipeline for rule-based data rating and selection in five steps (detailed in Section~\ref{subsec:Algo})
    }
    \label{fig:pipeline}
\end{figure*}

Our rule-based data selection framework proceeds in five steps (Figure~\ref{fig:pipeline}):

\textbf{Step 1. Rule generation.}
We prompt GPT-4 with a task description and source data context to generate $R$ candidate rules. These rules are then manually refined for clarity and relevance.

\textbf{Step 2. Rule-based rating}:
We employ Llama3-8B-Instruct \citep{llama3modelcard}, to rate the batch data $\mathcal{B}$ according to $R$ rules, resulting in the score matrix $\mS \in \R^{n \times R}$.

\textbf{Step 3. Rule selection using DPP}:
From $\mS$, we aim to select $r$ relatively independent columns using a DPP, forming the submatrix $\bmS \in \R^{n \times r}$. We define the kernel matrix of DPP as follows:
\begin{equation}\label{eq:kernel}
    \mK \bydef\mS^\top \mS \in \R^{R\times R},
\end{equation}
where each entry $K_{i,j} = \langle S_i, S_j \rangle$ (each $S_i$ is the $i$-th column of $\mS$), representing the similarity between rule $i$ and rule $j$. We then employ the DPP sampling algorithm to select $r$ indices from $\{1, 2, \dots, R\}$, corresponding to the $r$ chosen rules.

Note that the cost of generating $R$ rules is negligible, requiring just a single GPT-4 query, and the cost of obtaining the rating matrix $\mS$ can be managed by adjusting the batch size $n$.
The motivation to select a fixed small number of $r$ rules is driven by the computational costs associated with using LLMs for data rating and the need to maintain a consistent dimensionality for explaining data quality. These practical considerations lead us to treat $r$ as a hyperparameter. Discussions on the optimal choices of $r$ are explored in Section~\ref{sec:EvaluationA}  and Appendix~\ref{subsec:Appendix-EvalB-20rules}.

Another important remark is that, even with the same set of rules, they could have different correlations conditioned on a specific task or dataset. Therefore during DPP selection, instead of employing fixed representations such as semantic encodings---which result in static rule representations and selections across all tasks---we use \textit{task-aware} score vectors to adaptively represent the rules. These vectors allow the entire pipeline to be customized for a particular downstream task.

\textbf{Step 4. Stochastic data selection}: 
We extend the rating process to cover all data samples using the selected $r$ rules, expanding the rating matrix $\bmS$ from $n \times r$ to $N \times r$. We then aggregate these fine-grained ratings by averaging across the $r$ columns of $\bmS$, resulting in a score vector $\vv = [v_1, v_2, \dots, v_N]^\top$ that assigns a quality score to each of the $N$ samples.

Given the $N$ scores and a fixed budget of selecting $k$ samples for training ($k=20000$ in Section~\ref{sec:EvaluationB}), rather than choose the traditional top-$k$ approach, (selecting the highest scored samples), we adopt a stochastic sampling strategy, where we sample $k$ data points according to the distribution:
\begin{equation}\label{eq:top-k-sampling}
    p(\vx_i) = \frac{e^{v_i /\tau }}{\sum_{j=1}^N e^{v_i /\tau }}
\end{equation}
for each data point $\vx_i \in \mathcal{D}$, and $\tau$ is the ``temperature'' balancing between top-$k$ and uniform sampling (we use $\tau=1$ by default). This stochastic data selection mechanism introduces greater diversity into the sampling process and is used in several other works \citep{wettig2024qurating, sachdeva2024train}. 

\textbf{Step 5.} Apply the selected data on given downstream tasks, such as for LLM domain fine-tuning.

\section{Preliminary Experiments: Evaluating Against Ground Truth Ratings}\label{sec:EvaluationA}

We evaluate our method through two complementary approaches:
(A) comparing rule-based ratings \textbf{against ground truth human ratings}, where smaller deviations from human scores indicate better performance, and
(B) training LLMs (Pythia-1B and Llama3-8B) on the selected data and \textit{evaluating their downstream performance on domain-specific benchmarks}.
This section focuses on the first evaluation setup (A), while the full-scale LLM training experiments (B) are presented in Section~\ref{sec:EvaluationB}. Because Evaluation A avoids expensive LLM training, it enables a broader exploration of factors such as the effect of rule set size and the influence of model scale (Llama3-8B vs. Llama3-70B). These experiments offer deeper insights into the robustness and effectiveness of our approach.

\subsection{Experiments Setup}
\textbf{Datasets and ground-truth ratings:}
We use four datasets spanning different domains: StanfordNLP-IMDB \citep{maas-EtAl:2011:ACL-HLT2011} (IMDB), MedQA \citep{jin2021disease} (Medical), GSM8K \citep{cobbe2021training} (Math), and MBPP \citep{austin2021program} (Code). For each dataset, we randomly sample 300 examples and ask five human annotators to assign quality scores, with the average scores treated as ground truth (GT). In this section, we present results on IMDB as a representative example; results on the remaining domains are provided in Appendix~\ref{subsec:Appendix-EvalA-AllDomains}.

\textbf{Rule-based rating:}
We apply our rule-based approach to rate the same data. For each of the $R = 50$ rules, we obtain ratings from two LLMs: Llama3-8B-Instruct and Llama3-70B-Instruct, to evaluate the effect of rater capability. Each rule $i$ produces a score vector $S_i \in \R^n$ over $n = 300$ samples, forming a full rating matrix $\mS \in \R^{n \times R}$. We then select a subset of $r$ rules, forming a submatrix $\bmS \in \R^{n \times r}$. To evaluate how well this rule subset aligns with ground-truth scores $S_{\text{GT}} \in \R^n$, we compute the mean squared error (MSE):
\begin{equation}\label{eq:mse}
   \epsilon(\bmS) \bydef \frac{1}{n} \left\|\frac{1}{r}\sum_{j=1}^r \bar{S}_j - S_{GT} \right\|^2_2 
\end{equation}
where $\bar{S}_j$ is the $j$-th column of $\bmS$. As baselines, we include:
(1) the four human-designed rules in QuRating \citep{wettig2024qurating}, and
(2) a rule-free method where Llama3 directly produces an overall quality score (``NoRule'').

Our experiments in this section aim to explore the following research questions:
\begin{itemize}
    \item \textbf{(Q1)} Does greater rule diversity lead to more accurate ratings? 
    \item \textbf{(Q2)} Does rule-based selection generally outperform rule-free methods? 
    \item \textbf{(Q3)} Is the rating quality based on our DPP-selected rules better than that based on human-created rules and NoRule setting?

    \item \textbf{(Q4)} Does DPP select better rules than randomly chosen ones? 
    \item \textbf{(Q5)} Can the method possibly generalize well to pre-train datasets?
\end{itemize}

\subsection{Results}\label{subsec:EvalA-Results}
\begin{figure}[ht]
\centering
\begin{subfigure}[b]{0.48\linewidth} 
    \centering
    \includegraphics[width=\linewidth]{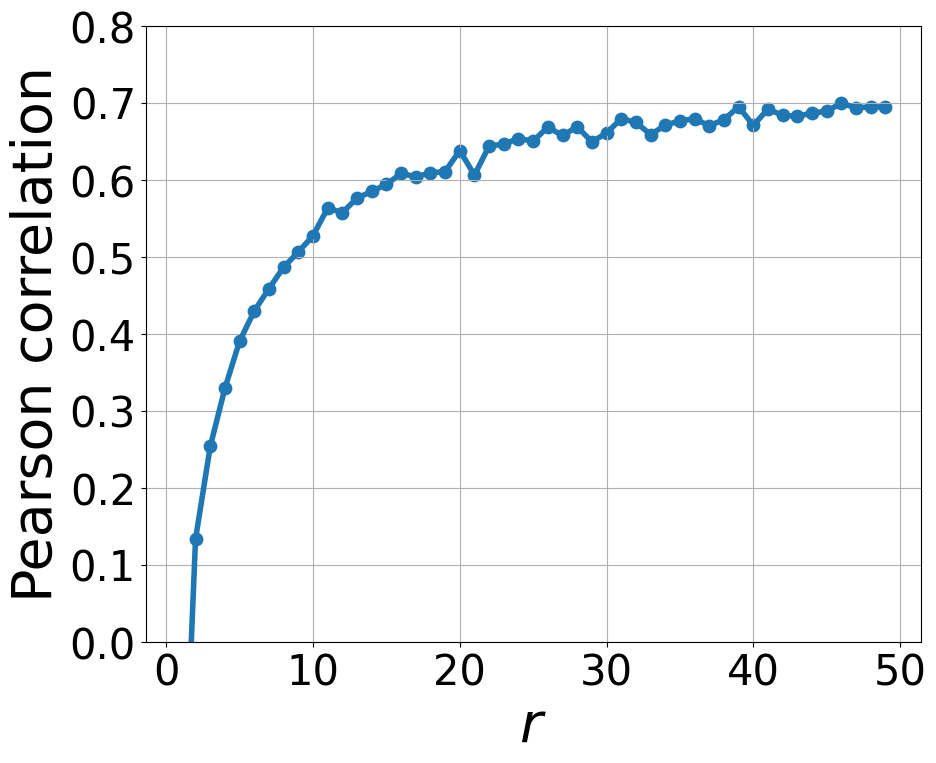}
    \caption{}
    \label{fig:EvalA_IMDB_Single_pearson}
\end{subfigure}
\begin{subfigure}[b]{0.48\linewidth} 
    \centering
    \includegraphics[width=\linewidth]{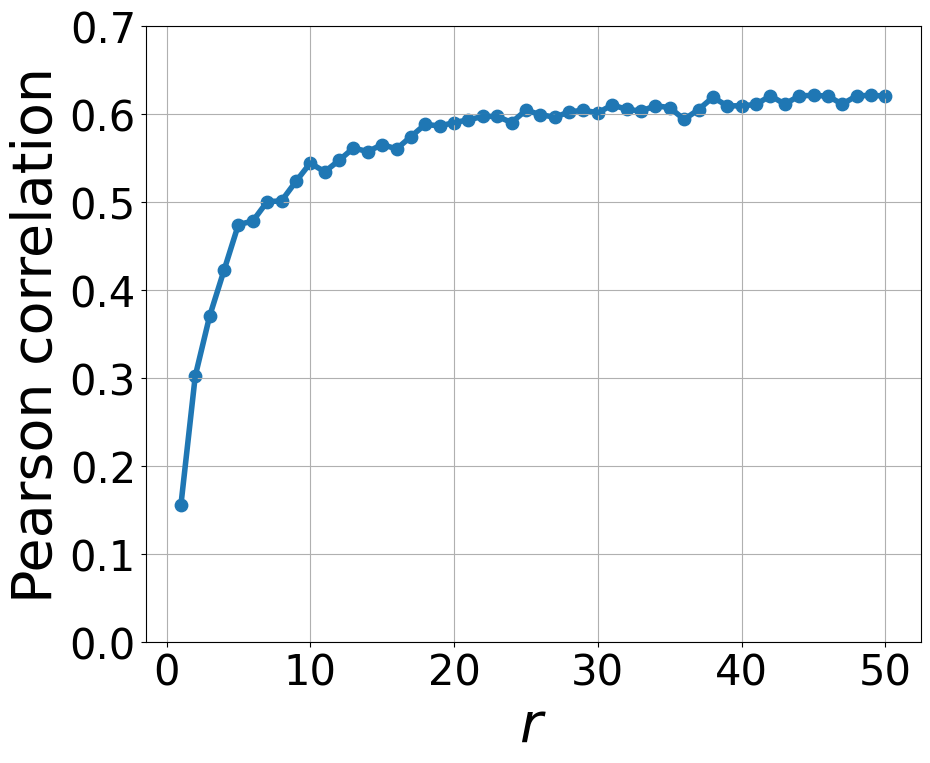}
    \caption{}
    \label{fig:EvalA_IMDB_Single70B_pearson}
\end{subfigure}

\vspace{0.5em} 

\begin{subfigure}[b]{0.48\linewidth}
    \centering
    \includegraphics[width=\linewidth]{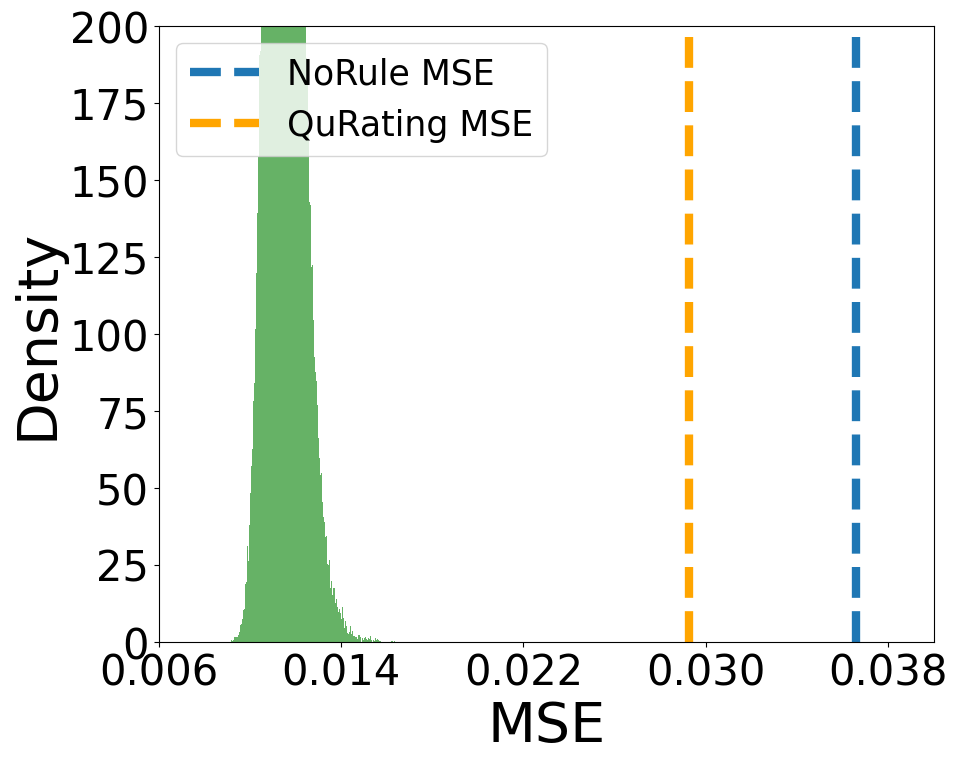}
    \caption{}
    \label{fig:EvalA_IMDB_Single_histogram}
\end{subfigure}
\begin{subfigure}[b]{0.48\linewidth}
    \centering
    \includegraphics[width=\linewidth]{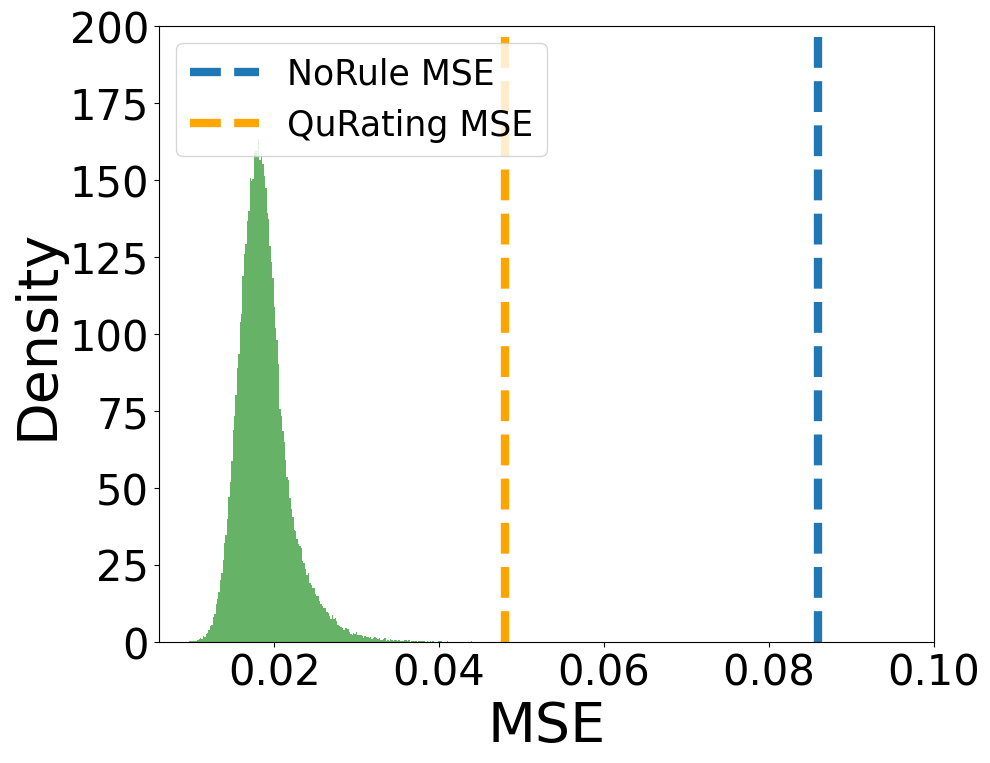}
    \caption{}
    \label{fig:EvalA_IMDB_Single70B_histogram}
\end{subfigure}
\caption{(a) and (b): Pearson correlation of the rule correlation $\rho(\bmS)$ and the MSE $\epsilon(\bmS)$, using Llama3 8B and 70B raters respectively. (c) and (d): Distribution of MSE from $10^6$ possible rule subsets with size $r$, using Llama3 8B and 70B raters respectively, where two vertical lines represent the MSE values of QuRating and NoRule.}
\end{figure}

\textbf{Correlation of $\rho(\bmS)$ and the MSE $\epsilon(\bmS)$ (answer to \textbf{Q1}).} For each rule subset size $r \in \{1,2,\dots, 50\}$, we sample $\min\{10000, \binom{50}{r}\}$ sets of indices of size $r$, which are used to choose rules and form $\bmS$. We then calculate its rule correlation $\rho(\bmS)$ and MSE $\epsilon(\bmS)$, as well as the Pearson correlation between them.
The Pearson correlation is surprisingly strong across both rating model (Llama3 8B and 70B) (Figures~\ref{fig:EvalA_IMDB_Single_pearson} and \ref{fig:EvalA_IMDB_Single70B_pearson}). This confirms that higher rule diversity is positively correlated with the accuracy of rating results. In other words, the highly correlated or redundant rules can potentially lead to non-robust ratings.

\textbf{Rule-based v.s. Rule-free (answer to \textbf{Q2}):} We sample $10^6$ random rule subsets of size $r$ and compare their MSEs against that of the NoRule baseline. Figures~\ref{fig:EvalA_IMDB_Single_histogram} and \ref{fig:EvalA_IMDB_Single70B_histogram} show that rule-based methods consistently achieve lower MSE. Interestingly, even randomly chosen domain-specific rules outperform the fixed QuRating rules, illustrating the limitations of static, general-purpose rule sets.

\begin{figure*}[ht]
\centering
\begin{subfigure}{0.26\textwidth}
  \centering
  \includegraphics[width=1.0\linewidth]{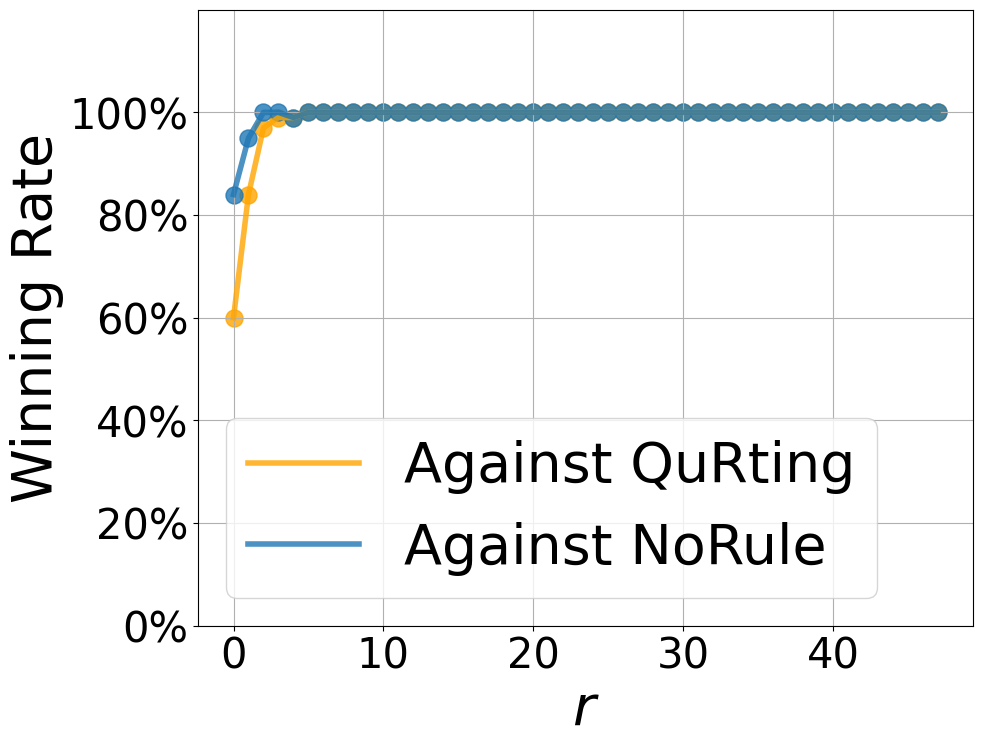}
  \caption{}
  \label{fig:EvalA_IMDB_Single_winning_rate}
\end{subfigure}
\begin{subfigure}{0.26\textwidth}
  \centering
  \includegraphics[width=1.0\linewidth]{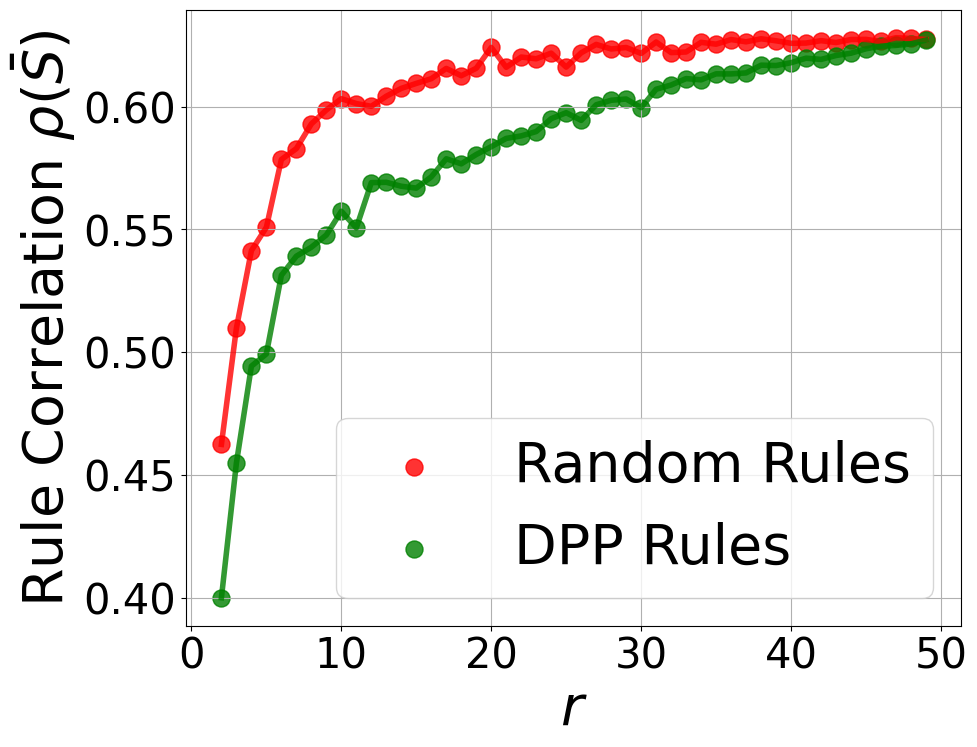}
  \caption{}
  \label{fig:EvalA_IMDB_Single_dpp_vs_random_RC}
\end{subfigure}
\begin{subfigure}{0.26\textwidth}
  \centering
  \includegraphics[width=1.0\linewidth]{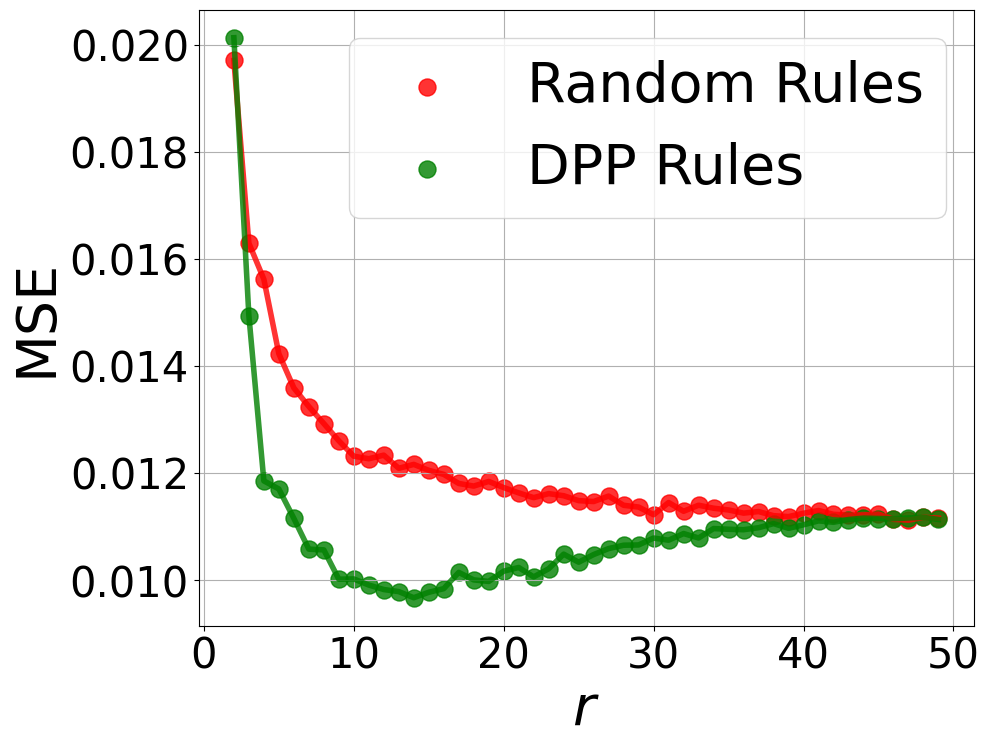}
  \caption{}
  \label{fig:EvalA_IMDB_Single_dpp_vs_random_MSE}
\end{subfigure}
\begin{subfigure}{0.26\textwidth}
  \centering
  \includegraphics[width=1.0\linewidth]{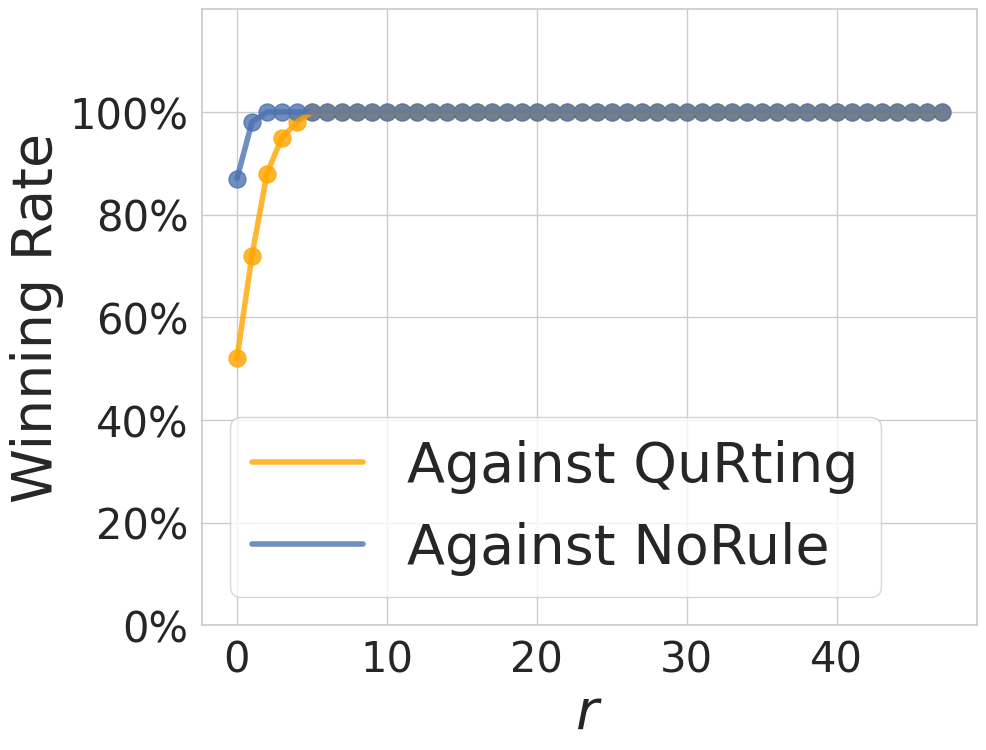}
  \caption{}
  \label{fig:EvalA_IMDB_Single70B_winning_rate}
\end{subfigure}
\begin{subfigure}{0.26\textwidth}
  \centering
  \includegraphics[width=1.0\linewidth]{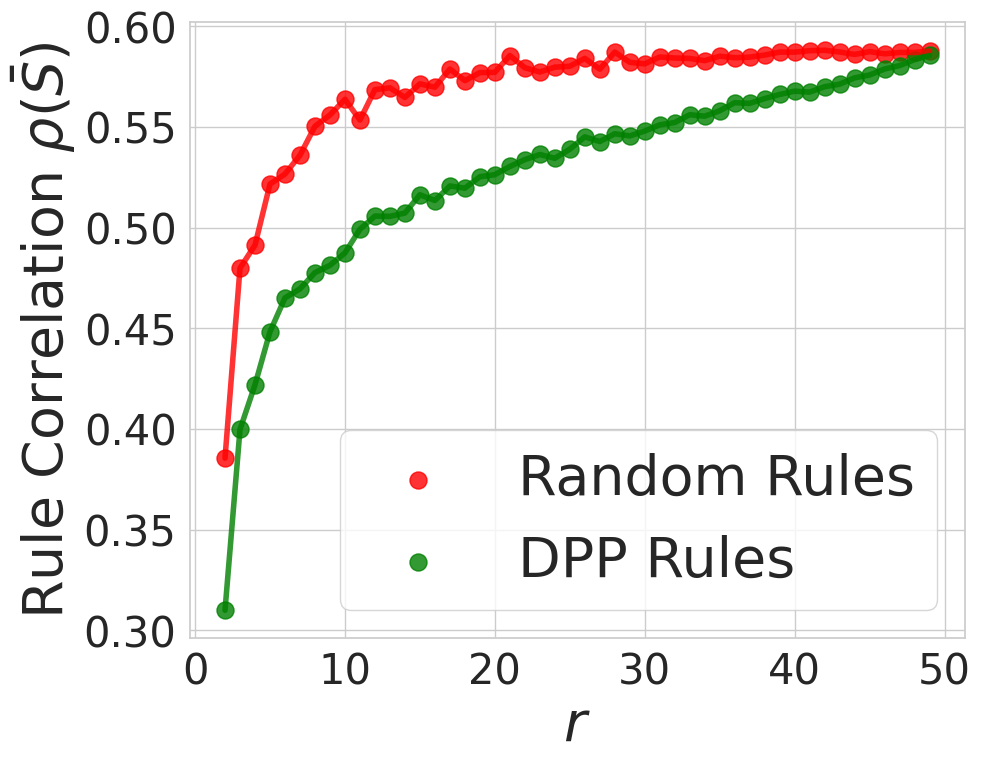}
  \caption{}
  \label{fig:EvalA_IMDB_Single70B_dpp_vs_random_RC}
\end{subfigure}
\begin{subfigure}{0.26\textwidth}
  \centering
  \includegraphics[width=1.0\linewidth]{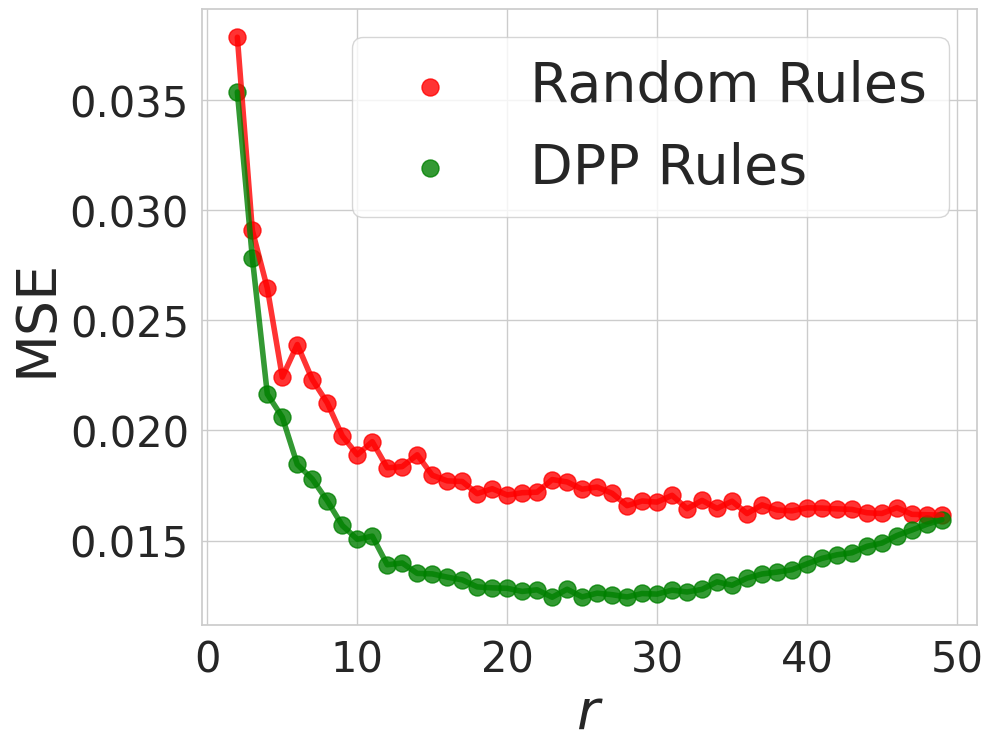}
  \caption{}
  \label{fig:EvalA_IMDB_Single70B_dpp_vs_random_MSE}
\end{subfigure}
\caption{(a) Winning rate of DPP-selected rules compared to QuRating’s four rules and the NoRule setting respectively, based on MSE across 100 DPP trials. (b) Comparison of rule correlation $\rho$ between DPP-selected and randomly selected rules, averaged across 100 trials. (c) Comparison of MSE between DPP-selected and randomly selected rules, averaged across 100 trials. Plots (a), (b), and (c) display results using Llama3-8B rater, while (d), (e), and (f) for the Llama3-70B rater.}
\label{fig:DPP_Compare_WinRates_RC_MSE}
\end{figure*}

\textbf{DPP v.s. QuRating v.s. NoRule (answer to \textbf{Q3}).} Using DPP, we select $r$ rules and repeat this for 100 trials. We report both the average MSE and the winning rate compared to QuRating and NoRule (Figures~\ref{fig:EvalA_IMDB_Single_winning_rate}, \ref{fig:EvalA_IMDB_Single70B_winning_rate}). Results show that once $r$ exceeds a modest threshold, DPP-selected rules almost always outperform both baselines.

\textbf{DPP rules v.s. Randomly selected rules (answer to \textbf{Q4}).} 
We compare DPP-selected rule subsets to randomly sampled ones of the same size, evaluating both the rule correlation $\rho(\bmS)$ and MSE $\epsilon(\bmS)$ for their corresponding score submatrices $\bmS$. As shown in Figures~\ref{fig:EvalA_IMDB_Single_dpp_vs_random_RC} and \ref{fig:EvalA_IMDB_Single_dpp_vs_random_MSE}, DPP consistently selects rules with lower correlation and MSE, regardless of the value $r$. Notably, MSE increases when $r$ is too small or too large, which matches our intuition: when $r$ is too small, there are too few rules to cover all important data quality aspects for rating, and when $r$ is too large, rule redundancy can negatively affect the rating outcomes. Therefore, we use $r=10$ in subsequent experiments.


\textbf{Apply to pre-training datasets (answer to \textbf{Q5})}.
So far, our experiments have focused on domain-specific datasets. We also applied the rule-based method to the CommonCrawl dataset \citep{commoncrawl}, a general-purpose web corpus commonly used for LLM pre-training. We find that the plots of rule correlation show similar patterns, and the Pearson correlation between $\rho$ and $\epsilon$ is consistently positive. However, the trends and Pearson values are less pronounced than in domain-specific settings. This may be because domain-specific datasets support more concrete and well-defined quality rules, whereas quality assessment in general pretraining data tends to be more ambiguous and less structured. Detailed results and analysis are provided in Appendix~\ref{subsec:Appendix-EvalA-CommonCrawl}.

\section{Experiments: Data Selection for LLM Fine-tuning}\label{sec:EvaluationB}

We now evaluate our full framework in a realistic setting by following the pipeline in Section~\ref{subsec:Algo} to fine-tune LLMs (Pythia-1B and Llama3-8B) on selected data, then measure their downstream performance. This setup reflects practical applications of LLM data selection in domain-specific fine-tuning.


\subsection{Experiments Setup}
\textbf{Evaluation Benchmarks.} 
We assess performance across four domains using standard benchmarks: \textbf{IMDB}: Sentiment classification using the IMDB dataset \citep{maas-EtAl:2011:ACL-HLT2011}, \textbf{Code}: Code generation evaluated on HumanEval \citep{chen2021evaluating}, MBPP \citep{austin2021program}, and Multiple-py/cpp \citep{cassano2022multipl}, \textbf{Math/Medical}: Relevant subsets of MMLU \citep{hendrycks2020measuring}.
We set the inference temperature to 0, unless the the benchmark protocol itself requires top-$k$ sampling. Full benchmark details are provided in Appendix~\ref{appendix:benchmark}.

\textbf{Data Source.}
We use SlimPajama \citep{CerebrasSlimPajama2023}, a large, deduplicated multi-source corpus for LLM training. From it, we randomly sample 1M examples (around 1B tokens) as our raw dataset $\mathcal{D}$, from which we select a high-quality subset of $k = 20{,}000$ samples for model training.

\textbf{Models.}
We intentionally choose Pythia-1B \citep{biderman2023pythia} as the base model for the IMDB and Medical tasks, as it was pre-trained on the Pile dataset \citep{gao2020pile}, offering a more controlled comparison than using models that may have been trained on SlimPajama. To validate the generalization ability of our framework across different LLMs, we train Llama3-8B \citep{llama3modelcard} with LoRA \citep{hu2021lora} for the Math and Code domains.

\textbf{Baselines.} We compare our method against the following baselines, including both rule-free and rule-based data selection methods:
\begin{itemize}[left=10pt,topsep=2pt,itemsep=1pt]
    \item \textbf{Rule-Free Methods}
    \begin{itemize}[left=10pt,topsep=1pt,itemsep=1pt]
        \item \textit{Uniform Sampling}: Randomly select 20K samples from the dataset.
        \item \textit{No Rule}: Use Llama3-8B to assign a quality score to each sample without any rule prompts, then apply the same sampling procedure as in Section~\ref{subsec:Algo}.
        \item \textit{DSIR} \citep{xie2024data}: Importance resampling based on similarity to a high-quality reference dataset (we use test sets from evaluation benchmarks).
        \item \textit{LESS} \citep{xia2024less}: Estimate data influence on downstream performance and select samples with the highest impact.
    \end{itemize}

    \item \textbf{Rule-Based Methods}
    \begin{itemize}[left=10pt,topsep=1pt,itemsep=1pt]
        \item \textit{QuRating} \citep{wettig2024qurating}: Rate and select data using a fixed set of four manually designed rules.
        \item \textit{GPT-Uncorrelated}: Prompt GPT-4 to directly generate 10 “uncorrelated” rules, then use them to rate and select data.
    \end{itemize}

    \item \textbf{All Data}
    \begin{itemize}[left=10pt,topsep=1pt,itemsep=1pt]
        \item \textit{AllData}: Use the entire 1M SlimPajama samples without applying any selection.
    \end{itemize}
\end{itemize}


For our automated rule-based selection algorithm, we set $R=50$ and $r=10$. The choice of $r$ as a hyperparameter is based on experimental observations from Section~\ref{sec:EvaluationA}. As inferences of LLM are a lot less computing-consuming than model training, we set $n=10^4$ in all our experiments (the size of a randomly selected batch for generating score vectors and selecting orthogonal rules).

\subsection{Fine-tuning Results}\label{subsec:EvalB_domain_finetune}

\begin{table*}[ht]
    \centering
    \mytiny
    \renewcommand{\arraystretch}{1.2} 
    \setlength{\tabcolsep}{3.2pt}
\begin{tabular}{ c | c | c  c  c  c | c  c  c  c | c  c  c  c  c } \hline 
   \multirow{2}{*}{\textbf{Method}} & \textbf{IMDB} & \multicolumn{4}{c|}{\textbf{Medical}} & \multicolumn{4}{c|}{\textbf{Math}} & \multicolumn{5}{c}{\textbf{Code}} \\
    \cline{2-15} 
    & \makecell{SA} & \makecell{CM} & \makecell{PM} & \makecell{MG} & \makecell{Med\\Avg}
& \makecell{ES} & \makecell{HS} & \makecell{CMath} & \makecell{Math\\Avg}
& \makecell{HE} & mbpp & \makecell{MPy} & \makecell{MCpp} & \makecell{Code\\Avg}  \\ \hline 
            Backbone & 44.5 &  21.4 & 34.2 & 23.0 & 26.2 & 41 &  39.6 & 34 & 38.2 & 46.3 &  42.9 & 44 & 48.4 & 45.4 \\
            Uniform Sampling & 43.9 & 23.0 & 42.1 & 22.5 & 28.9 & 40.5 & 39.2 & 35 & 38.2 & 38.7 & 38.2 & 38.2 & 39.7 & 38.7\\
            No Rule & \underline{51.1} & 23.1 & \underline{42.6} & 22.0 & 29.2 & \underline{42.3} & 37.4 & 37 & 38.9 & 45.1 & \underline{43.9} & 42.8 & 52.1 & 45.9\\
            DSIR & 50.2 & 22.5 & 32.4 & 17.0 & 23.9 & 41.5 & \textbf{41.1} & 34 & 38.9 &  45.1 & 43.6 & \textbf{49.1} & \underline{52.2} & \underline{47.5}\\
            LESS & 46.6 & 23.6 & 40.4 & \underline{24} & 29.3 & 41.5 & 40.4 & 33 & 38.3 & 41.4 & 43.5 & 43.9 & 45.3 & 43.5\\
            QuRating & 47.7	& 21.3 & 42.2 & 22 & 28.5 & 41.5 & 38.1 & 35 & 38.2 & 43.2 & 43.4 & 40.5 & 45.6 & 43.1\\
            GPT-Uncorrelated & 50.9 & \underline{23.7} & 42	& 22.7	& 29.4 & 41.4 & 39 & \underline{37.3} & \underline{39.2} & 41.2 & 43.5 & 39.6 & 48.6 & 43.2\\
            \hline
            AllData & 45.5 & 21.9 & 37.8 & 24.0	& 27.9 & 41.8 & 37.7 & 33.0 & 37.5 & 45.1 & 43.8 & 42.2 & 52.1 & 45.8\\
            \hline
            DPP 10 Rules (ours) & \textbf{53.5} & \textbf{24.6} &  \textbf{43.3} & \textbf{26.8} & \textbf{31.6} & \textbf{43.7} &  \underline{40.6} &  \textbf{38} & \textbf{40.8} &\textbf{50.5} &  \textbf{44.2} & \underline{46.9} & \textbf{52.7} & \textbf{48.6}\\ \hline
    \end{tabular}
    \caption{Fine-tuning on IMDB, Medical, Math and Code domains, each using 20K selected data samples from SlimPajama. The first row shows the performance of the original backbone model (Pythia-1B for IMDB and Medical and Llama3-8B for Math and Code). Abbreviations for subtasks: SA = Sentimental Analysis accuracy, CM = College Medicine, PM = Professional Medicine, MG = Medical Genetics, ES = Elementary School Math, HS = High School Math, CMath = College Math, HE = HumanEval, MPy = Multiple-py, MCpp = Multiple-cpp. For \textit{Uniform Sampling} and methods involving randomness in rule selections, we conducted 3 independent trials and averaged the results. The standard deviations are reported in Appendix~\ref{subsec:Appendix-EvalB-Trials}.
    }
    \label{tab:Finetune}
\end{table*}

 For each domain, we select 20K relevant training samples using different selection strategies. Results are shown in Table~\ref{tab:Finetune}. We first note that training with all 1M data gives underperforming results, aligning with findings from prior works such as \citet{zhou2023lima}. \textit{Uniform Sampling} generally fails to yield significant improvements. While \textit{DSIR} performs well on certain tasks, it relies on access to a high-quality reference dataset or even the test dataset, which may not always be feasible.
Among all rule-based methods, \textit{QuRating} underperforms. As previously noted, its limited hand-crafted rules may reflect subjective preferences or omit task-dependent correlations between rules. The \textit{GPT-Uncorrelated} rules face a similar issue where the rule selection process is entirely independent of the data. In contrast, our method first generates a large rule pool and then selects a subset based on rule orthogonality derived from data-specific score vectors. This task-aware design, grounded in a rigorous metric, leads to improved fine-tuning outcomes across all domains.

\subsection{Ablation Study}\label{subsec:AblationStudy}
To demonstrate the effectiveness of DPP in rule selection, we conduct an ablation study and evaluate the performance of the following variations developed within our framework:
\begin{itemize}
    \setlength\itemsep{0em}
    \item \textit{GPT-selected 10 Rules}: Query GPT to select 10 ``uncorrelated'' rules from the rule pool.
    \item \textit{All 50 Rules}: Average score vectors from all 50 rules to rate and select data.
    \item \textit{Random 10 Rules}: Randomly choose 10 rules and average the score vectors to rate and select data.
\end{itemize}
Results are provided in Appendix~\ref{sec:Appendix-AblationStudy}. Key findings include:
\textit{DPP 10 Rules} consistently outperforms both \textit{All 50 Rules} and \textit{Random 10 Rules}. This aligns with our argument of the importance of rule orthogonality, as well as the intuition that the optimal $r$ is not near the boundaries (both validated by experiments in Section~\ref{sec:EvaluationA}). 
\textit{GPT-selected 10 Rules} provides a strong baseline, but since it relies solely on semantic features and not task-specific scores, it is generally outperformed by our DPP-based selection. These results confirm that our DPP-based selection mechanism effectively identifies a compact, high-quality, and diverse rule subset tailored to the target task, resulting in superior data selection and model performance.

\section{Conclusion}\label{sec:Conclusion}

We presented an automated, rule-based framework for selecting high-quality training data for LLMs, combining LLM-generated rules with a DPP-based selection mechanism to promote rule diversity and reduce redundancy. Our work is the first to introduce a principled, automated metric for rule evaluation, integrated into a fully data-driven pipeline that generalizes effectively across diverse domains and tasks. We first validated our approach on a dataset with ground truth quality scores, demonstrating improved rating accuracy. We then fine-tuned LLMs using data selected by our method and showed consistent improvements over strong baselines across four domains. These results confirm that selecting a compact set of diverse, task-aware rules leads to higher-quality data and ultimately better model performance.

Our framework is broadly applicable to various scenarios, including general LLM pre-training, domain-specific fine-tuning, and RLHF data annotation. Moreover, it supports natural extensions such as rule re-weighting, enabled by the orthogonality of selected rules. Future work includes exploring these extensions and adapting the framework for settings requiring dynamic or task-conditioned rule weighting.

\bibliography{aaai2026}


\setcounter{secnumdepth}{2}
\clearpage
\appendix





\section{Proof of Theorem~\ref{thm:RuleCorrelation}} \label{sec:Appendix-ProofTheorm}

Recall we have $R$ candidate rules and a batch of $n$ data samples, denoted $\{\vx_1, \vx_2,\dots, \vx_n\}$. Each rule $j\in\{1,\dots,R\}$ ideally corresponds to an unknown ``ground-truth'' rating function $f_j(\vx_i)$, which assesses the data sample $\vx_i$ along direction of rule $j$. However, in practice, we do not observe $f_j(\vx_i)$ directly. Instead, we rely on an LLM-as-a-judge to produce an estimation
\begin{equation}
    S_{i,j} \;=\; f_j(\vx_i) + \epsilon_{i,j},
\end{equation}
where $\epsilon_{i,j}$ is an annotation error (or noise) incurred by the LLM. Hence we can model each rating score as a random variable. Then we can invoke concentration results to show that the empirical correlations converge to the true correlations as $n$ grows. This justifies that for a large enough $n$, such as the batch size $n=10^4$ used in Section~\ref{sec:EvaluationB}, is enough to generate rating scores to approximate the correlation between the rules.

\begin{lemma}[Max Entry-wise Deviation]\label{lem:Appendix-MaxEntryDeviation}
Define the random vector
\[
\vx^{(i)}
=
\left( S_{i,1}, S_{i,2}, \dots, S_{i,r} \right)
\in [0,1]^r,
\]
so that $\vx^{(i)}$ collects all rule scores for sample $i$. Assume $\{\vx^{(1)}, \dots, \vx^{(n)}\}$ are i.i.d.\ draws in $[0,1]^r$. Let  $\mSigma \bydef \mathrm{Cov}(\vx) \in \mathbb{R}^{r\times r}$ be the true covariance matrix of $\vx$, and 
\[
\rho_{j,\ell} \bydef \frac{\mSigma_{j,\ell}}{\sqrt{\mSigma_{j,j} \mSigma_{\ell,\ell}}}
\]
be the true correlation between rules $j$ and $\ell$, and denote true correlation matrix $\mC \bydef \left(\rho_{j,\ell}\right)_{1 \le j,\ell \le r}$. From the $n$ samples, we form the empirical covariance $\widehat{\mSigma}$ and correlation matrix $\widehat{\mC} \bydef \left(\widehat{\rho}_{j,\ell}\right)_{1 \le j,\ell \le r}$, where
\[
\widehat{\rho}_{j,\ell}
\bydef 
\frac{\widehat{\mSigma}_{j,\ell}}{\sqrt{\widehat{\mSigma}_{j,j} \widehat{\mSigma}_{\ell,\ell}}}.
\]

Assume each rule $j$ has nontrivial variance $(\mSigma_{j,j}$ bounded below by a positive constant), and all entries remain in $[0,1]$. Then there exists a universal constant $c > 0$ such that for any $\delta > 0$, with probability at least $1 - \delta$,
\[
\left\| \widehat{\mC} - \mC \right\|_{\max}
\bydef
\max_{1 \le j,\ell \le r}
\left| \widehat{\rho}_{j,\ell} - \rho_{j,\ell} \right|
\leq
c \sqrt{\frac{\ln\left(\frac{r^2}{\delta}\right)}{n}}.
\]

\end{lemma}

\begin{proof}
    
We use Hoeffding’s inequality and a union bound over all pairs of coordinates $(i,j)$ to show the concentration. Define the true mean of $\vx_i$ as $\mu_i = \E[\vx_i]$. The sample mean is

\[
\widehat{\mu}_i = \frac{1}{n} \sum_{k=1}^{n} x_i^{(k)}.
\]

Since $x_i^{(k)} \in [0,1]$, the Hoeffding’s inequality on the bounded random variables implies that for any $\delta>0$,

\[
\mathbb{P} \left( \left| \widehat{\mu}_i - \mu_i \right| > \epsilon \right)
\leq 2 \exp \left( -2 n \epsilon^2 \right).
\]

Taking $\epsilon = \sqrt{\frac{\ln(r/\delta)}{2n}}$, we see that $\widehat{\mu}_i$ concentrates around $\mu_i$. By applying a union bound over $i = 1, \dots, r$, all $\widehat{\mu}_i$ are close to $\mu_i$ with high probability.

A similar argument applies to each pair $(i,j)$ when estimating covariance. Recall that

\[
\mSigma_{i,j} \bydef \E \left[ (x_i - \mu_i)(x_j - \mu_j) \right],
\]

and the sample covariance is given by

\[
\widehat{\mSigma}_{i,j} \bydef \frac{1}{n} \sum_{k=1}^{n} 
\left( x_i^{(k)} - \widehat{\mu}_i \right) 
\left( x_j^{(k)} - \widehat{\mu}_j \right).
\]

\textbf{Step 1: Concentration Using True Means.}

First, we define an ``ideal'' version of the sample covariance using the \emph{true} means $\mu_j$:

\[
\widetilde{\mSigma}_{j,\ell}
\bydef
\frac{1}{n} \sum_{k=1}^{n}
\left( x_j^{(k)} - \mu_j \right)
\left( x_\ell^{(k)} - \mu_\ell \right).
\]

It is easy to see that  $\E \left[ \widetilde{\mSigma}_{j,\ell} \right] = \mSigma_{j,\ell}$ since $\mu_j = \E[x_j]$. Thus, to bound $\widetilde{\mSigma}_{j,\ell}$ around $\mSigma_{j,\ell}$, we notice that each product
$\left( x_j^{(k)} - \mu_j \right) \left( x_\ell^{(k)} - \mu_\ell \right) $ lies in $[-1,1]$ (since $x_j^{(k)} \in [0,1]$ implies $(x_j^{(k)} - \mu_j) \in [-1,1]$, etc.). One can apply Hoeffding’s inequality to the i.i.d. random variables
\[
Z^{(k)}_{j,\ell}
\bydef
\left( x_j^{(k)} - \mu_j \right)
\left( x_\ell^{(k)} - \mu_\ell \right),
\]

each bounded in $[-1,1]$. Then:

\[
\mathbb{P} \left( \left| \widetilde{\mSigma}_{j,\ell} - \mSigma_{j,\ell} \right|
> \epsilon \right)
\leq
2 \exp \left( -\frac{n \epsilon^2}{2} \right),
\]

up to constant factors. Setting $\epsilon = \sqrt{\frac{2 \ln(r^2/\delta)}{n}}$, we can get that each $(j, \ell)$-th entry fails with probability at most $\frac{2\delta}{r^2}$. Then applying a union bound over all $\frac{r(r+1)}{2} \approx \frac{r^2}{2}$ pairs $(j,\ell)$ yields

\begin{equation}
    \max_{j,\ell} \left| \widetilde{\mSigma}_{j,\ell} - \mSigma_{j,\ell} \right|
\leq
C_1 \sqrt{\frac{\ln\left(\frac{r^2}{\delta}\right)}{n}}
\quad
\text{with probability at least } 1 - \delta,
\end{equation}

for some absolute constant $C_1$.

\textbf{Step 2: Relating $\widetilde{\mSigma}$ to $\widehat{\mSigma}$.}

Next, we note that $\widetilde{\mSigma}_{j,\ell}$ uses true means $\mu_j$, whereas $\widehat{\mSigma}_{j,\ell}$ uses empirical means $\widehat{\mu}_j$. We claim that the discrepancy between these two forms is also $\mathcal{O}(\sqrt{\ln(r^2/\delta)/n})$. Note that we have
\begin{align*}
    \widehat{\mSigma}_{j,\ell}
&=
\frac{1}{n}
\sum_{k=1}^{n}
\left( x_j^{(k)} - \widehat{\mu}_j \right)
\left( x_\ell^{(k)} - \widehat{\mu}_\ell \right) \\
&= \frac{1}{n} \sum_{k=1}^{n}
\left( x_j^{(k)} - \mu_j + \mu_j - \widehat{\mu}_j \right)
\left( x_\ell^{(k)} - \mu_\ell + \mu_\ell - \widehat{\mu}_\ell \right)\\
&= \frac{1}{n} \sum_{k=1}^{n}
\left( x_j^{(k)} - \mu_j \right) \left( x_\ell^{(k)} - \mu_\ell \right)
+ \left(\mu_j - \widehat{\mu}_j \right) \left( x_\ell^{(k)} - \mu_\ell \right)\\
&\quad + \left( x_j^{(k)} - \mu_j \right) \left(\mu_\ell - \widehat{\mu}_\ell \right)
+ \left(\mu_j - \widehat{\mu}_j \right) \left(\mu_\ell - \widehat{\mu}_\ell \right)\\
&= 
\widetilde{\mSigma}_{j,\ell}
- (\widehat{\mu}_j - \mu_j)(\widehat{\mu}_\ell - \mu_\ell)
+ \Delta_{j,\ell},
\end{align*}
where $\Delta_{j,\ell}$ collects a cross-term that depends on $\widehat{\mu}_j - \mu_j$ and the average $\frac{1}{n}\sum_{k=1}^{n} (x_j^{(k)} - \mu_j)$.

Recall we have shown that $\left| \widehat{\mu}_j - \mu_j \right|$ is typically $\mathcal{O}\left(\sqrt{\frac{\ln(r/\delta)}{n}}\right)$, and each average $\frac{1}{n}\sum_{k=1}^{n} \left( x_j^{(k)} - \mu_j \right)$ (using Hoeffding's again). Hence, the difference $\widehat{\mSigma}_{j,\ell} - \widetilde{\mSigma}_{j,\ell}$ is bounded by the same order $\sqrt{\frac{\ln(r^2/\delta)}{n}}$. Therefore, we get
\begin{equation}
    \max_{j,\ell} \left| 
\widehat{\mSigma}_{j,\ell} - \mSigma_{j,\ell}
\right|
\leq
C_3 \sqrt{\frac{\ln\left(\frac{r^2}{\delta}\right)}{n}}
\end{equation}

with probability at least $1 - \delta$, for some universal constant $C_3$.

\textbf{Step 3: Translating Covariance Error to Correlation Error.}

Finally, we translate the result on covariance to correlation. Recall that

\[
\rho_{j,\ell}
=
\frac{\mSigma_{j,\ell}}{\sqrt{\mSigma_{j,j} \mSigma_{\ell,\ell}}},
\quad
\widehat{\rho}_{j,\ell}
=
\frac{\widehat{\mSigma}_{j,\ell}}{\sqrt{\widehat{\mSigma}_{j,j} \widehat{\mSigma}_{\ell,\ell}}}.
\]

Under the assumption that $\mSigma_{j,j}$ are bounded below by some positive constant, say $\sigma_{\min}^2$. We have $\mSigma_{j,j} \geq \sigma_{\min}^2 > 0$ and  $\widehat{\mSigma}_{j,j} \approx \mSigma_{j,j}$ with high probability.
We can make $n$ large enough such that $\mathcal{O} \left(\sqrt{\tfrac{\ln(r^2/\delta)}{n}}\right)\le \frac{1}{2}\,\sigma_{\min}^2$, then it follows that  $\widehat{\mSigma}_{j,j}$ remains above $\frac{1}{2} \sigma_{\min}^2$ for large $n$. After showing the closeness of $\widehat{\mSigma}_{j,\ell}$ and $\mSigma_{j,\ell}$ in Step 2, the closeness of $ \widehat{\rho}_{j,\ell}$ and $\rho_{j,\ell}$ is straightforward. For example, since denominators $\widehat{\mSigma}_{j,j}, \widehat{\mSigma}_{\ell,\ell}$ are not too small, we can use a ratio-Lipschitz argument to prove 
\begin{equation}
    \left| \widehat{\rho}_{j,\ell} - \rho_{j,\ell} \right|
\leq
\mathrm{poly} \left( \max_{j,\ell} \left| \widehat{\mSigma}_{j,\ell} - \mSigma_{j,\ell} \right| \right)
=
\mathcal{O} \left( \sqrt{\frac{\ln(r^2/\delta)}{n}} \right).
\end{equation}
This holds across all $(j,\ell)$, and hence we obtain our goal:
\begin{equation}
    \|\widehat{\mathbf{C}} - \mathbf{C}\|_{\max}
=
\max_{j,\ell}
\left| \widehat{\rho}_{j,\ell} - \rho_{j,\ell} \right|
=
\mathcal{O} \left( \sqrt{\frac{\ln(r^2/\delta)}{n}} \right),
\end{equation}
with probability at least $1 - \delta$. This completes the proof.
\end{proof}

\begin{lemma}[Frobenius-Norm Deviation]\label{lem:Appendix-MatrixDeviation}
Under the same setting of Lemma~\ref{lem:Appendix-MaxEntryDeviation}, we have that the Frobenius norm satisfies
\[
\|\widehat{\mC} - \mC\|_F
\leq
r \cdot c \sqrt{\frac{\ln\left(\frac{r^2}{\delta}\right)}{n}},
\]
Thus, when $n$ is sufficiently large such that $\sqrt{\ln(r^2/\delta)/n}$ is small, the entire sample correlation matrix $\widehat{\mC}$ is uniformly close to the true correlation matrix $C$ in all entries.
\end{lemma}

\begin{proof}
By definition,

\[
\rho_{i,j} = \frac{\mSigma_{i,j}}{\sqrt{\mSigma_{i,i} \mSigma_{j,j}}}
\quad \text{and} \quad
\widehat{\rho}_{i,j} = \frac{\widehat{\mSigma}_{i,j}}{\sqrt{\widehat{\mSigma}_{i,i} \widehat{\mSigma}_{j,j}}}.
\]

From Lemma~\ref{lem:Appendix-MaxEntryDeviation} we have
\[
\max_{j, \ell} \left| \widehat{\rho}_{j, \ell} - \rho_{j, \ell} \right|
\leq c \sqrt{\frac{\ln(r^2/\delta)}{n}},
\]
for some constant $c$. Once each entry $\widehat{\rho}_{i,j}$ is within $\epsilon$ of $\rho_{i,j}$, the Frobenius norm difference is controlled by

\[
\|\widehat{\mC} - \mC\|_F^2
= \sum_{i,j=1}^{r} \left| \widehat{\rho}_{i,j} - \rho_{i,j} \right|^2
\leq r^2 \epsilon^2.
\]

Taking the square root yields

\[
\|\widehat{\mC} - \mC\|_F
\leq r \epsilon.
\]

Choosing $\epsilon$ on the order of $\sqrt{\frac{\ln(r^2/\delta)}{n}}$ completes the proof.

\end{proof}

\textbf{Proof of Theorem~\ref{thm:RuleCorrelation}.}
By Lemma~\ref{lem:Appendix-MaxEntryDeviation} and Lemma~\ref{lem:Appendix-MatrixDeviation}, we know that  
\[
\|\widehat{\mC} - \mC\|_F
=
\mathcal{O} \left( r \sqrt{\frac{\ln(r^2/\delta)}{n}} \right),
\]
with high probability $1 - \delta$. Using the triangle inequality, we get
\begin{align*}
    \left\|\widehat{\mC} - \mathbf{I}_r\right\|_F
&=
\left\| (\widehat{\mC} - \mC) + (\mC - \mathbf{I}_r) \right\|_F\\
&\leq
\|\widehat{\mC} - \mC\|_F + \|\mC - \mathbf{I}_r\|_F.
\end{align*}
Hence,
\[
\frac{1}{r} \left\|\widehat{\mC} - \mathbf{I}_r\right\|_F
\leq
\frac{1}{r} \|\mC - \mathbf{I}_r\|_F
+
\frac{1}{r} \|\widehat{\mC} - \mC\|_F.
\]

Similarly, reversing the roles of $\widehat{\mC}$ and $\mC$ in that inequality shows  

\[
\frac{1}{r} \|\mC - \mathbf{I}_r\|_F
\leq
\frac{1}{r} \left\|\widehat{\mC} - \mathbf{I}_r\right\|_F
+
\frac{1}{r} \|\widehat{\mC} - \mC\|_F.
\]

Combining both results,

\[
\left|
\frac{1}{r} \|\widehat{\mC} - \mathbf{I}_r\|_F - \frac{1}{r} \|\mC - \mathbf{I}_r\|_F
\right|
\leq
\frac{1}{r} \|\widehat{\mC} - \mC\|_F.
\]
Now apply Lemma~\ref{lem:Appendix-MatrixDeviation}, we can bound $\|\widehat{\mC} - \mC\|_F \leq r c \sqrt{\frac{\ln(r^2/\delta)}{n}}$ with probability $1 - \delta$. Thus, $ \frac{1}{r} \|\widehat{\mC} - \mC\|_F \leq c \sqrt{\frac{\ln(r^2/\delta)}{n}}$, which implies
\begin{align*}
    \left|
\rho(\bmS) - \frac{1}{r} \|\mC - \mathbf{I}_r\|_F
\right|
&=
\left|
\frac{1}{r} \|\widehat{\mC} - \mathbf{I}_r\|_F - \frac{1}{r} \|\mC - \mathbf{I}_r\|_F
\right|\\
&\leq
c \sqrt{\frac{\ln(r^2/\delta)}{n}}
\end{align*}
with probability at least $1-\delta$. This completes the proof. $\hfill\square$

\section{Ablation Study}\label{sec:Appendix-AblationStudy}
For the ablation study, we compare with these methods to demonstrate the advantages of DPP in rule selection:
\begin{itemize}
    \setlength\itemsep{0em}
    \item \textit{GPT-selected 10 Rules}: Query GPT to select 10 ``uncorrelated'' rules from the rule pool.
    \item \textit{All 50 Rules}: Average score vectors from all 50 rules to rate and select data.
    \item \textit{Random 10 Rules}: Randomly choose 10 rules and average the score vectors to rate and select data.
    \item \textit{DPP 10 Rules}: Use DPP to sample 10 rules, and then apply them to rate and select data.
\end{itemize}
Note that for uniform sampling and the methods involving randomness in the selection of rules, we considered 3 independent trials and averaged the results.
\begin{table*}[ht]
    \centering
    \mytiny
    \renewcommand{\arraystretch}{1.1} 
    \setlength{\tabcolsep}{0.6pt}
\begin{tabular}{ c | c | c  c  c  c | c  c  c  c | c  c  c  c  c } \hline 
   \multirow{2}{*}{\textbf{Method}} & \textbf{IMDB} & \multicolumn{4}{c|}{\textbf{Medical}} & \multicolumn{4}{c|}{\textbf{Math}} & \multicolumn{5}{c}{\textbf{Code}} \\
    \cline{2-15} 
    & \makecell{SA\\accuracy} & \makecell{college\\medicine} & \makecell{professional\\medicine} & \makecell{medical\\genetics} & \makecell{Medical\\Average} 
    & \makecell{elementary\\school}  & \makecell{high\\school}  &   \makecell{college\\math} & \makecell{Math\\Average} 
    & \makecell{human\\-eval} & mbpp & \makecell{multiple\\-py} & \makecell{multiple\\-cpp}  & \makecell{Code\\Average}  \\ \hline 
            
            GPT-selected 10 Rules & 51.6 & 21.9 & 42.2 & \underline{24} & 29.4 & 42.5 & 40.3 & \underline{36} & \underline{39.6} & 40.8 & \underline{43.4} & 42.8 & \underline{50.3} & 44.3 \\
            All 50 Rules & 51 &  23.1 & \underline{42.6} & 23 & \underline{29.6} & 41.8 &  \textbf{40.7} & 33 & 38.5 & 43.9 &  \underline{43.4} & \underline{46.6} & 49.1 & 45.8\\
            Random 10 Rules & \underline{51.7} & \underline{24} & 41.2 & 23.5 & \underline{29.6} & \underline{42.9} & 39.8 & 35.2 & 39.3 & \underline{48.5} & 41 & \underline{46.6} & 48.1 & \underline{46}\\
            DPP 10 Rules & \textbf{53.5} & \textbf{24.6} &  \textbf{43.3} & \textbf{26.8} & \textbf{31.6} & \textbf{43.7} &  \underline{40.6} &  \textbf{38} & \textbf{40.8} &\textbf{50.5} &  \textbf{44.2} & \textbf{46.9} & \textbf{52.7} & \textbf{48.6}\\ \hline
    \end{tabular}
    \caption{Fine-tuning Llama3-8B on IMDB, Medical, Math and Code domains, each using 20K selected data samples from SlimPajama. The first row shows the performance of the original backbone model (Pythia-1B for IMDB and Medical and Llama3-8B for Math and Code).
    }
    \label{tab:AblationStudy}
\end{table*}

\textbf{Rule Correlation of selected rules.}
Here we provide in Table~\ref{tab:Appendix-EvalB-RuleIndices} the rule indices (in the range $\{0,1,\dots, 49\}$) for the rules selected by DPP and random selection (in one trial). The complete list of all generated $50$ rules for each domain is provided in \ref{subsec:Appendix-EvalB-Prompts-Rules}. For each set of selected rules, we also calculate their rule correlation value $\rho$ (defined in \ref{eq:rule_correlation}). We confirm that indeed DPP selects rules with lower rule correlation than random selected rules.

\begin{table}[H]
\centering
\small 
\setlength{\tabcolsep}{1pt}            
\renewcommand{\arraystretch}{1.2}      
\caption{Rule correlation and indices of the selected rules by DPP and random selection (two trials ``A'' and ``B'').}
\label{tab:Appendix-EvalB-RuleIndices}
\resizebox{1.0\columnwidth}{!}{%
\begin{tabular}{c|c|c} 
\hline
\textbf{Domain} & \textbf{Method} & \textbf{Rule Indices} \\ 
\hline
\multirow{3}{*}{IMDB} & DPP 10 rules ($\rho=0.42$) & [2, 3, 13, 21, 28, 36, 37, 45, 46, 49] \\
 & Random 10 Rules A ($\rho=0.53$) & [2, 6, 10, 11, 15, 21, 28, 42, 43, 48] \\
 & Random 10 Rules B ($\rho=0.51$) & [1, 9, 12, 14, 25, 26, 27, 37, 38, 40] \\
\hline
\multirow{3}{*}{Medical} & DPP 10 rules ($\rho=0.55$) & [1, 9, 10, 25, 29, 30, 32, 38, 42, 47] \\
 & Random 10 Rules A ($\rho=0.69$) & [11, 13, 14, 16, 25, 33, 34, 43, 45, 49] \\
 & Random 10 Rules B ($\rho=0.66$) & [6, 17, 20, 28, 29, 37, 40, 41, 47, 48] \\
\hline
\multirow{3}{*}{Math} & DPP 10 rules ($\rho=0.40$) & [0, 4, 13, 26, 27, 31, 33, 38, 44, 45] \\
 & Random 10 Rules A ($\rho=0.65$) & [0, 2, 11, 16, 17, 18, 27, 28, 34, 39] \\
 & Random 10 Rules B ($\rho=0.61$) & [3, 4, 25, 20, 23, 13, 15, 24, 35, 39] \\
\hline
\multirow{3}{*}{Code} & DPP 10 rules ($\rho=0.54$) & [2, 3, 13, 21, 28, 36, 37, 45, 46, 49] \\
 & Random 10 Rules A ($\rho=0.59$) & [5, 7, 10, 13, 17, 19, 21, 26, 30, 34] \\
 & Random 10 Rules B ($\rho=0.58$) & [2, 4, 8, 14, 16, 20, 23, 33, 37, 44] \\
\hline
\end{tabular}
}
\end{table}

In Table~\ref{tab:Appendix-EvalB-RuleIndices-GPTselected10} below, we calculate the rule correlation selected by GPT. By comparing to the rule correlations of DPP-selected rules in Table~\ref{tab:Appendix-EvalB-RuleIndices}, we see that although we prompt GPT-4 to select ``uncorrelated'' rules, the rule correlations of the selected 10 rules are still higher than our DPP-selected rules. 

\begin{table}[H]
\centering
\small 
\setlength{\tabcolsep}{1pt}
\renewcommand{\arraystretch}{1.2}
\caption{Rules selected by GPT and their rule correlation values.}
\label{tab:Appendix-EvalB-RuleIndices-GPTselected10}
\resizebox{1.0\columnwidth}{!}{%
\begin{tabular}{c|c|c} 
\hline
\textbf{Domain} & \textbf{Method} & \textbf{Rule Indices} \\ 
\hline
\multirow{1}{*}{IMDB} & GPT selected 10 rules ($\rho = 0.67$) & [0, 1, 4, 10, 13, 17, 25, 31, 40, 49] \\
\cline{2-3}
\hline
\multirow{1}{*}{Medical} & GPT selected 10 rules  ($\rho = 0.40$) & [0, 4, 7, 11, 15, 24, 29, 34, 42, 49] \\
\hline
\multirow{1}{*}{Math} & GPT selected 10 rules  ($\rho = 0.56$) & [0, 3, 7, 11, 17, 24, 28, 38, 44, 48]\\
\hline
\multirow{1}{*}{Code} & GPT selected 10 rules  ($\rho = 0.65$) & [0, 4, 9, 12, 16, 23, 29, 34, 43, 49] \\
\hline
\end{tabular}
}
\end{table}


\section{Limitations and Future Directions}\label{sec:Appendix-Limitations-FutureDirections}
We have developed an automated, rule-based selection framework for identifying high-quality LLM data. Below, we outline some limitations of our approach and suggest potential directions for future research:

\textbf{Adjusting hyperparameters.} Recall that our hyperparameter $r$ determines the number of rules selected for rating, influencing the diversity and coverage of the selected rules. We have explored the effect of $r$ in Section~\ref{sec:EvaluationA} and also in Appendix~\ref{subsec:Appendix-EvalB-20rules}. We leave a comprehensive study of its optimal values for future work.

\textbf{Data sampling method.}
There are variations of the stochastic top-$k$ sampling, such as incorporating a temperature parameter $\tau$ (see \citet{wettig2024qurating}). Replacing \eqref{eq:top-k-sampling} with its variations or exploring other data sampling methods represents another research direction.
    
\textbf{Rule format.} In this study, we only focus on natural language rules, which are straightforward to design and offer significant explainability. However, rules in other formats can also be integrated into our pipeline.

\textbf{Other rule evaluations metrics.} We propose multiple metrics (in Equations~\eqref{eq:rule_correlation} and \eqref{eq:volume} below) to measure rule quality, but all based on the correlation/orthogonality of rules. Evaluating rules from other aspects is another intriguing topic for future work.

\section{Orthogonality Measures}\label{sec:Appendix-orthogonality-measures}

\textbf{Volume of parallelepiped.}
In our experiments, we also considered another measure of orthogonality, defined as the ``volume'' of the parallelepiped formed by vectors. This is mathematically described as:
\begin{equation}\label{eq:volume}
    \textbf{Vol}(\bmS) \bydef \frac{\sqrt{\det(\bmS^\top \bmS)}}{\Pi_{i=1}^r \|\vv_i\|},
\end{equation}
where $\vv_i$ are the columns of $\bmS$. The determinant of $\bmS^\top \bmS$ geometrically represents the squared volume of the parallelepiped formed by the columns of $\bmS$ \citep{kulesza2012determinantal}. We normalize by the product of the vector norms since both the magnitude of the vectors and their mutual correlation influence the volume: larger norms increase the volume, whereas higher correlation reduces it. Thus, after normalization, the value of \textbf{Vol} serves as an indicator of the overall orthogonality among the column vectors of $\bmS$. The phenomena under the usage of this measure are similar to the ones under \ref{eq:rule_correlation}. Therefore we only presented results using the rule correlation $\rho$.

\section{DPP Sampling}\label{sec:Appendix-DPPSampling}

\textbf{Intuition by $r=2$ case.} Here we use the $r=2$ case to illustrate the intuition behind DPP and explain why it tends to choose items that are relatively uncorrelated. Using the same notation as in \ref{subsec:DefinitionsNotations}, let $\mK$ be the kernel matrix and $\mathcal{Y}, Y$ be the ground set and selected subset respectively. When $r=2$, consider items $A = \{i,j\}$. Then the probability of both items being selected together is given by:
\begin{align*}
    &\mathbb{P}(A \subseteq  Y) 
    = K_{i,i}K_{j,j} - K_{i,j}K_{j,i}\\
    &= \mathbb{P}(i \in Y)\mathbb{P}(j \in Y) - K_{i,j}^2\\
    &= \mathbb{P}(i\text{ is chosen})\mathbb{P}(j\text{ is chosen}) - (\text{similarity of items $i,j$})^2,
\end{align*}
since $\mK$ is symmetric by our definition. Larger similarity of $i,j$ reduces the probability $\mathbb{P}(A \subseteq  Y)$, indicating that similar items are less likely to be chosen simultaneously. This underscores the DPP’s capacity to promote diversity by favoring the selection of dissimilar items.

\textbf{DPP Sampling Algorithm: } The sampling algorithm can be found in Algorithm 1 of \citet{kulesza2012determinantal}. The sampling process starts by decomposing the kernel matrix $\mK$ and involves two main stages: 1. Selecting eigenvectors by sampling from a Bernoulli distribution based on the eigenvalues, and 2. Sampling a subset from the ground set using an iterative conditional distribution method to ensure diversity, as detailed in \citep{kulesza2012determinantal}. We utilize the \texttt{DPPy} Python library \citep{GPBV19} for efficient DPP initialization and sampling.

\textbf{Time Complexity:} Finding the submatrix (subset of columns) of a matrix to maximize the orthogonality is NP-hard \citep{civril2007finding, kulesza2012determinantal}. DPP provides us a relatively good solution. In practice, the computational complexity of sampling from a DPP depends primarily on the eigendecomposition of the kernel matrix $\mK$. In our case, $\mK \in \R^{R \times R}$ and therefore it requires $O(R^3)$ time, where $R$ is the number of rules. In the \texttt{DPPy} package \citep{GPBV19} it uses the spectral sampler by default, so the actual run-time of our DPP implementation is $O(R^3)$.

\textbf{DPP Sampling for Data Selection:} We noticed that in a concurrent work \citep{yang2024p3}, the authors also use DPP to perform data selection, but directly applied to the data itself. However, the approach to directly perform data selection using DPP requires the computation based on the kernel matrix with dimension $N$ (number of samples), which is usually huge in the context of LLM data. Moreover, while DPP inherently prioritizes diversity in data selection, it does not address other quality dimensions. In contrast, our rule-based approach assesses multiple aspects of data quality, ensuring a more comprehensive and robust selection process.

\section{Stochastic data selection: Gumbel top-$k$ trick:}
Imagine the cases where the target dataset distribution shows a long-tail pattern with respect to our quality measure, using a deterministic quality score as the cutoff could exclude many possibly valuable data \citep{albalak2024survey}. Hence, our stochastic sampling in Equation~\eqref{eq:top-k-sampling} effectively balances the quality and diversity of the selected data. Nonetheless, instead of doing actual sampling according to Equation~\eqref{eq:top-k-sampling}, we use the Gumbel top-$k$ trick similar as in \citep{wettig2024qurating}, which is a sampling technique used to efficiently and probabilistically select the top-$k$ items from a discrete probability distribution. Specifically, each item $i$ in the distribution is assigned a score using the formula:
\[
s_i = \log p_i + g_i,
\]
where $p_i$  is the probability of item $i$, and $g_i$ is a noise term drawn from a Gumbel distribution, which can be generated using $g_i = -\log(-\log(u_i))$. In other words, we could add a Gumbel noise vector to the log of the sampling probability in \eqref{eq:top-k-sampling} and then choose the top-$k$ data points with the highest sums. This is statistically equivalent to sampling according to \eqref{eq:top-k-sampling} \citep{kool2019stochastic}.

\section{Rule Generator Variation: Claude-3.5-Sonnet}
To verify that GPT-4 is a reliable rule generator, we compare it with Claude-3.5-Sonnet. For each of the five tasks (General, IMDB, Medical, Math, Code), we prompt GPT and Claude to generate 100 rules for each, and then study the distribution of the rules. Specifically, we use Sentence-Transformer \citep{reimers2019sentence} to generate the embedding vectors and then use PCA to project them onto the top two principal components for 2-dimensional visualization. From Figure~\ref{fig:rules_embeddings} below, we observe that the two groups of rules generated by the two models completely overlap, demonstrating no distinct separation. This suggests that using strong LLMs such as GPT or Claude does not introduce clear model-specific bias during rule generation.

\begin{figure}[ht]
    \centering
    \begin{subfigure}[b]{0.49\columnwidth}
        \centering
        \includegraphics[width=\textwidth]{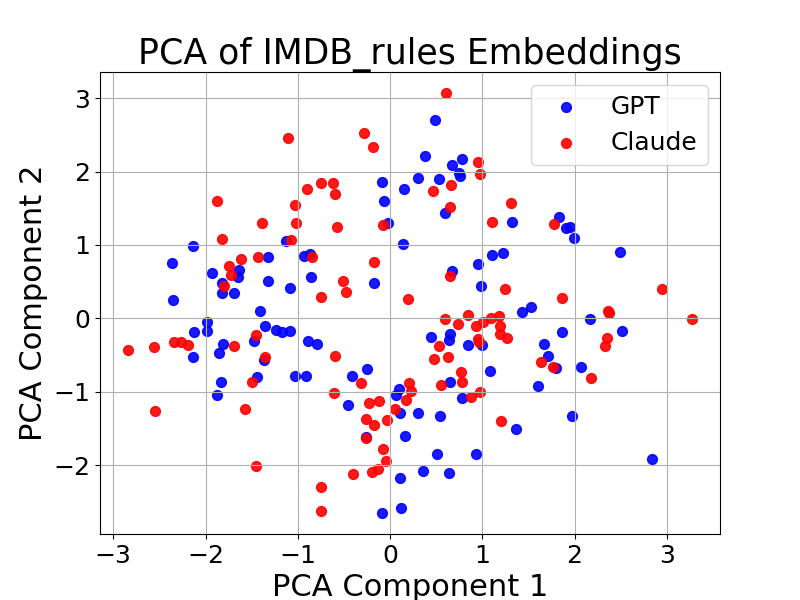}
        \caption{IMDB Rules}
        \label{fig:imdb_rules}
    \end{subfigure}
    \hfill
    \begin{subfigure}[b]{0.49\columnwidth}
        \centering
        \includegraphics[width=\textwidth]{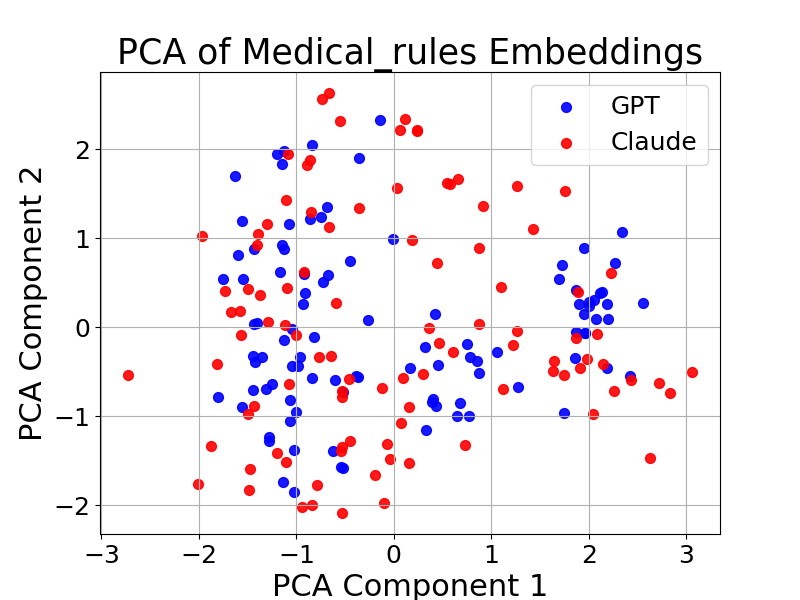}
        \caption{Medical Rules}
        \label{fig:medical_rules}
    \end{subfigure}
    \begin{subfigure}[b]{0.49\columnwidth}
        \centering
        \includegraphics[width=\textwidth]{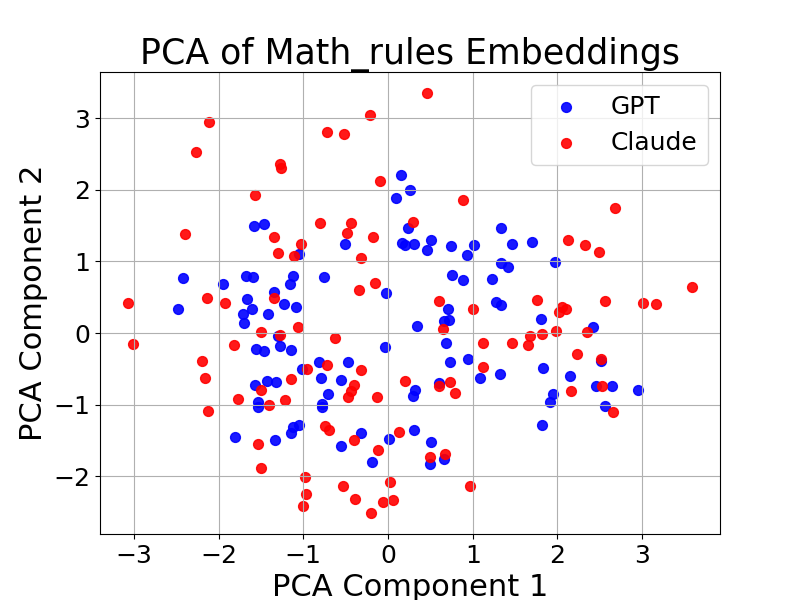}
        \caption{Math Rules}
        \label{fig:math_rules}
    \end{subfigure}
    \begin{subfigure}[b]{0.49\columnwidth}
        \centering
        \includegraphics[width=\textwidth]{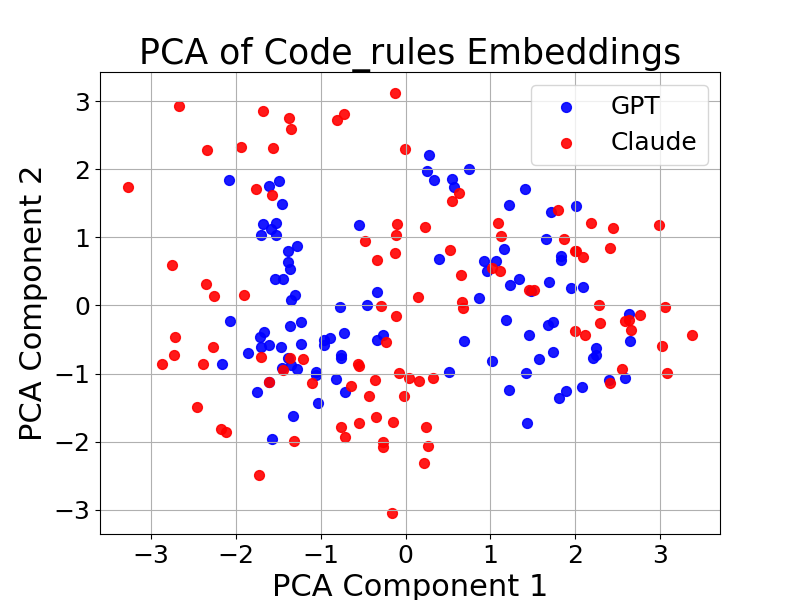}
        \caption{Code Rules}
        \label{fig:code_rules}
    \end{subfigure}
    \caption{Embeddings of rules generated by GPT and Claude across different domains.}
    \label{fig:rules_embeddings}
\end{figure}

To further quantify the distribution differences, we studied the Wasserstein distance within and between the two rule sets. Specifically, we compute the distance between the GPT rules and Claude rules. Then we randomly split GPT rules into two parts and computed their Wasserstein distance and similarly for the Claude rules (averaged over 10 trials). By comparing these values (see Table~\ref{tab:Appendix-EvalB-GPTClaude-Wasserstein}), we found no clear distribution bias when switching from one rule generator to the other.

\begin{table}[H]
\centering
\caption{Comparison of Intra-Model and Inter-Model Metrics across Domains}
\label{tab:model_metrics}
\begin{tabular}{@{}lcccccc@{}}
\toprule
               & General & IMDB  & Medical & Math  & Code  \\ \midrule
Intra-GPT      & 0.432   & 0.438 & 0.502   & 0.584 & 0.673 \\
Intra-Claude   & 0.455   & 0.445 & 0.515   & 0.592 & 0.668 \\
Inter-Model    & 0.468   & 0.548 & 0.522   & 0.612 & 0.684 \\ \bottomrule
\end{tabular}
\end{table}
\label{tab:Appendix-EvalB-GPTClaude-Wasserstein}

We have compared the rule generators using GPT and Claude above, both are LLMs. In order to address potential biases in rule generation by these LLMs compared to human-generated rules, we prompt GPT-4 to generate 133 ethical and safety rules and compare the GPT-generated rules with the public and standard constitutions in \citep{huang2024collective} (we make the rules all start from ``Choose the response that'' for a fair comparison). We asked 3 authors who have not seen the constitutions in \citep{huang2024collective} to distinguish the rules blindly. We get an average accuracy of $20.7\%$, suggesting it is indeed hard to distinguish between rules generated by GPT and those designed by humans. All these discussions underscore the potential of GPT as a reliable rule generator that is capable of producing rules that are comparable to those crafted by human experts.


\section{Appendix for Preliminary Experiments}\label{sec:Appendix-EvalA}

\subsection{Error Metrics}
\textbf{Ranking-difference error.}
To assess the deviation of rating scores from the ground truth, instead of using the mean squared error in \ref{eq:mse}, an alternative intuitive approach is to compare the rankings derived from the data scores with those of the ground truth. This approach is based on the premise that for data selection purposes, if two sets of scores yield identical rankings, they will select the same high-scoring data samples. An example of such a ranking metric is the Kendall rank correlation coefficient (Kendall’s tau) \citep{kendall1938new}. However, we opted against this type of metric for two critical reasons: First, it lacks the granularity needed to evaluate errors effectively. For instance, two sets of scores like [0.01, 0.98, 0.99] and [0.01, 0.02, 0.03] share exactly the same ranking yet differ significantly in their actual scores. Second, our method involves stochastic data selection, not a straightforward top-$k$ selection, meaning that a higher score increases the likelihood of a data point being chosen. Hence, a ranking difference, which overlooks the absolute values of scores and focuses solely on their relative comparisons, is not ideal here.


\subsection{Other domains: Medical, Math and Code}\label{subsec:Appendix-EvalA-AllDomains}
\begin{figure*}[ht]
\centering

\begin{subfigure}[b]{0.19\linewidth}
    \centering
    \includegraphics[width=\linewidth]{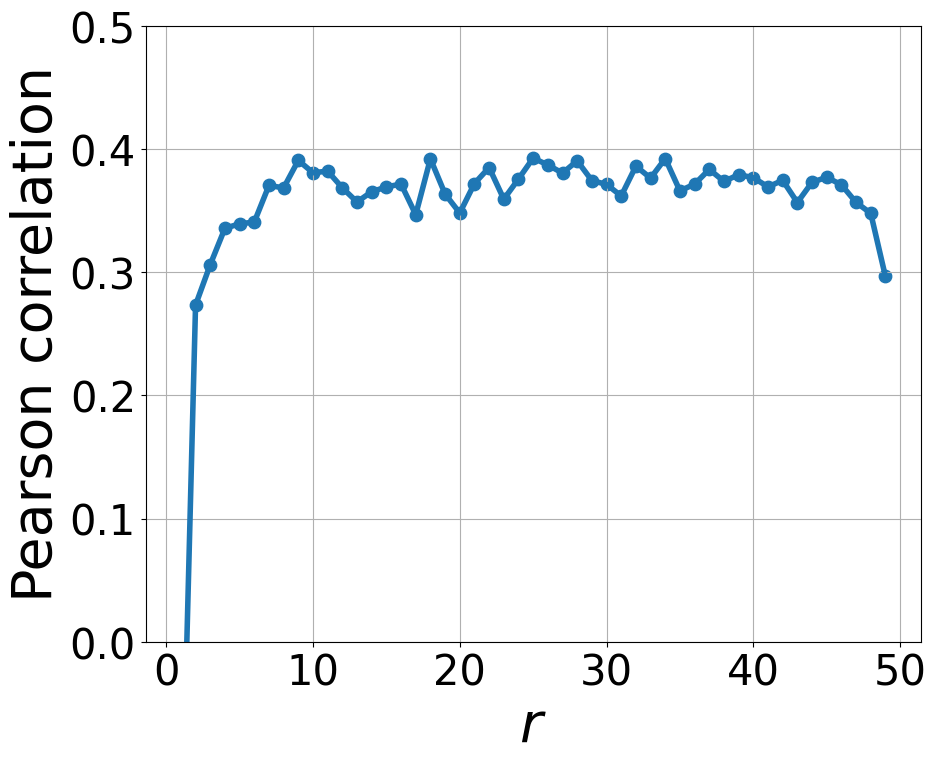}
    \caption{}
\end{subfigure}
\begin{subfigure}[b]{0.19\linewidth}
    \centering
    \includegraphics[width=\linewidth]{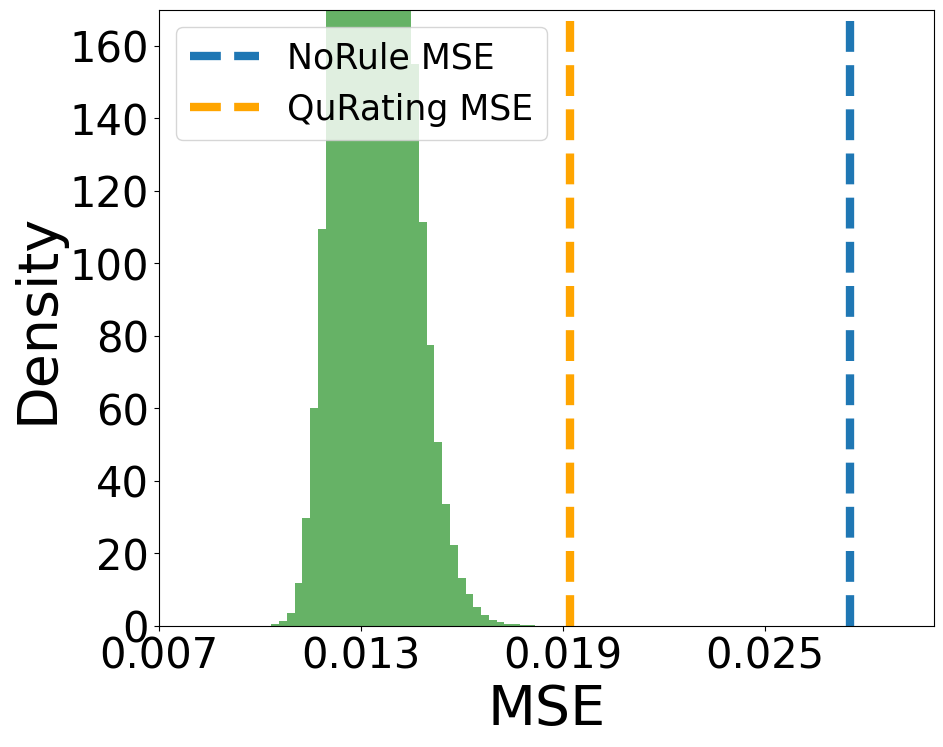}
    \caption{}
\end{subfigure}
\begin{subfigure}[b]{0.19\linewidth}
    \centering
    \includegraphics[width=\linewidth]{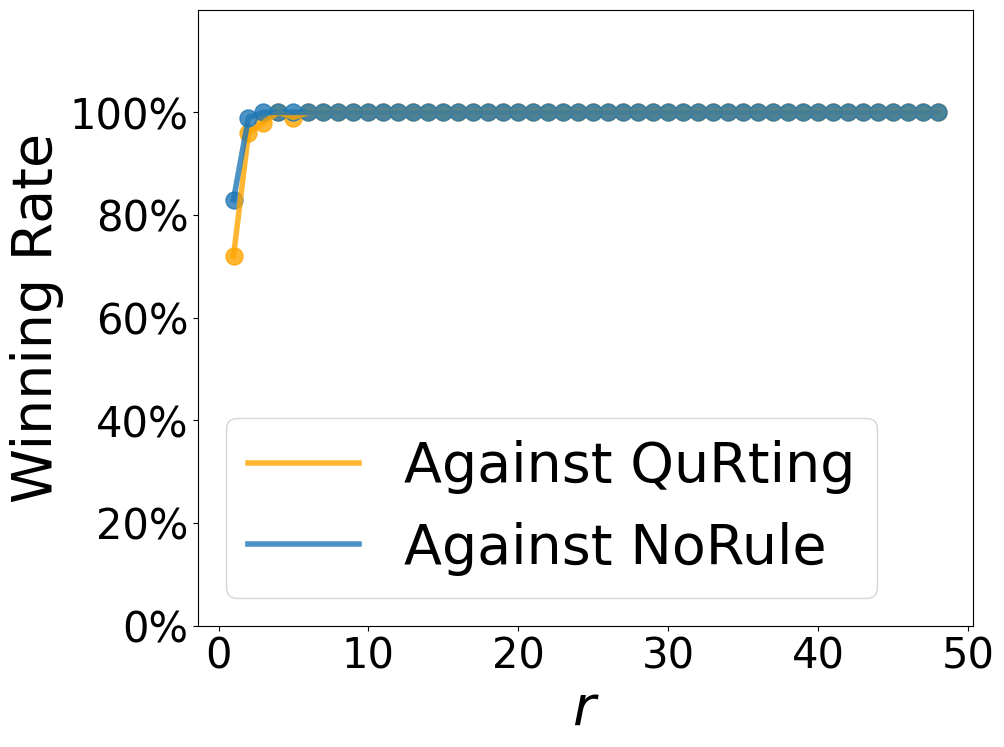}
    \caption{}
\end{subfigure}
\begin{subfigure}[b]{0.19\linewidth}
    \centering
    \includegraphics[width=\linewidth]{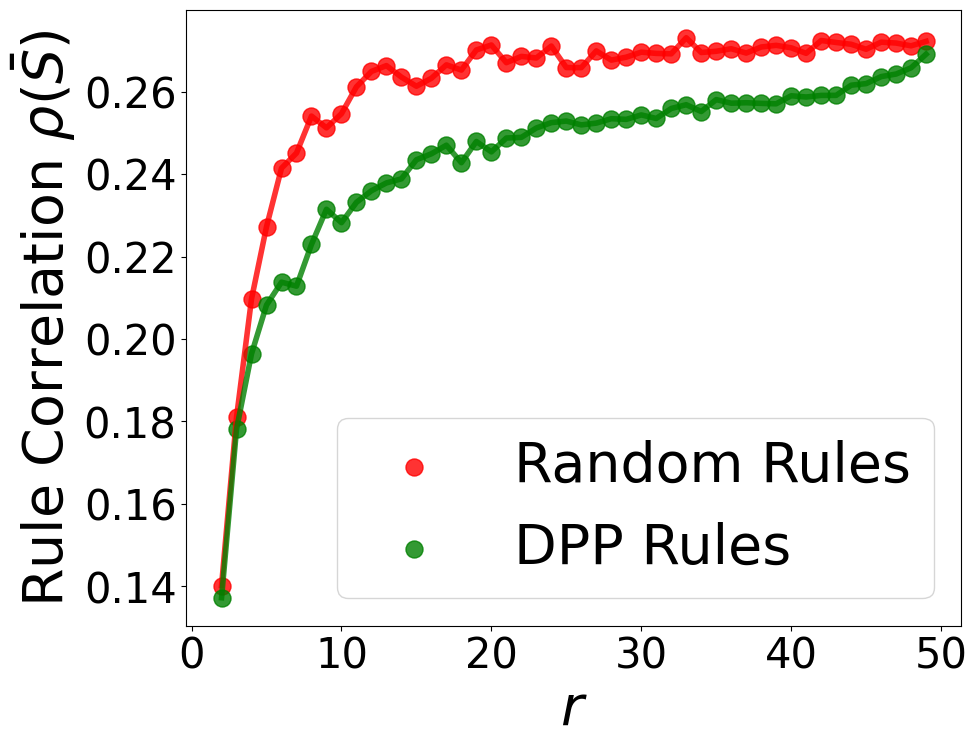}
    \caption{}
\end{subfigure}
\begin{subfigure}[b]{0.19\linewidth}
    \centering
    \includegraphics[width=\linewidth]{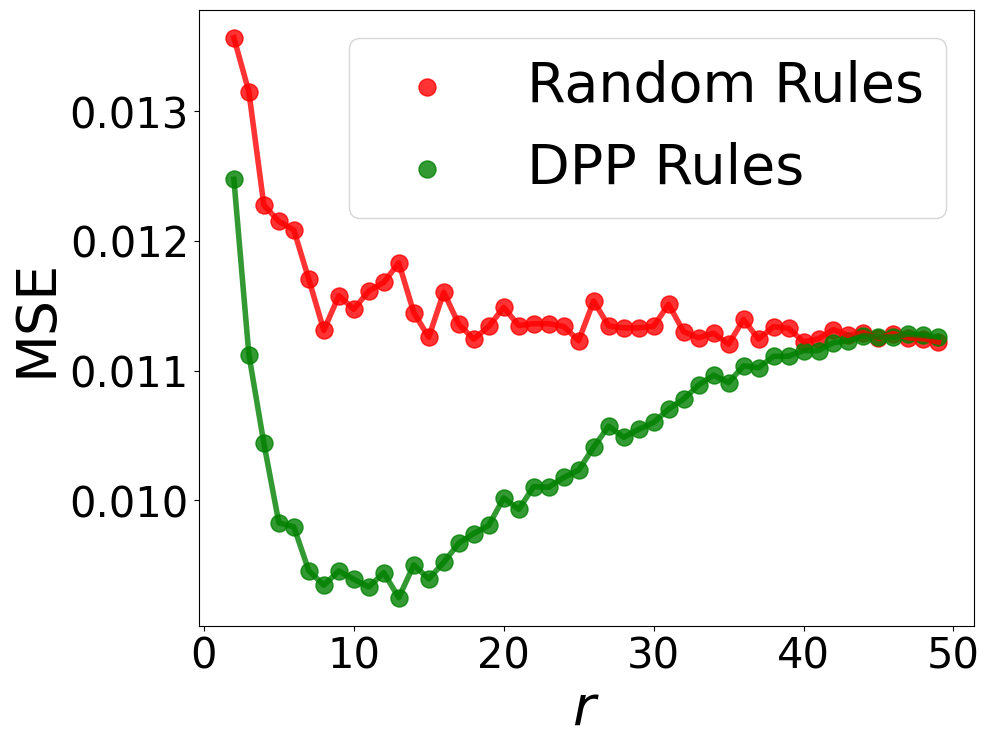}
    \caption{}
\end{subfigure}

\vspace{0.5em}

\begin{subfigure}[b]{0.19\linewidth}
    \centering
    \includegraphics[width=\linewidth]{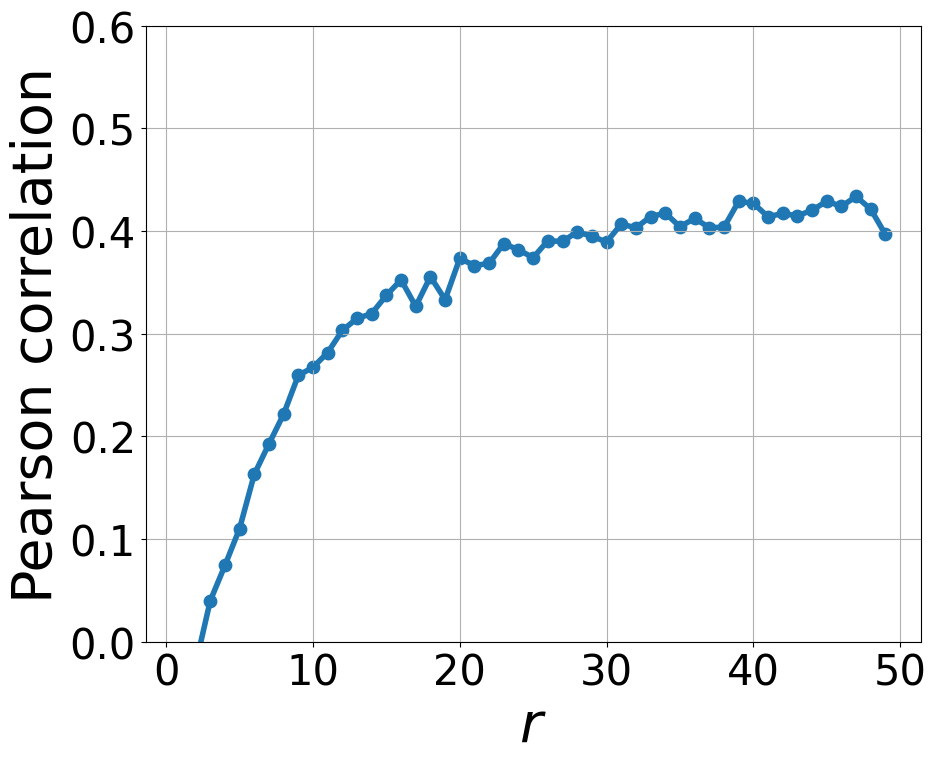}
    \caption{}
\end{subfigure}
\begin{subfigure}[b]{0.19\linewidth}
    \centering
    \includegraphics[width=\linewidth]{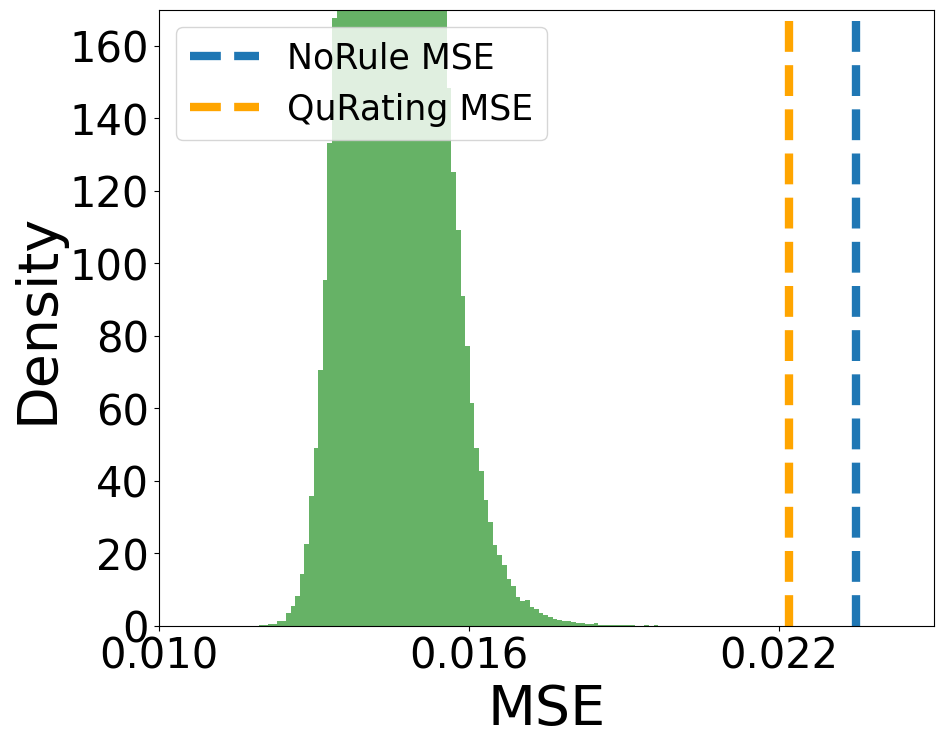}
    \caption{}
\end{subfigure}
\begin{subfigure}[b]{0.19\linewidth}
    \centering
    \includegraphics[width=\linewidth]{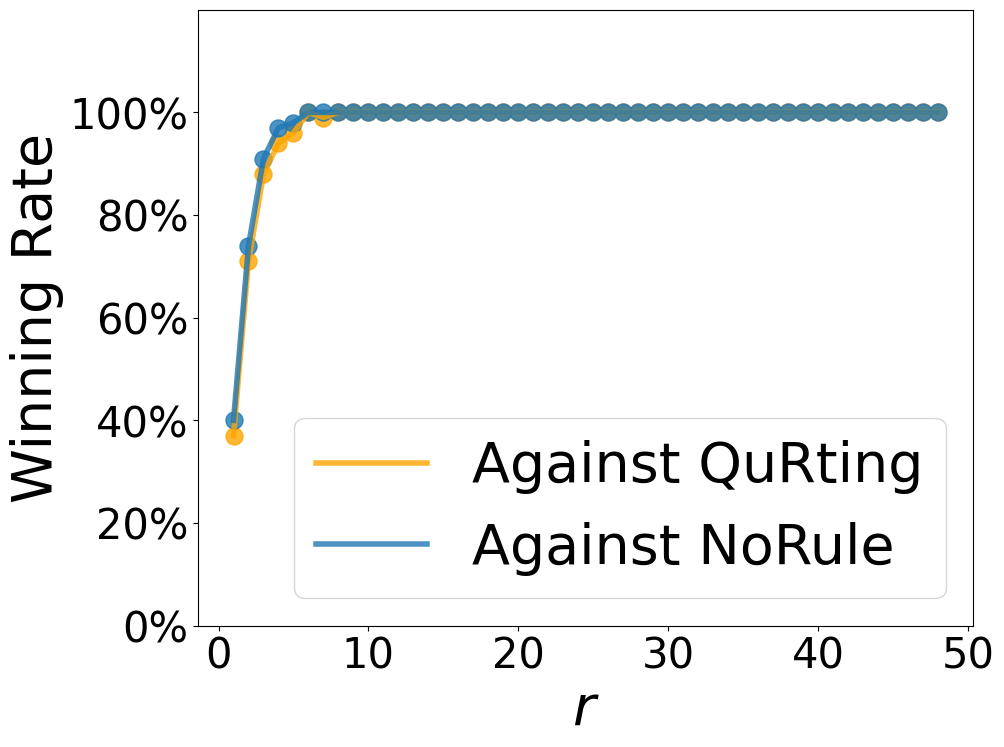}
    \caption{}
\end{subfigure}
\begin{subfigure}[b]{0.19\linewidth}
    \centering
    \includegraphics[width=\linewidth]{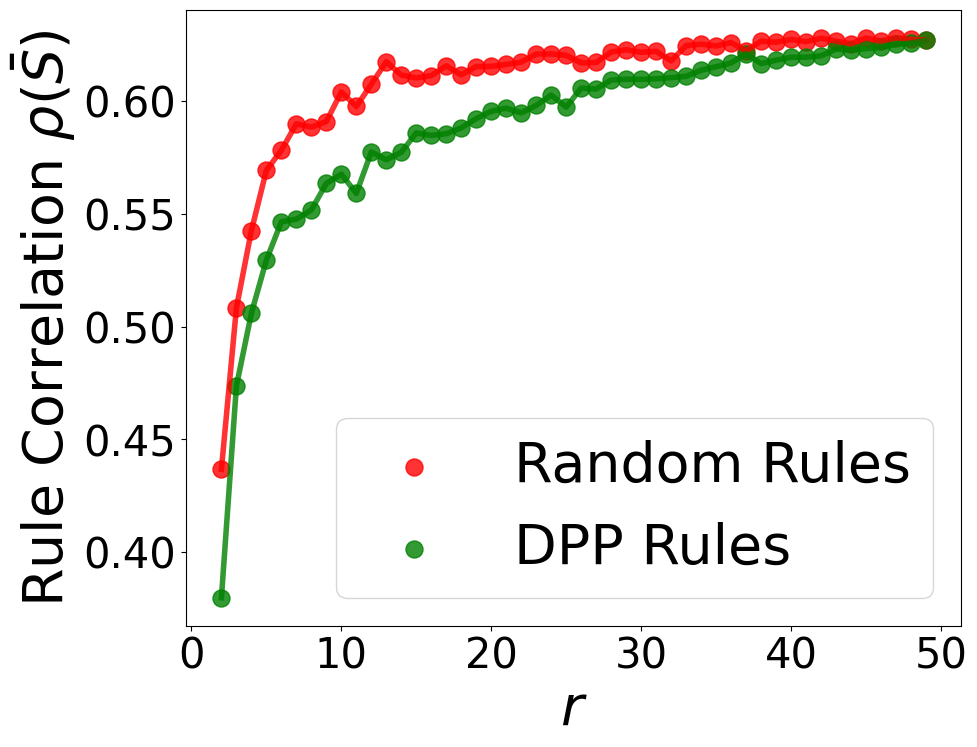}
    \caption{}
\end{subfigure}
\begin{subfigure}[b]{0.19\linewidth}
    \centering
    \includegraphics[width=\linewidth]{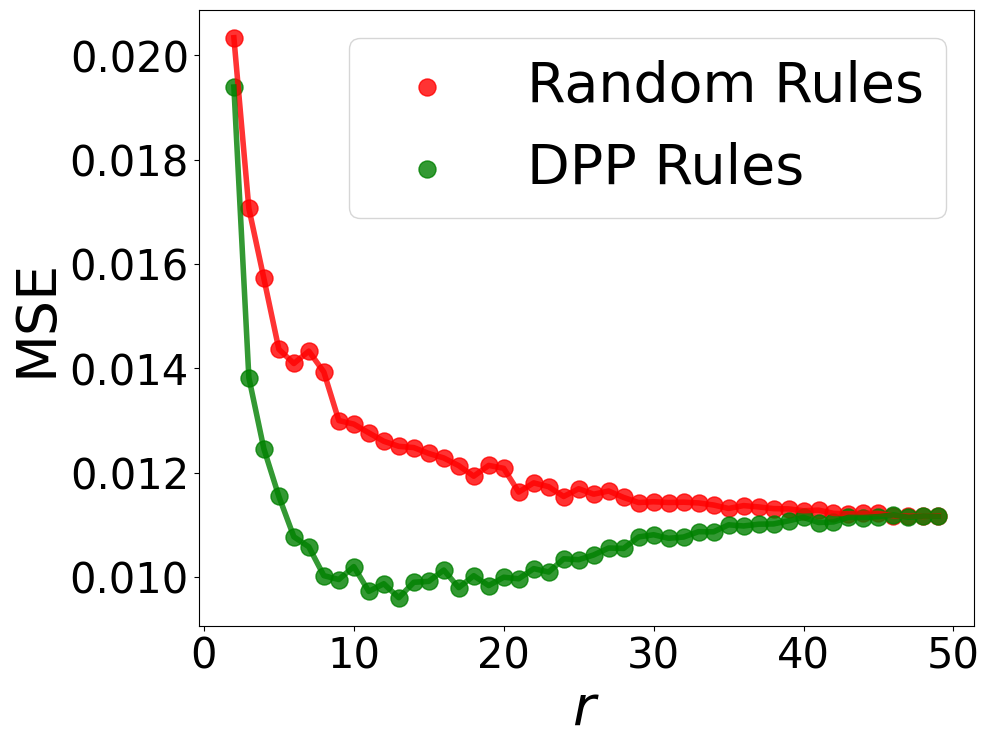}
    \caption{}
\end{subfigure}

\vspace{0.5em}

\begin{subfigure}[b]{0.19\linewidth}
    \centering
    \includegraphics[width=\linewidth]{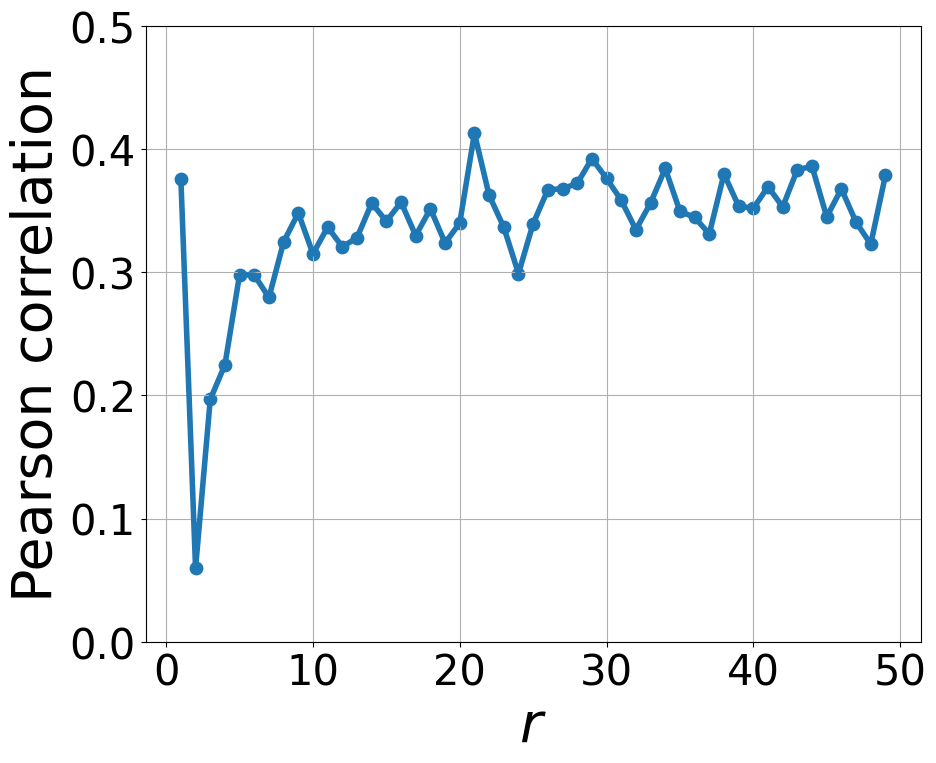}
    \caption{}
\end{subfigure}
\begin{subfigure}[b]{0.19\linewidth}
    \centering
    \includegraphics[width=\linewidth]{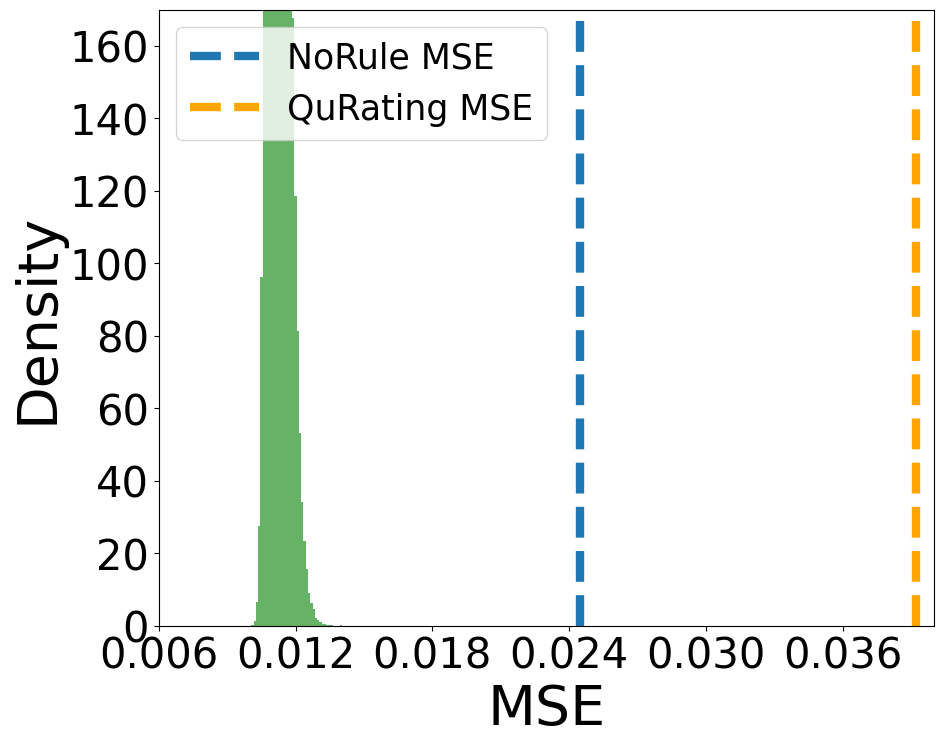}
    \caption{}
\end{subfigure}
\begin{subfigure}[b]{0.19\linewidth}
    \centering
    \includegraphics[width=\linewidth]{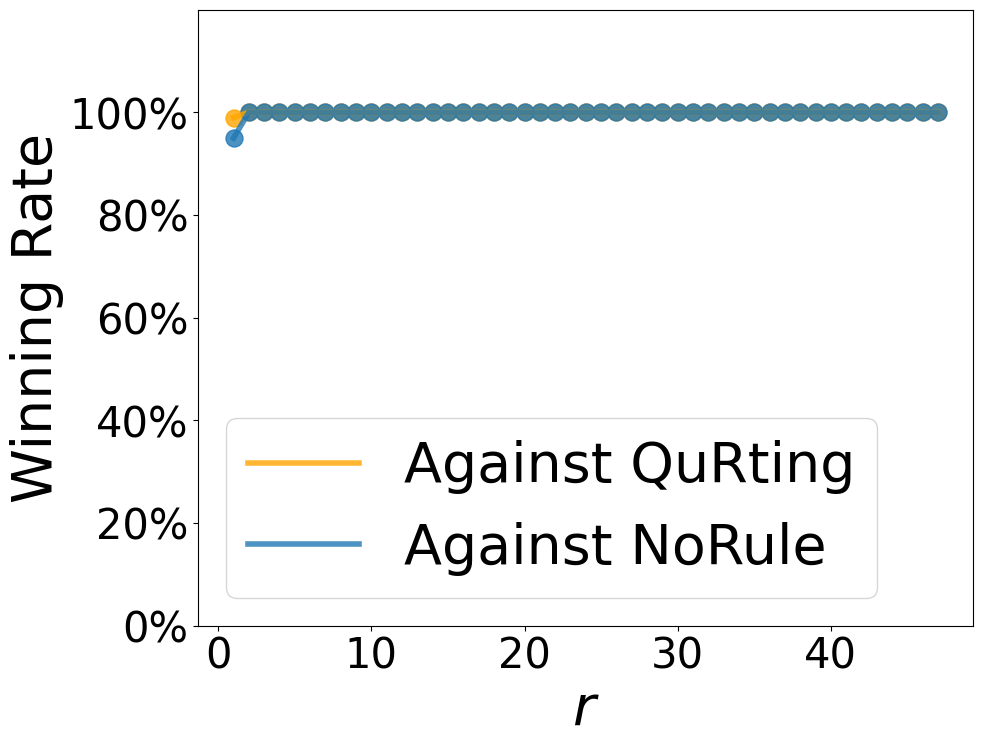}
    \caption{}
\end{subfigure}
\begin{subfigure}[b]{0.19\linewidth}
    \centering
    \includegraphics[width=\linewidth]{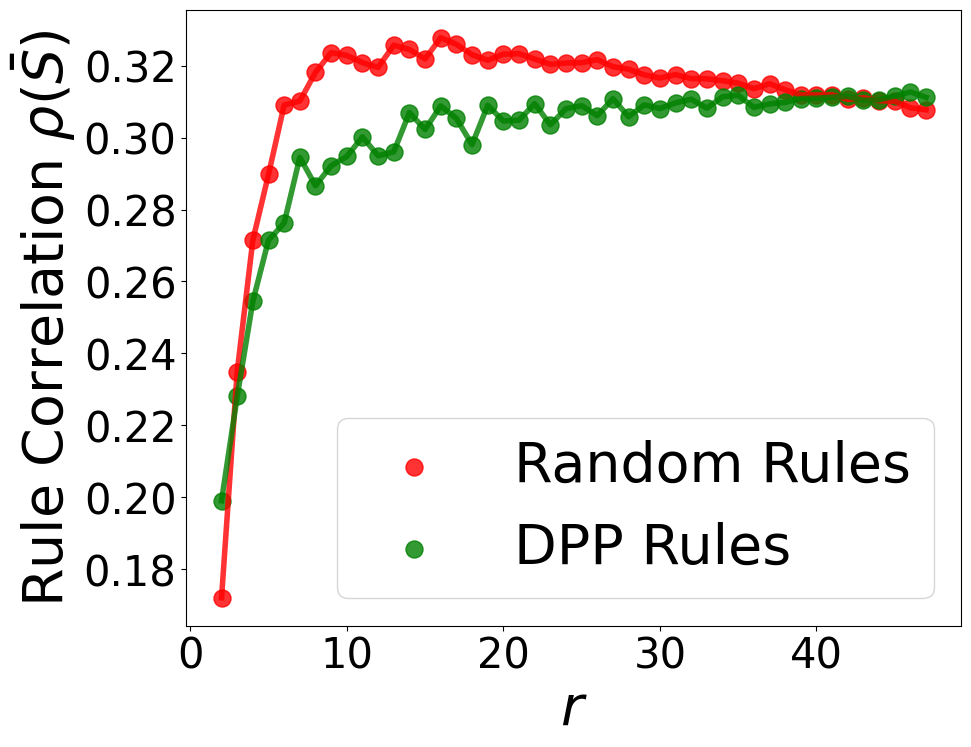}
    \caption{}
\end{subfigure}
\begin{subfigure}[b]{0.19\linewidth}
    \centering
    \includegraphics[width=\linewidth]{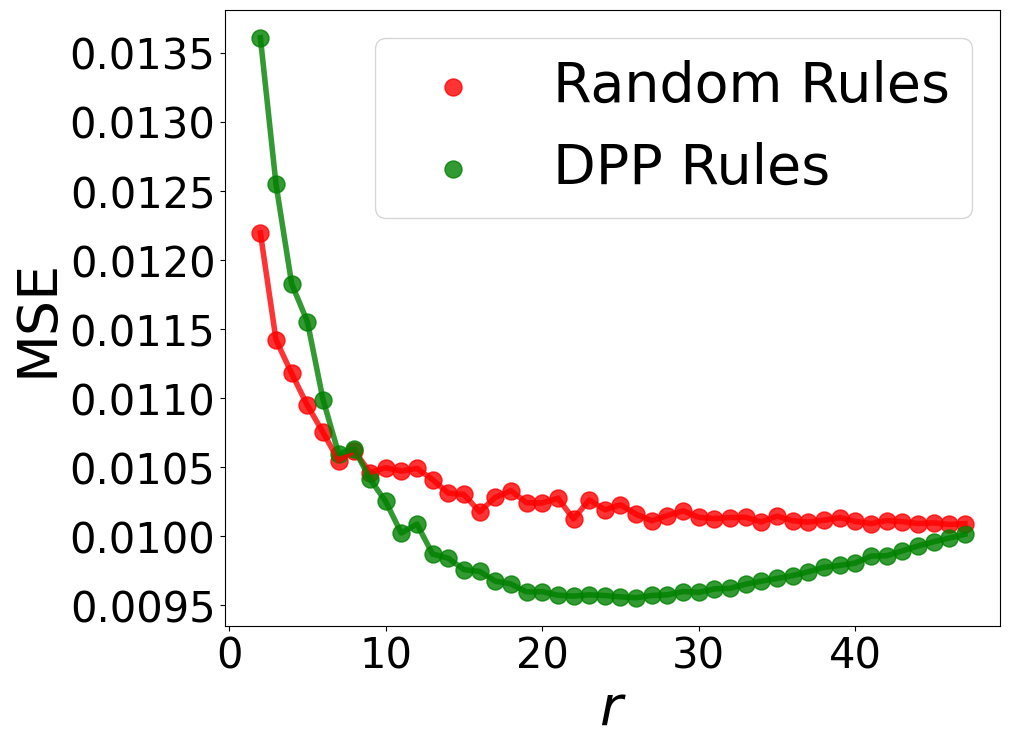}
    \caption{}
\end{subfigure}

\caption{
Each row corresponds to a domain: (top) Medical, (middle) Math, and (bottom) Code. 
(a): Pearson correlation between rule correlation $\rho(\bmS)$ and MSE $\epsilon(\bmS)$. 
(b): Distribution of MSE from $10^6$ possible rule subsets with size $r$, where vertical lines represent QuRating and NoRule. 
(c): Winning rate of DPP-selected rule subsets compared to QuRating and NoRule baselines based on MSE. 
(d,e): Comparison of DPP-selected versus randomly selected rules on rule correlation $\rho(\bmS)$ and MSE $\epsilon(\bmS)$, respectively.
}
\label{fig:human_eval_summary_three_domains}
\end{figure*}

In our preliminary experiments (Section~\ref{sec:EvaluationA}), we focused on the IMDB dataset to illustrate the effectiveness of our method for rule-based rating. Here, we extend the same Evaluation A analysis to three additional domains: Medical, Math, and Code. The corresponding results are shown in Figure~\ref{fig:human_eval_summary_three_domains}. These include the correlation between rule diversity and rating accuracy, comparisons between rule-based and rule-free settings, and evaluations of DPP-selected rules versus human-designed and randomly selected ones. The findings further confirm the generalizability of our method across diverse domains.

\subsection{General Dataset: CommonCrawl}\label{subsec:Appendix-EvalA-CommonCrawl}
\begin{figure*}[ht]
\centering

\begin{subfigure}[b]{0.19\linewidth}
    \centering
    \includegraphics[width=\linewidth]{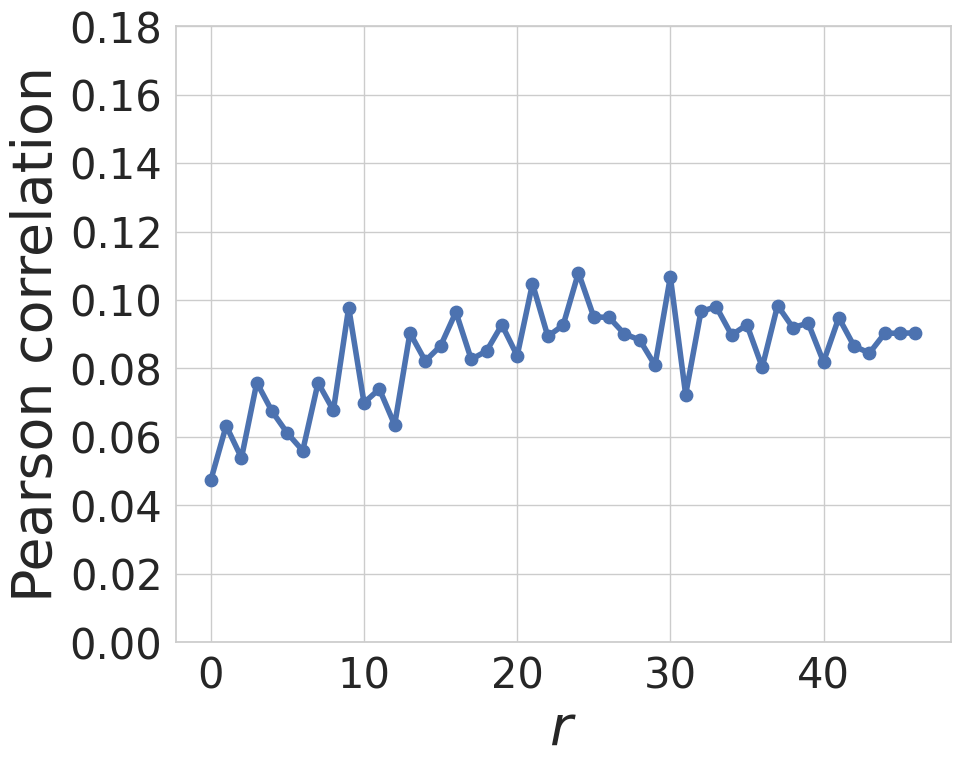}
    \caption{}
    \label{fig:EvalA_CommonCrawl_Single_pearson}
\end{subfigure}
\begin{subfigure}[b]{0.19\linewidth}
    \centering
    \includegraphics[width=\linewidth]{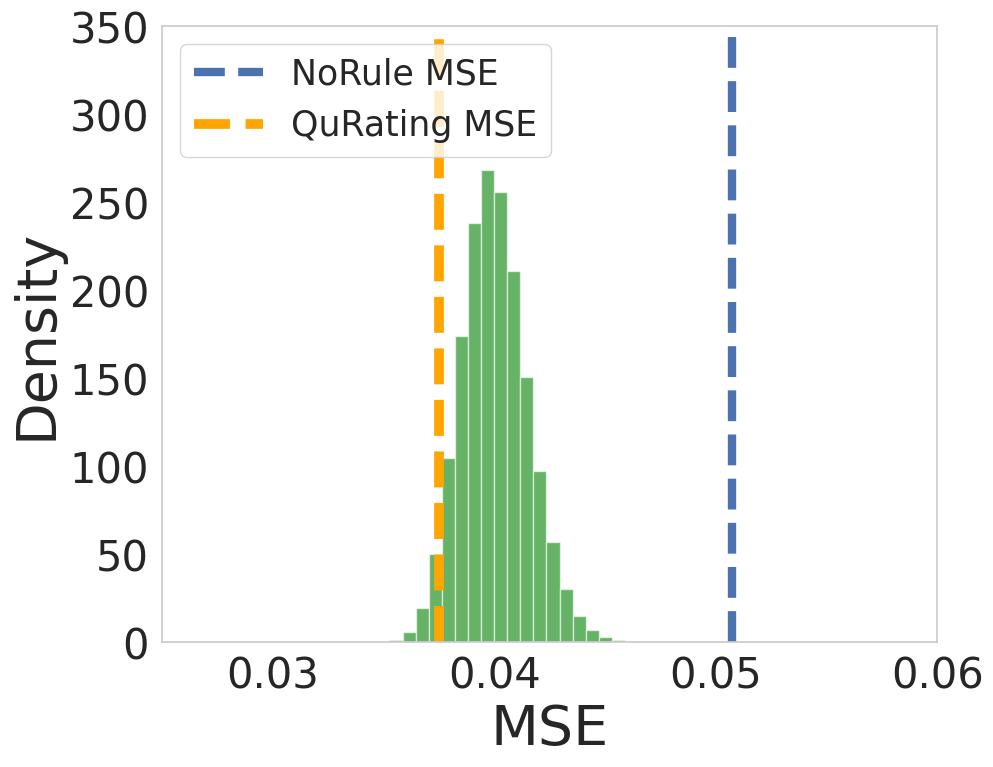}
    \caption{}
    \label{fig:EvalA_CommonCrawl_Single_histogram}
\end{subfigure}
\begin{subfigure}[b]{0.19\linewidth}
    \centering
    \includegraphics[width=\linewidth]{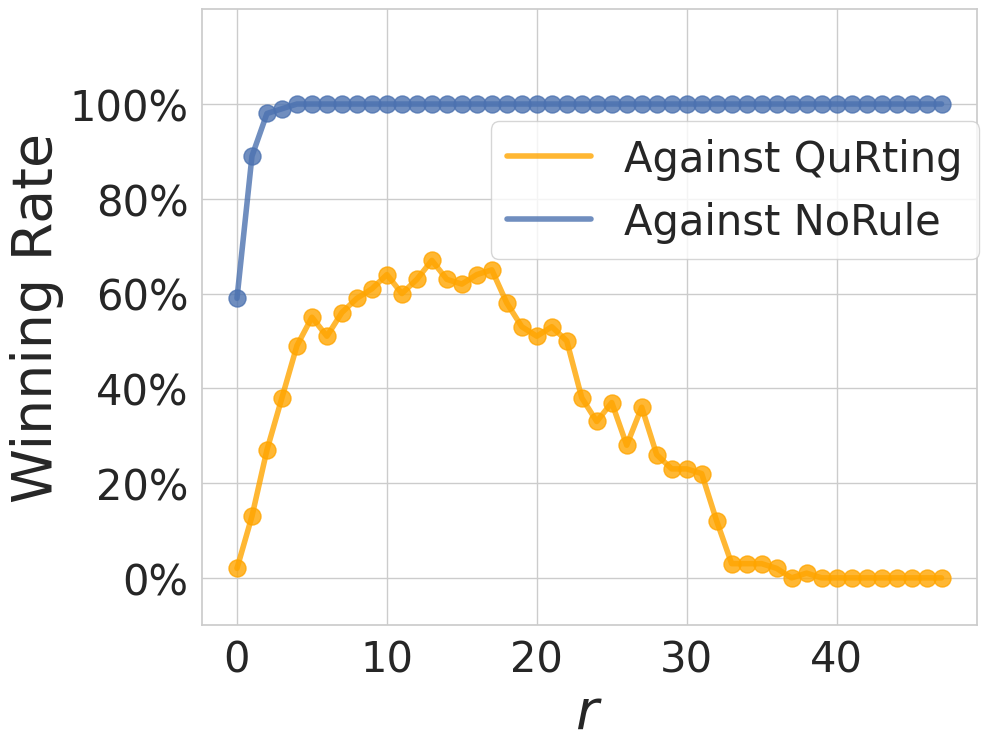}
    \caption{}
    \label{fig:EvalA_CommonCrawl_Single_winning_rate}
\end{subfigure}
\begin{subfigure}[b]{0.19\linewidth}
    \centering
    \includegraphics[width=\linewidth]{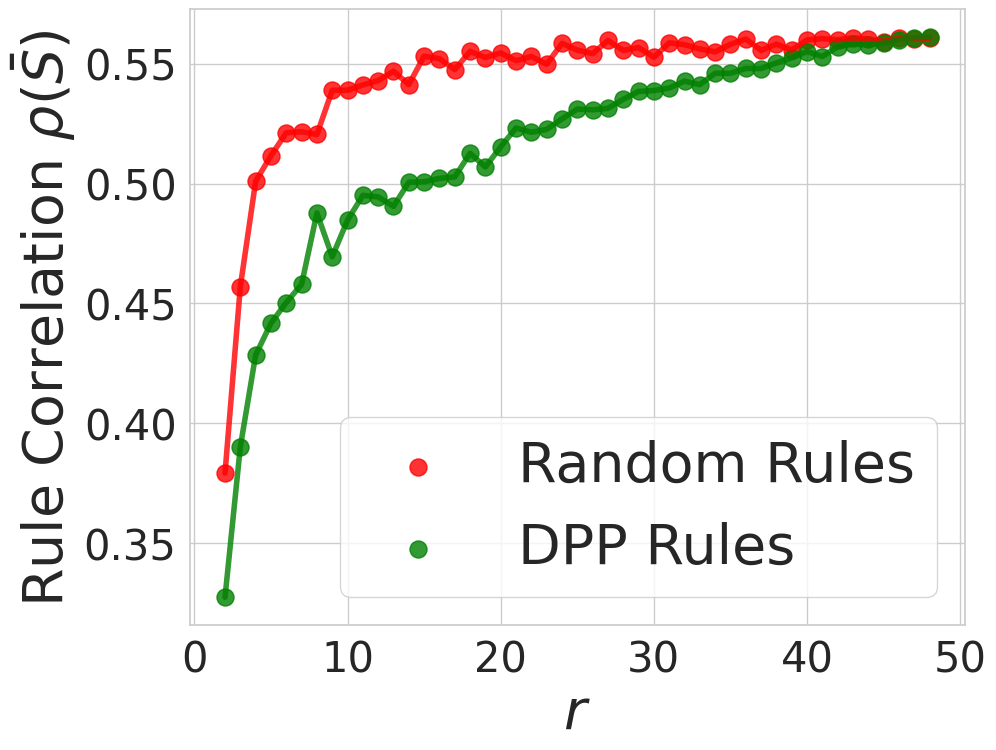}
    \caption{}
    \label{fig:EvalA_CommonCrawl_Single_dpp_vs_random_RC}
\end{subfigure}
\begin{subfigure}[b]{0.19\linewidth}
    \centering
    \includegraphics[width=\linewidth]{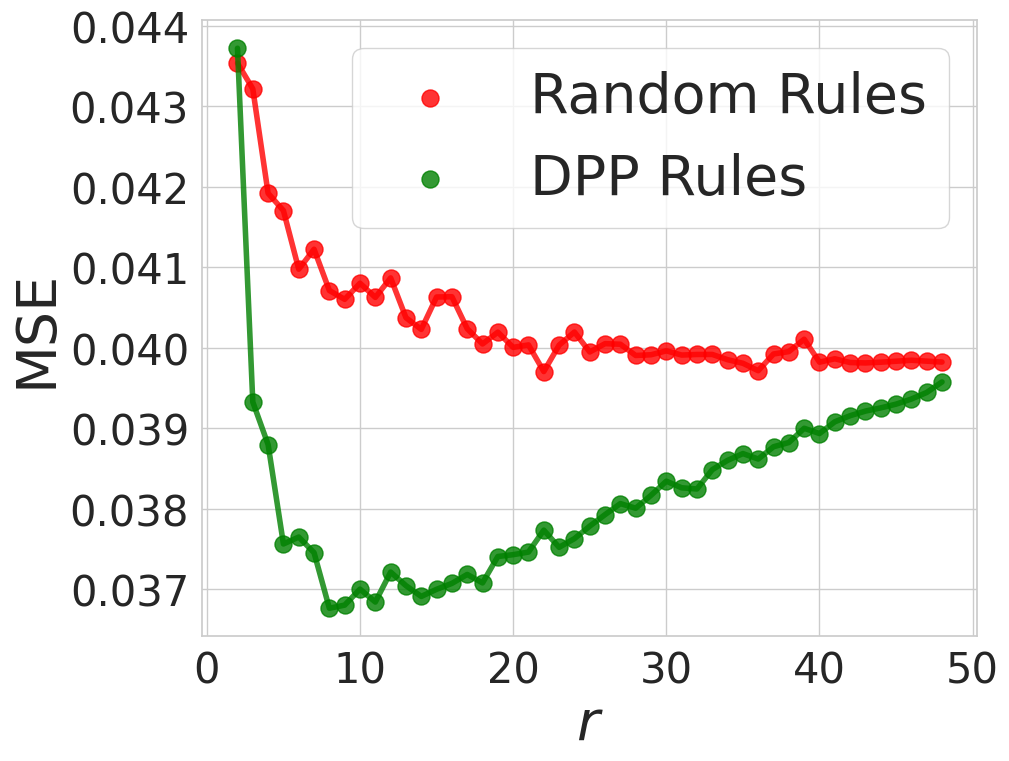}
    \caption{}
    \label{fig:EvalA_CommonCrawl_Single_dpp_vs_random_MSE}
\end{subfigure}

\caption{
Evaluation on the CommonCrawl dataset. 
(a): Pearson correlation between rule correlation $\rho(\bmS)$ and MSE $\epsilon(\bmS)$. 
(b): Distribution of MSE across $10^6$ random rule subsets with size $r$, with vertical lines indicating MSE values of QuRating and NoRule. 
(c): Winning rate of DPP-selected rule subsets compared to QuRating and NoRule. 
(d,e): Comparison of DPP-selected versus randomly selected rules on rule correlation $\rho(\bmS)$ and MSE $\epsilon(\bmS)$, averaged over 100 trials.
}
\label{fig:EvalA_CommonCrawl_All}
\end{figure*}

In Figure~\ref{fig:EvalA_CommonCrawl_All}, we observe that QuRating rules here have better performance than on IMDB dataset: it can at least surpass most of the randomly selected rules. This is reasonable since the QuRating themselves are general high-level rules. It suggests that while QuRating rules are effective for general pre-training data, they lack the flexibility to adapt to other settings or domains. When compared to DPP, interestingly, DPP underperforms QuRating when $r$ is too small or too large. 

\subsection{Empirical Distribution of $\rho$, $\epsilon$, and Ground Truth Ratings}\label{subsec:Appendix-EvalA-EmpiricalDistribution}

\begin{figure*}[t!]
\centering

\includegraphics[width=0.7\textwidth]{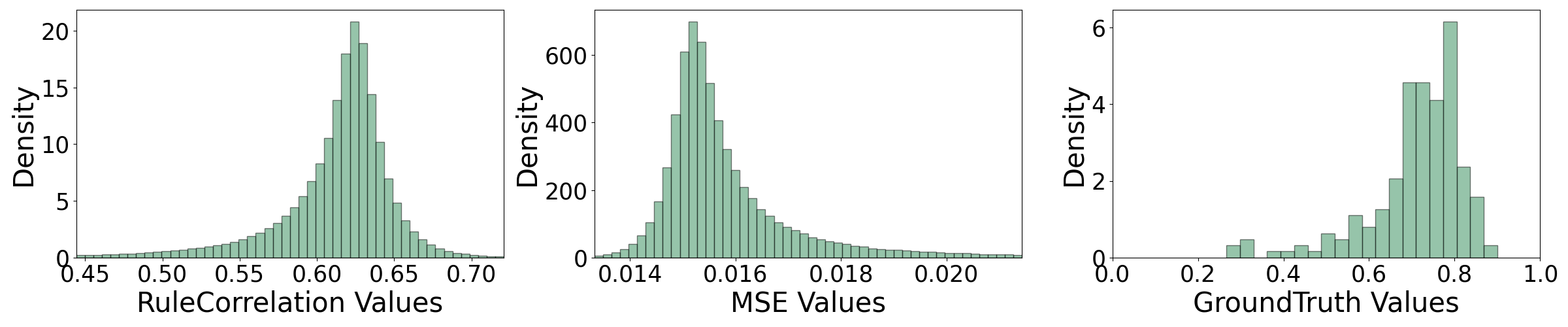}
\vspace{1em}

\includegraphics[width=0.7\textwidth]{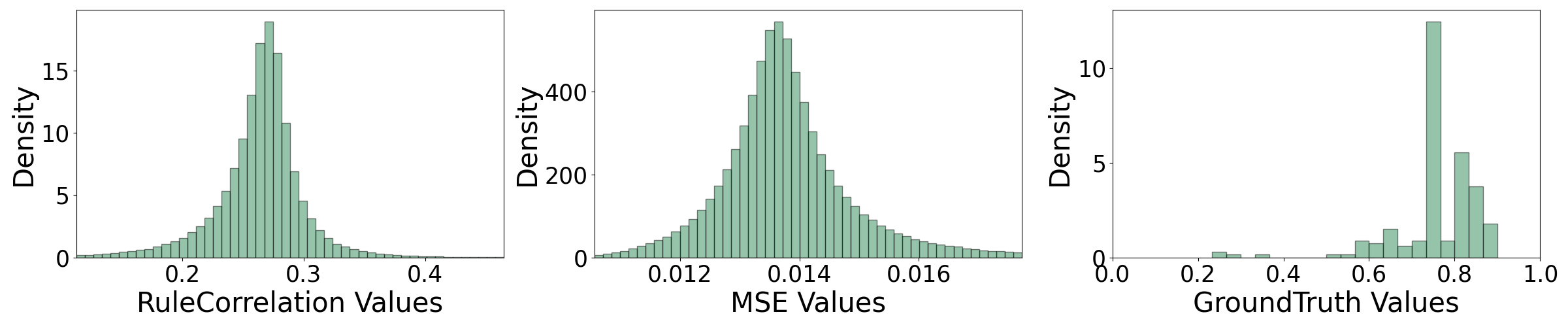}
\vspace{1em}

\includegraphics[width=0.7\textwidth]{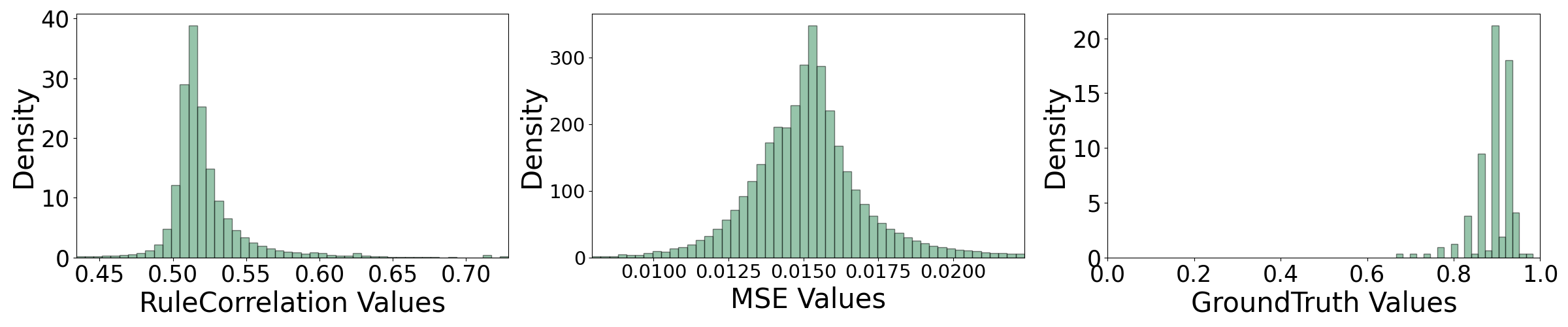}
\vspace{1em}

\includegraphics[width=0.7\textwidth]{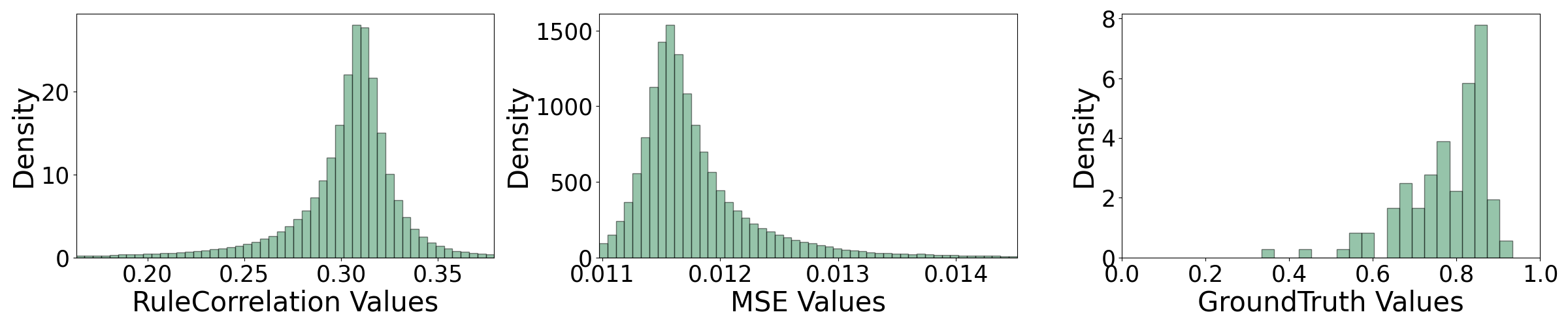}

\caption{Distributions of rule correlation $\rho(\bmS)$, MSE $\epsilon(\bmS)$, and ground truth scores across four datasets: IMDB, Medical, Math, and Code.}
\label{fig:distribution_rho_mse_gt_all}
\end{figure*}

To verify that the rule correlation $\rho(\bmS)$, the MSE $\epsilon(\bmS)$, and the ground truth ratings $S_{GT}$ are not trivially concentrated around a narrow range or even single value, we plot their empirical distributions across different domains. As shown in Figure~\ref{fig:distribution_rho_mse_gt_all}, each metric exhibits substantial variance, indicating meaningful diversity among rule subsets and human-provided quality scores. This supports the validity of using these signals for evaluating rule selection strategies.

\subsection{Prompts and Generated Rules}\label{subsec:Appendix-EvalA-Prompts-Rules}
Below are the templates used to rate a sample according to a specific rule. For the rule-free version, simply omit the sentence involving the rule. Replace DATASET\_NAME\_AND\_DESCRIPTION with the dataset's name and one sentence of description to correspond to the two data sources discussed in Section~\ref{sec:EvaluationA}.

\begin{figure}[H]
\centering
\small
\begin{tcolorbox}[colback=cyan!10!white, 
                  colframe=cyan!30!white, 
                  width=0.50\textwidth, 
                  arc=4mm, 
                  auto outer arc,
                  ]

Evaluate the following example from \texttt{\color{red}<DATASET\_NAME\_AND\_DESCRIPTION>} and assign a quality score between 0 and 1 (0 indicates the worst quality, and 1 indicates perfect quality) according to the provided rule:\\
\texttt{\color{red}<RULE>}\\
\\
Example:\\
\texttt{\color{red}<DATA\_SAMPLE>}\\\\
Respond only with a single float number.
\end{tcolorbox}
\caption{Template of rule-based comparison prompt.}
\label{tab:EvalA-rule_individual_prompt}
\end{figure}






The rules are then manually checked by 5 human annotations (all five are PhD students in CS or Math) to make sure the meaning is clear and not too abstract, and to remove the rules that are completely unrelated to the task/domain.

\textbf{Generated IMDB rules:}

\begin{table*}[t]
\small 
\caption{Generated rules for the IMDB examples.} \label{tab:rules_for_imdb}
\setlength{\tabcolsep}{6pt}
\renewcommand{\arraystretch}{1.15}
\begin{tabularx}{\textwidth}{c|X}
\toprule
0  & Clearly state the main opinion or sentiment of the reviewer. \\
\hline
1  & Be free of spelling errors. \\
\hline
2  & Be free of grammatical errors. \\
\hline
3  & Have a coherent structure with a clear beginning, middle, and end. \\
\hline
4  & Be relevant to the movie being reviewed. \\
\hline
5  & Avoid using offensive or inappropriate language. \\
\hline
6  & Provide specific reasons for the given sentiment. \\
\hline
7  & Include details that support the overall sentiment. \\
\hline
8  & Be free from excessive use of exclamation marks. \\
\hline
9  & Not contain any personal attacks on individuals. \\
\hline
10 & Not be overly repetitive. \\
\hline
11 & Avoid vague statements and provide concrete examples. \\
\hline
12 & Not include spoilers without a spoiler warning. \\
\hline
13 & Be written in complete sentences. \\
\hline
14 & Not use excessive capitalization for emphasis. \\
\hline
15 & Have a logical flow and avoid jumping between unrelated points. \\
\hline
16 & Not contain any irrelevant information. \\
\hline
17 & Be at least 100 words long. \\
\hline
18 & Not exceed 500 words. \\
\hline
19 & Not contain any text that is irrelevant to the movie review. \\
\hline
20 & Not include any links or advertisements. \\
\hline
21 & Provide a balanced perspective, mentioning both positives and negatives if applicable. \\
\hline
22 & Not be biased or prejudiced. \\
\hline
23 & Be written from a first-person perspective. \\
\hline
24 & Not contain any misleading information. \\
\hline
25 & Mention the movie title at least once. \\
\hline
26 & Be engaging and hold the reader's attention. \\
\hline
27 & Avoid overly technical language that might confuse readers. \\
\hline
28 & Not be a duplicate of another review in the dataset. \\
\hline
29 & Mention specific scenes or elements of the movie when providing critiques. \\
\hline
30 & Provide a final summary of the reviewer's overall opinion. \\
\hline
31 & Not include excessive punctuation marks such as multiple question marks or exclamation points. \\
\hline
32 & Not use text abbreviations or slang. \\
\hline
33 & Be written in a formal or semi-formal tone. \\
\hline
34 & Provide context for any cultural or historical references. \\
\hline
35 & Not make unsupported generalizations. \\
\hline
36 & Maintain a consistent tone throughout. \\
\hline
37 & Not contradict itself. \\
\hline
38 & Indicate whether the reviewer recommends the movie or not. \\
\hline
39 & Not contain any unnecessary filler words or phrases. \\
\hline
40 & Be respectful and considerate in its critique. \\
\hline
41 & Address the acting, direction, and cinematography if possible. \\
\hline
42 & Be free from any copy-pasted text from other sources. \\
\hline
43 & Not include any personal anecdotes unrelated to the movie. \\
\hline
44 & Be specific about what worked and what didn't in the movie. \\
\hline
45 & Mention the genre of the movie. \\
\hline
46 & Include a rating or score if available. \\
\hline
47 & Be written with the target audience in mind. \\
\hline
48 & Provide insights into the movie's themes and messages. \\
\hline
49 & Not contain any text that is purely promotional in nature. \\
\hline
\bottomrule
\end{tabularx}
\end{table*}

\textbf{Generated CommonCrawl rules:}

\begin{table*}[t]
\small 
\caption{Generated 50 Rules for rating CommonCrawl examples.} 
\setlength{\tabcolsep}{6pt}
\renewcommand{\arraystretch}{1.15}
\begin{tabularx}{\textwidth}{c|X}
\toprule
0 & The text should be free of spelling errors. \\
\hline
1 & The grammar should be correct and appropriate for the context. \\
\hline
2 & The content should be relevant to the topic described in the title or metadata. \\
\hline
3 & The text should not contain any offensive or inappropriate language. \\
\hline
4 & The information presented should be factually accurate. \\
\hline
5 & The text should not be overly repetitive. \\
\hline
6 & The sentences should be clear and concise. \\
\hline
7 & The text should provide useful and meaningful information. \\
\hline
8 & The content should be engaging and interesting to the reader. \\
\hline
9 & The text should have a logical flow and coherent structure. \\
\hline
10 & The text should not contain broken or incomplete sentences. \\
\hline
11 & The metadata should accurately reflect the content of the text. \\
\hline
12 & The text should not include excessive jargon or overly complex language. \\
\hline
13 & The content should be relevant to the intended audience. \\
\hline
14 & The text should be free of any advertisements or promotional material. \\
\hline
15 & The text should not contain any personal information or sensitive data. \\
\hline
16 & The content should be original and not plagiarized. \\
\hline
17 & The text should not include any irrelevant or off-topic information. \\
\hline
18 & The text should be formatted properly with appropriate headings and paragraphs. \\
\hline
19 & The text should be free of any links or URLs unless relevant and necessary. \\
\hline
20 & The content should be up-to-date and not outdated. \\
\hline
21 & The text should be free of any empty or meaningless filler words. \\
\hline
22 & The content should provide a balanced and unbiased perspective. \\
\hline
23 & The text should not contain any images or multimedia unless relevant and properly embedded. \\
\hline
24 & The text should have proper punctuation marks. \\
\hline
25 & The text should not contain any placeholders or unfinished sentences. \\
\hline
26 & The text should be suitable for the language model's training purposes. \\
\hline
27 & The text should maintain a consistent tone and style throughout. \\
\hline
28 & The text should not contain any HTML or other markup language unless specified. \\
\hline
29 & The text should avoid slang or colloquial expressions unless contextually appropriate. \\
\hline
30 & The content should have proper citations or references if necessary. \\
\hline
31 & The text should be sufficiently detailed to provide value to the reader. \\
\hline
32 & The text should be free of any biased or prejudiced language. \\
\hline
33 & The text should not contain any technical errors or glitches. \\
\hline
34 & The content should have a clear beginning, middle, and end. \\
\hline
35 & The text should be free of any redundant phrases or statements. \\
\hline
36 & The text should adhere to any specified length requirements. \\
\hline
37 & The text should not contain any duplicate content. \\
\hline
38 & The text should be relevant to the specified geographic location if mentioned. \\
\hline
39 & The text should not include any speculative or unverified information. \\
\hline
40 & The content should encourage reader engagement and interaction. \\
\hline
41 & The text should maintain a professional tone unless otherwise specified. \\
\hline
42 & The text should be free of any ambiguities or unclear statements. \\
\hline
43 & The content should not promote any illegal activities or behaviors. \\
\hline
44 & The text should have a neutral point of view unless otherwise specified. \\
\hline
45 & The text should be free of any distracting formatting errors. \\
\hline
46 & The content should address any specified keywords or topics effectively. \\
\hline
47 & The text should be free of any content that violates copyright or intellectual property rights. \\
\hline
48 & The text should have a clear and relevant title or headline. \\
\hline
49 & The text should be suitable for training language models for general downstream tasks. \\
\bottomrule
\end{tabularx}
\end{table*}

\section{Appendix for Main Experiments}\label{sec:Appendix-EvalB}

\subsection{Model training}\label{subsec:Appendix-EvalB-model_training}
For training Pythia-1B and Llama3-8B, we loaded both models using \texttt{bfloat16} precision and used one \texttt{NVIDIA A100-80GB} for each training job. Below are the training parameters:
\begin{table}[H]
\centering
\mytiny
\caption{Comparison of Model Parameters}
\label{tab:Appendix-EvalB-training}
\begin{tabular}{l|c|c}  
\hline  
\textbf{Model}      & Pythia-1B & Llama3-8B \\
\hline  
\textbf{Num of epochs}           & 1                  & 1                  \\
\hline  
\textbf{Learning rate}           & $2\cdot 10^{-5}$   & $2\cdot 10^{-5}$   \\
\hline
\textbf{Token max length}        & 2048               & 4096               \\
\hline
\textbf{LoRA}                    & No                 & Rank=64      \\
\hline  
\end{tabular}
\end{table}


\subsection{GPT 10 uncorrelated rules}
Another straightforward rule generation method is to directly prompt GPT-4 to generate 10 uncorrelated rules and rely on its understanding of the correlation between the rules. We have explored this by using a similar rule generation prompt as in \ref{subsec:Appendix-EvalB-Prompt-Rules}, where we provide the task description and data description, but this time we request for 10 rules and added one sentence ``make sure the rules are uncorrelated'' to further require the independence of the rules. The 10 ``uncorrelated'' GPT rules are provided in Table~\ref{tab:Appendix-EvalB-GPT10UncorrelatedRules-Code} and \ref{tab:Appendix-EvalB-GPT10UncorrelatedRules-Math}  below. Following this, we rated the data according to the 10 GPT rules and calculated the rule correlation $\rho$ of the score vectors (in one \textit{GPT-Uncorrelated} trial). We tested for Code and Math domain and got $\rho_{Code} = 0.65$ and $\rho_{Math}=0.56$, both are significantly higher than DPP correlation values in Tabel~\ref{tab:Appendix-EvalB-RuleIndices}. For Code, even random 10 rules selected from a pool of 50 rules provide lower correlation than the 10 rules directly generated by GPT that are claimed to be ``uncorrelated''. This shows that our two-step approach---first generating enough rules to ensure diversity, followed by employing DPP on the rating vectors to select rules---is superior and also more task-specific.

\begin{table*}[t]
\small 
\caption{Generated 10 “uncorrelated” rules by GPT-4 for the Code domain.}\label{tab:Appendix-EvalB-GPT10UncorrelatedRules-Code}
\setlength{\tabcolsep}{6pt}
\renewcommand{\arraystretch}{1.15}
\begin{tabularx}{\textwidth}{c|X}
\toprule
\textbf{Index} & \textbf{Rule Description} \\
\midrule
0 & Code Snippet Integrity: Select examples that contain complete and syntactically correct code snippets, avoiding those with partial or pseudo code which may confuse the model. \\
\hline
1 & Language Diversity: Include examples in a variety of programming languages, ensuring that no single language dominates the dataset to promote versatility in code generation. \\
\hline
2 & Comment Quality: Prioritize data that includes well-documented code with comments that clearly explain the logic and functionality. \\
\hline
3 & Algorithmic Complexity: Choose examples that demonstrate a range of algorithmic solutions from basic to advanced. \\
\hline
4 & Relevance to Modern Programming: Favor examples that utilize current libraries, frameworks, and features of programming languages. \\
\hline
5 & Balanced Domain Representation: Ensure a balanced representation of code from different domains to prevent model bias. \\
\hline
6 & Error Handling: Include examples that demonstrate robust error handling and debugging practices. \\
\hline
7 & Executable Code: Select training examples where the code is functional and executable without errors. \\
\hline
8 & Contextual Coherence: Ensure that the selected texts provide meaningful context that relates logically to the code. \\
\hline
9 & Code Formatting and Style: Include examples that adhere to common coding standards and formats. \\
\bottomrule
\end{tabularx}
\end{table*}

\begin{table*}[t]
\small 
\caption{Generated 10 “uncorrelated” rules by GPT-4 for the Math domain.}\label{tab:Appendix-EvalB-GPT10UncorrelatedRules-Math}
\setlength{\tabcolsep}{6pt}
\renewcommand{\arraystretch}{1.15}
\begin{tabularx}{\textwidth}{c|X}
\toprule
\textbf{Index} & \textbf{Rule Description} \\
\midrule
0 & Lexical Diversity Rule: Select data samples with a diverse vocabulary, especially those rich in mathematical terminology. Exclude texts with high repetition of common words and low occurrence of domain-specific terms. \\
\hline
1 & Complexity and Structure Rule: Prioritize texts that exhibit complex sentence structures and logical argumentation, indicative of advanced reasoning skills. \\
\hline
2 & Numerical Data Presence Rule: Include texts that contain numerical data, charts, or graphs, along with explanatory text that interprets or analyzes the numerical information. \\
\hline
3 & Mathematical Concept Explanation Rule: Favor texts that explicitly explain mathematical concepts, theories, or problem-solving steps. \\
\hline
4 & Contextual Relevance Rule: Select texts related to mathematical applications in real-world scenarios, such as physics problems or economics calculations. \\
\hline
5 & Historical and Evolutionary Math Content Rule: Include content discussing the historical development and evolution of mathematical theories. \\
\hline
6 & Cross-Disciplinary Integration Rule: Opt for texts that integrate mathematical concepts with other disciplines like science and engineering. \\
\hline
7 & Error-free Mathematical Notation Rule: Ensure that the texts contain accurate and error-free mathematical notation wherever applicable. \\
\hline
8 & Problem-Solving Narratives Rule: Select texts that include step-by-step problem-solving narratives or worked examples. \\
\hline
9 & Cultural and Application Diversity Rule: Include texts that discuss the application of mathematics in various cultural and practical contexts. \\
\bottomrule
\end{tabularx}
\end{table*}

\subsection{Variance of Trials}\label{subsec:Appendix-EvalB-Trials}
Due to computational resource constraints, we were unable to perform multiple repetitions of all experiments. However, as mentioned in Section~\ref{sec:EvaluationB}, we conducted 3 independent trials in four domains for \textit{Uniform Sampling} and methods involving randomness in rule selections, including \textit{GPT-Uncorrelated}, \textit{Random 10 Rules}, and \textit{DPP 10 Rules} (note that DPP sampling is also non-deterministic) to mitigate the effects of randomness, and we report their standard deviations here in Tables~\ref{tab:IMDB_Medical_STD} and \ref{tab:Math_Code_STD}


\begin{table*}[t!]
\centering
\small
\setlength{\tabcolsep}{4pt}
\renewcommand{\arraystretch}{1.2}
\caption{Mean and standard deviation over 3 independent trials for the IMDB \& Medical fine-tuning setting.}
\label{tab:IMDB_Medical_STD}
\resizebox{0.8\textwidth}{!}{%
    \begin{tabular}{ c | c | c  c  c  c} \hline 
           \multirow{2}*{Method} & IMDB & \multicolumn{4}{c}{Medical} \\
           \cline{2-6} 
           &  SA accuracy & CollegeMedicine & ProfessionalMedicine & MedicalGenetics & MedicalAverage\\ \hline 
            Uniform Sampling & $43.9_{1.1}$ & $23.0_{0.42}$ & $42.1_{0.62}$ & $22.5_{0.71}$ & $28.9_{0.77}$\\
            GPT-Uncorrelated & $50.9_{0.71}$ & $23.7_{0.11}$ & $42_{0.3}$ & $22.7_{0.76}$  & $29.4_{0.2}$\\
            \hline
            Random 10 Rules & $51.7_{0.21}$ & $24_{0.42}$ & $41.2_{1.2}$ & $23.5_{0.41}$ & $29.6_{0.58}$\\
            DPP 10 Rules & $53.5_{0.58}$ & $24.6_{0.37}$ &  $43.3_{0.43}$ & $26.8_{0.76}$ & $31.6_{0.41}$\\ \hline
    \end{tabular}
}
\end{table*}

\begin{table*}[t!]
\centering
\small
\setlength{\tabcolsep}{4pt}
\renewcommand{\arraystretch}{1.2}
\caption{Mean and standard deviation over 3 independent trials for the Math \& Code fine-tuning setting.}
\label{tab:Math_Code_STD}
\resizebox{1.0\textwidth}{!}{%
    \begin{tabular}{ c | c  c  c c | c  c  c  c c} \hline 
           \multirow{2}*{Method} & \multicolumn{4}{c|}{Math} & \multicolumn{5}{c}{Code} \\
           \cline{2-10} 
           &  elementary& high school & college & Math Average & humaneval& mbpp & multiple-py & multiple-cpp & Code Average\\ \hline 
            Uniform Sampling & $40.5_{0.35}$ & $39.2_{0.56}$ & $35_{0.2}$ & $38.2_{0.30}$ & $38.7_{1.27}$ & $38.2_{1.6}$ & $38.2_{0.42}$ & $39.7_{1.2}$ & $38.7_{0.94}$\\
            GPT-Uncorrelated & $41.4_{0.17}$ & $39_{0.21}$ & $37.3_{0.57}$ & $39.2_{0.28}$ & $41.2_{0.15}$ & $43.5_{0.11}$ & $39.6_{1.44}$ & $48.6_{0.15}$ & $43.2_{0.32}$\\
            \hline
            Random 10 Rules & $42.9_{0.1}$ & $39.8_{1.3}$ & $35.2_{0.28}$ & $39.3_{0.48}$ & $48.5_{0.6}$ & $41_{0.94}$ & $46.6_{0.85}$ & $48.1_{1.2}$ & $46_{0.78}$\\
            DPP 10 Rules & $43.7_{0.61}$ &  $40.6_{0.32}$ &  $38_{0}$ &  $40.8_{0.15}$ &$50.5_{0.36}$ &  $44.2_{0.26}$ & $46.9_{0.2}$ & $52.7_{0.15}$ & $48.6_{0.22}$\\ \hline
    \end{tabular}
}
\end{table*}

Here in Tables~\ref{tab:method_comparison_imdb_medical} and \ref{tab:method_comparison_math_code}, we perform the $t$-test to demonstrate that the advantage of \textit{DPP 10 Rules} is significant compared to other methods. We include the $t$-statistics and $p$-values in the table. If we choose the significance threshold $p=0.05$, then we see that all the comparisons are significant.

\begin{table*}[t!]
\centering
\scriptsize
\resizebox{0.8\textwidth}{!}{%
\begin{tabular}{@{}p{3.2cm}cc@{}}
\toprule
Comparison & IMDB & Medical average \\ \midrule
DPP vs GPT-Uncorrelated & \(t=4.912\), \(p=0.00881\) & \(t=8.353\), \(p=0.00408\) \\
DPP vs Uniform Sampling & \(t=13.371\), \(p=0.00086\) & \(t=5.361\), \(p=0.01218\) \\
DPP vs Random 10 Rules  & \(t=5.054\), \(p=0.02255\) & \(t=4.877\), \(p=0.01065\) \\
\bottomrule
\end{tabular}
}
\caption{Comparison of DPP method with other methods in IMDB and Medical AVG domains using Welch's t-test.}
\label{tab:method_comparison_imdb_medical}
\end{table*}

\begin{table*}[t!]
\centering
\scriptsize
\resizebox{0.8\textwidth}{!}{%
\begin{tabular}{@{}p{3.2cm}cc@{}}
\toprule
Comparison & Math average & Code average \\ \midrule
DPP vs GPT-Uncorrelated & \(t=8.724\), \(p=0.00293\) & \(t=24.085\), \(p=0.00005\) \\
DPP vs Uniform Sampling & \(t=13.426\), \(p=0.00099\) & \(t=17.762\), \(p=0.00195\) \\
DPP vs Random 10 Rules  & \(t=5.166\), \(p=0.02415\) & \(t=5.557\), \(p=0.02204\) \\
\bottomrule
\end{tabular}
}
\caption{Comparison of DPP method with other methods in Math and Code domains using Welch's t-test.}
\label{tab:method_comparison_math_code}
\end{table*}

\subsection{Evaluation Benchmarks}
\label{appendix:benchmark}
In this section, we provide detailed descriptions of the benchmarks utilized for our evaluation. 
We considered the following benchmarks for fine-tuning: we use zero-shot in IMDB, and 5-shot for Medical and Math (which uses subsets of MMLU). Moreover, Math and Medical domains, we use the subject-related subsets from MMLU, specifically ElementaryMathematics, HighSchoolMathematics, and CollegeMathematics for Math, and CollegeMedicine, ProfessionalMedicine, and MedicalGenetics for Medical. For Code, we tested code generation and for each code benchmark, we use the pass@k setting and specify the number of code generation samples. See detailed explanations below.

\begin{itemize}
    \item \textbf{IMDB} \citep{maas-EtAl:2011:ACL-HLT2011}: The IMDB dataset comprises 50,000 movie reviews and is designed for binary sentiment classification. For our evaluation, we select 25,000 test samples.
    \item \textbf{MMLU} \citep{maas-EtAl:2011:ACL-HLT2011}: MMLU is a comprehensive multitask test comprises multiple-choice questions from a wide range of knowledge domains. It spans subjects across the humanities, social sciences, hard sciences, and other critical learning areas, encompassing 57 tasks such as elementary mathematics, US history, computer science, law, and more. To achieve high accuracy on this test, models need to demonstrate extensive world knowledge and robust problem-solving capabilities.
    \item \textbf{HumanEval} \citep{chen2021evaluating}: The HumanEval benchmark evaluates Python programming skills with 164 problems, each comprising a function signature, docstring, function body, and unit tests. In a zero-shot setting, models generate code using top-p sampling (p=0.95) until stop words are reached. Pass@k metrics (k=1, 10, 100) are calculated with n=200 samples per problem, estimating the success rate following Chen et al.'s approach. Success is determined by whether at least one solution is correct within k attempts, with temperature controlling randomness in generation. This benchmark measures model performance in solving programming tasks with increasing attempts.
    \item \textbf{MBPP} \citep{austin2021program}: The MBPP benchmark contains around 1,000 crowd-sourced Python programming problems, designed for entry-level programmers. Each problem includes a task description, a code solution, and 3 test cases. The evaluation is performed on the test set from index 11 to 511. In a few-shot setting, the InCoder-style prompt is used, where the task description and one solution are provided to guide the model. The prompt format is \texttt{f'"""\{description\}\{test\_example\}"""'}.. By default, \texttt{prompt\_type\_mbpp} is set to \texttt{incoder}, and optionally, the solution can be included using \texttt{include\_solution\_mbpp=True}. We use single generation per problem (pass@1), and for pass@k estimation, we generate n=15 samples per problem, similar to the HumanEval approach. The evaluation focuses on pass@1 success rates.

  \item \textbf{Multiple-py and Multiple-cpp} \citep{cassano2022multipl}: 
MultiPL-E: is a benchmark for evaluating large language models for code generation that supports 18 programming languages. It takes the OpenAI ``HumanEval'' Python benchmark and uses little compilers to translate them to other languages. We use similar implementation as the original repository and evaluation parameters are similar to HumanEval.
\end{itemize}

\subsection{Number of selected rules}\label{subsec:Appendix-EvalB-20rules}
The exploration of the scaling of $r$ in Section~\ref{sec:EvaluationA} is feasible due to the low cost of these experiments. However, in Section~\ref{sec:EvaluationB}, which involves high-cost LLM training, it is impractical to test multiple values of  $r$ across all experiments.

Here in Table~\ref{tab:vary_num_rules}, we modified the number of rules, $r$, from 10 to 20 and repeated the experiments for the Code domain. Compared to the 10-rule results presented in Table~\ref{tab:Finetune}, we observed some discrepancies. For instance, the performance score on HumanEval is less than the 10-rule results, whereas the results for Multiple-cpp improved. The number of rules indeed alters the criteria used for data selection, thereby influencing the distribution of the selected data. Determining the optimal $r$ represents a valuable direction for our future exploration.

\begin{table*}[t!]
    \centering
    \mytiny
    \resizebox{0.8\textwidth}{!}{
    \begin{tabular}{c | c c c c c} \hline 
           \multirow{2}*{Method} & \multicolumn{5}{c}{Code} \\
           \cline{2-6} 
           & humaneval & mbpp & multiple-py & multiple-cpp & Code Average \\ \hline 
            Random 20 Rules & 43.90  & 43.93 &  46.57  & 50.50  & 46.23\\
            DPP 20 Rules & 45.10  & 44.80  & 49.10  & 53.40 &  48.10\\ \hline
    \end{tabular}
    }
    \caption{Code fine-tuning on Llama3-8B using 20K selected data samples from our SlimPajama data source. Instead of using 10 rules, 20 rules were selected during the rule selection step.}
    \label{tab:vary_num_rules}
\end{table*}

\subsection{Size of Sampled Data}
We investigated the impact of varying training data sizes on performance in Table~\ref{tab:vary_data_size}, specifically within the context of the \textit{DPP 10 Rules} and the Medical domain, due to limited computing resources. Our observations reveal that increasing the amount of training data does not always enhance performance; in fact, performance may decline beyond a certain data threshold. This phenomenon is consistent with findings from the LIMA paper \citep{zhou2024lima}, which suggests that data quality is often more important than quantity for LLMs. Balancing data quality with quantity is another challenging but valuable topic.
\begin{table*}[t!]
    \centering
    \resizebox{0.8\textwidth}{!}{
    \begin{tabular}{ c | c  c  c  c} \hline 
           \multirow{2}*{Training Size} & \multicolumn{4}{c}{Medical} \\
           \cline{2-5} 
           & college medicine & professional medicine & medical genetics & Medical average\\ \hline 
            10K &  23.1 & 41.5 & 27.0 & 30.5\\
            20K & 24.3 &  43.0 & 26.0 & 31.1\\
            50K & 23.7 &  44.5 & 24.0 & 30.7\\
            100K & 23.7 &  39.3 & 24.0 & 29.0\\
            200K & 23.1 &  42.6 & 21.0 & 28.9\\\hline
    \end{tabular}
    }
    \caption{Medical fine-tuning on Pythia-1B using various sizes of training data selected by DPP with 10 rules.}
    \label{tab:vary_data_size}
\end{table*}

\subsection{Distribution of Selected Data}
Evaluating and contrasting the quality of data subsets selected by different methods is challenging and often necessitates extensive human intervention. To address this, we examined the initial 100 examples selected by each method. This examination revealed notable distinctions in the relevance and domain specificity of the data selected. Specifically, our DPP rule-based approach demonstrated a marked ability to identify and select examples that were highly pertinent to specific domains. For instance, in experiments focused on the Code domain, this method favored the inclusion of data containing code. In contrast, other less targeted methods, such as QuRating and Uniform Sampling, often yield selections that lack domain-specific relevance. This insight underscores the efficacy of using tailored, rule-based methods over generic ones for tasks where domain alignment is critical.

Although it is hard to compare the distribution of the selected data,  we provide a visual representation in Figure~\ref{fig:data_distribution_Code_meta} below, showcasing the meta-data (categories of the data samples) distributions for the Code domain as a representative example. Notably, the DPP methods with 10 and 50 rules tend to select more data from GitHub and StackExchange for Code fine-tuning.

\begin{figure*}[t!]
\centering
\begin{subfigure}{0.31\textwidth}
    \includegraphics[width=\linewidth]{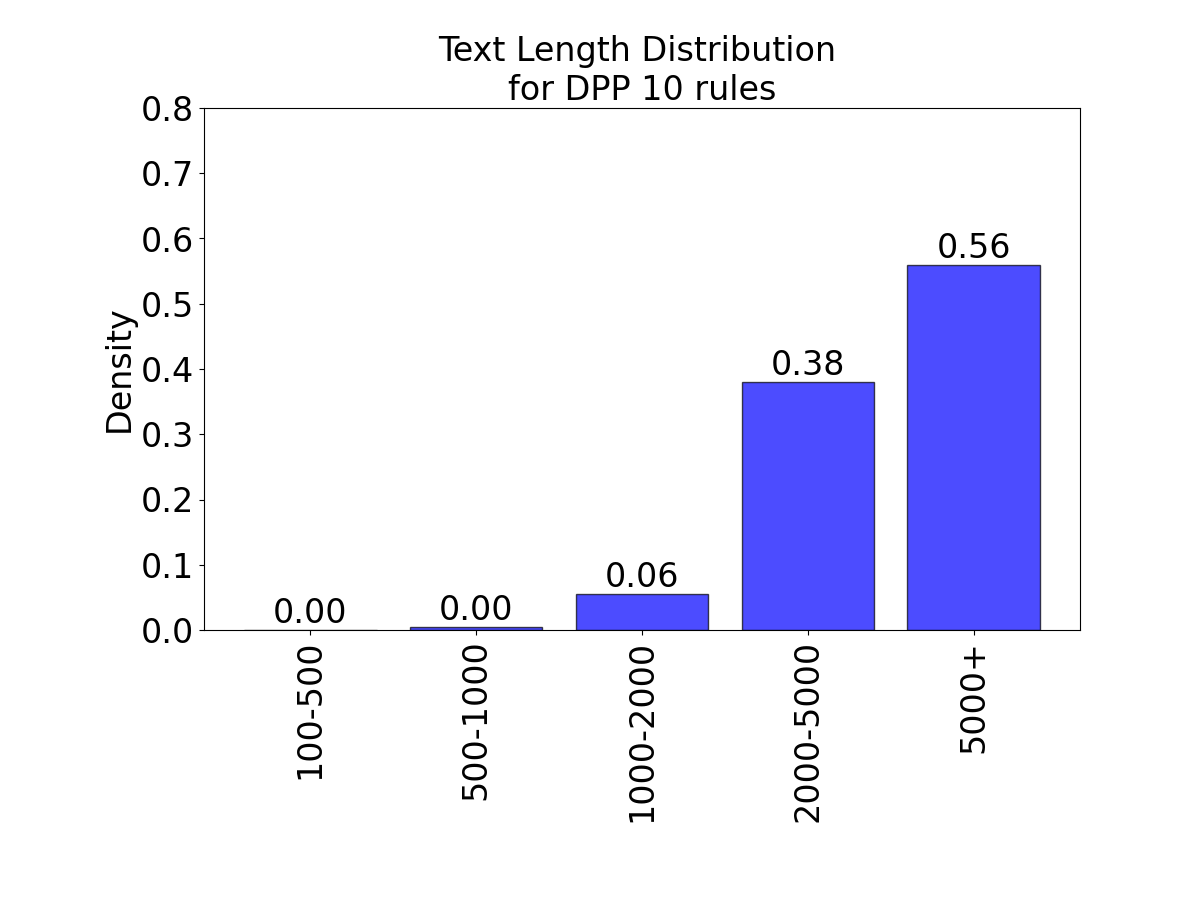}
    \caption{DPP 10 rules}
\end{subfigure}
\begin{subfigure}{0.31\textwidth}
    \includegraphics[width=\linewidth]{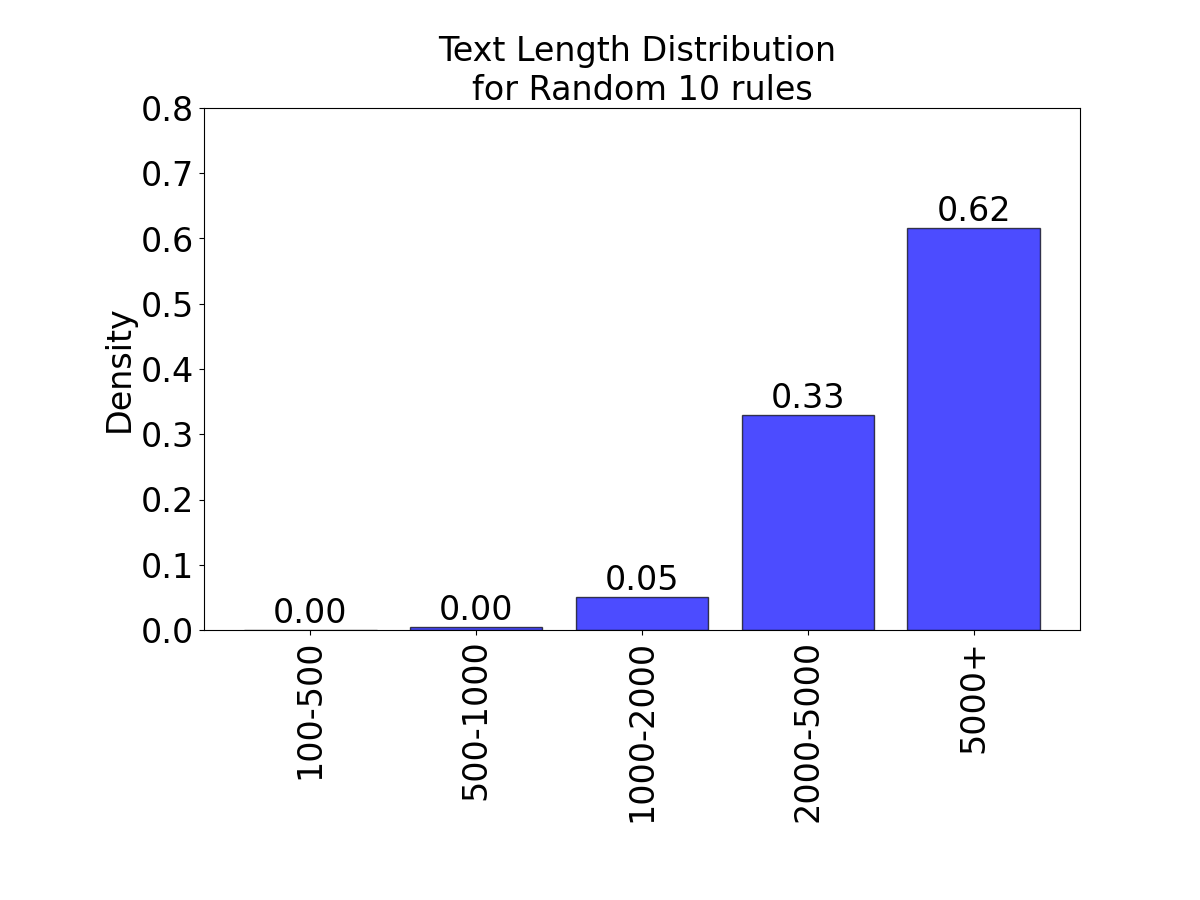}
    \caption{Random 10 rules}
\end{subfigure}
\begin{subfigure}{0.31\textwidth}
    \includegraphics[width=\linewidth]{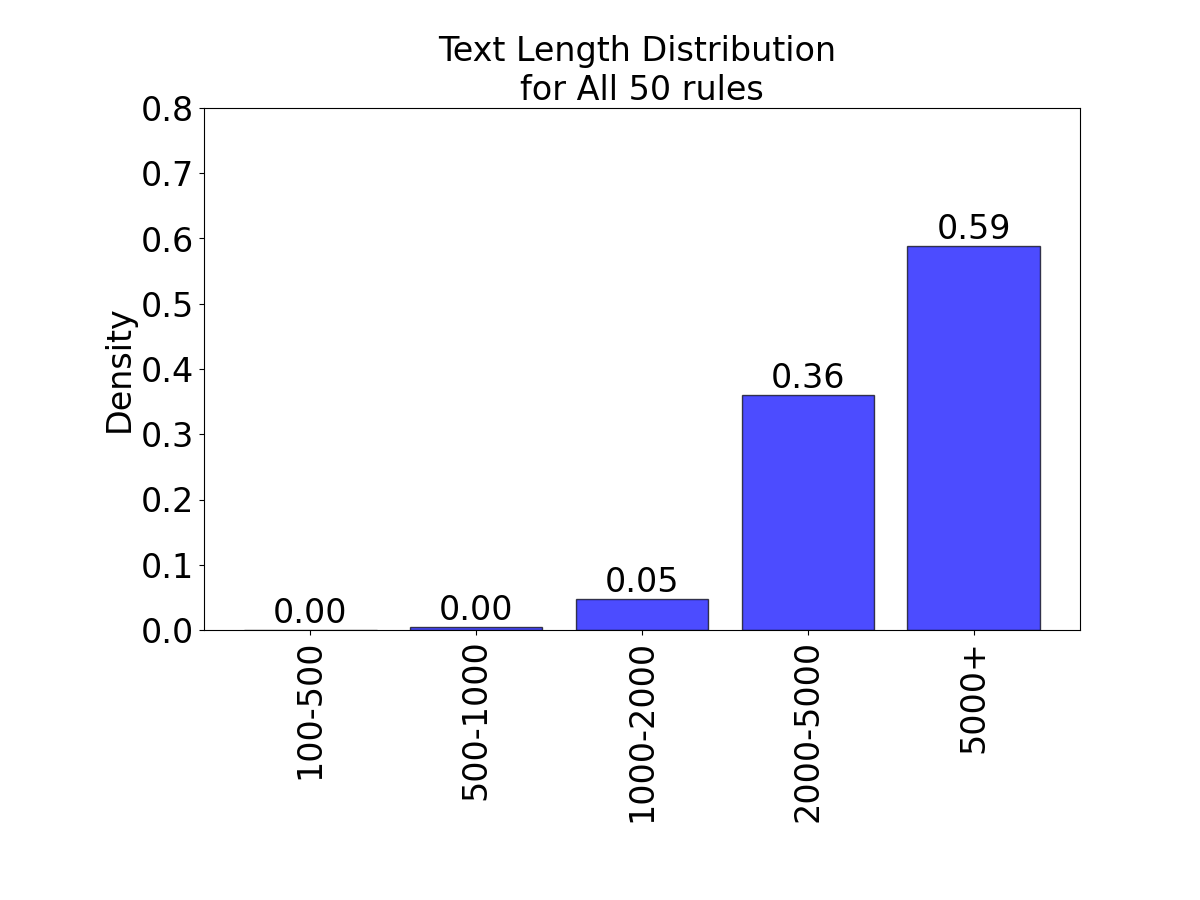}
    \caption{All 50 rules}
\end{subfigure}
\begin{subfigure}{0.31\textwidth}
    \includegraphics[width=\linewidth]{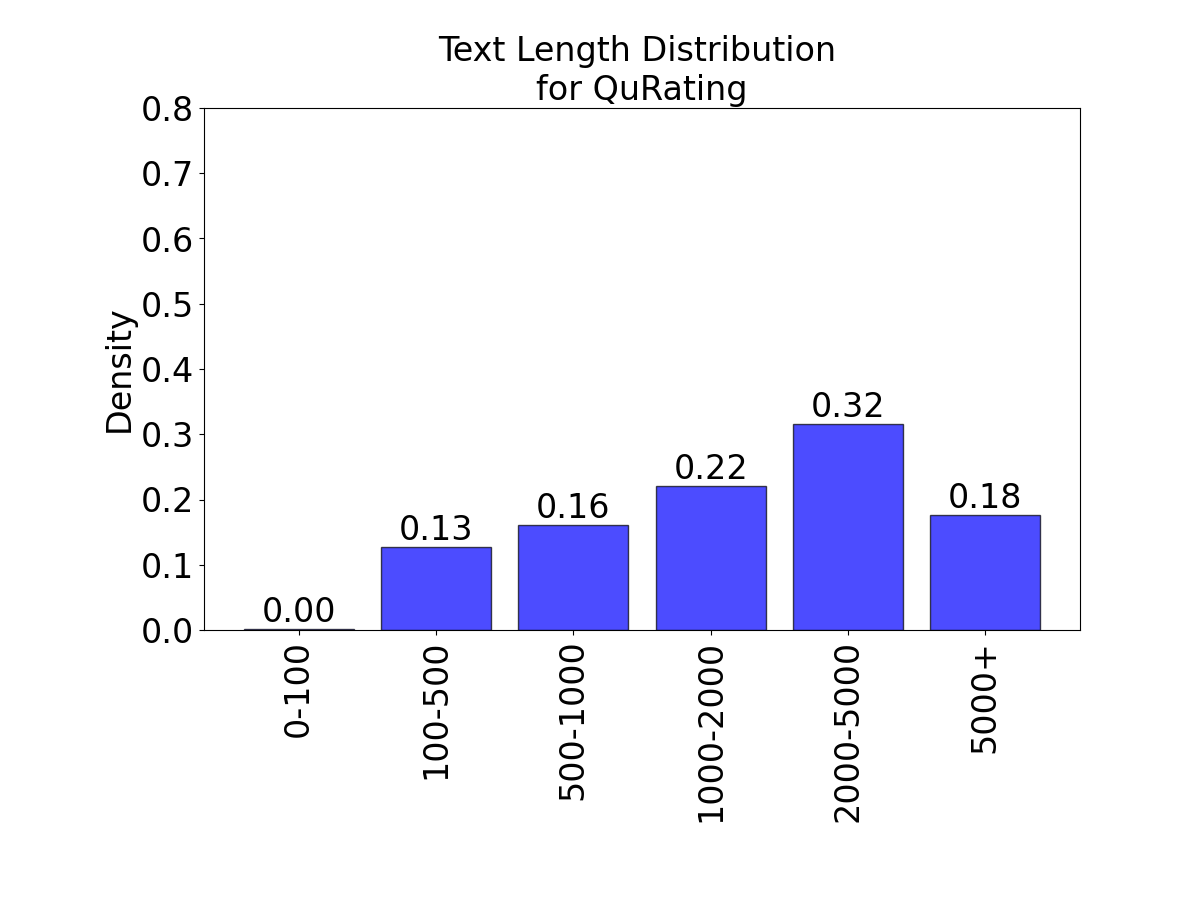}
    \caption{QuRating}
\end{subfigure}
\begin{subfigure}{0.31\textwidth}
    \includegraphics[width=\linewidth]{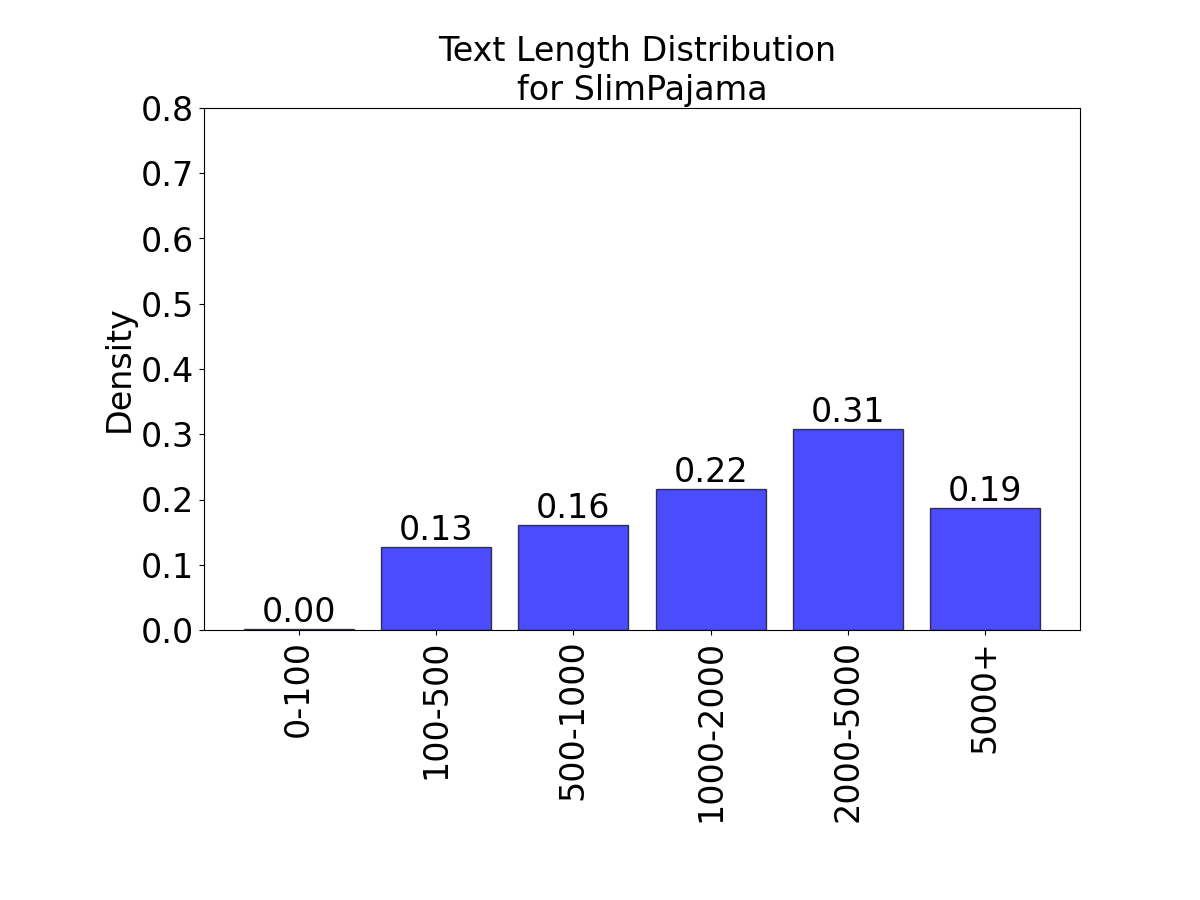}
    \caption{SlimPajama1M}
\end{subfigure}

\caption{Comparison of text length distribution across different methods. The last is the original distribution of our source data.} 
\label{fig:data_distribution_IMDB_length}
\end{figure*}

\begin{figure*}[t!]
\centering
\begin{subfigure}{0.31\textwidth}
    \includegraphics[width=\linewidth]{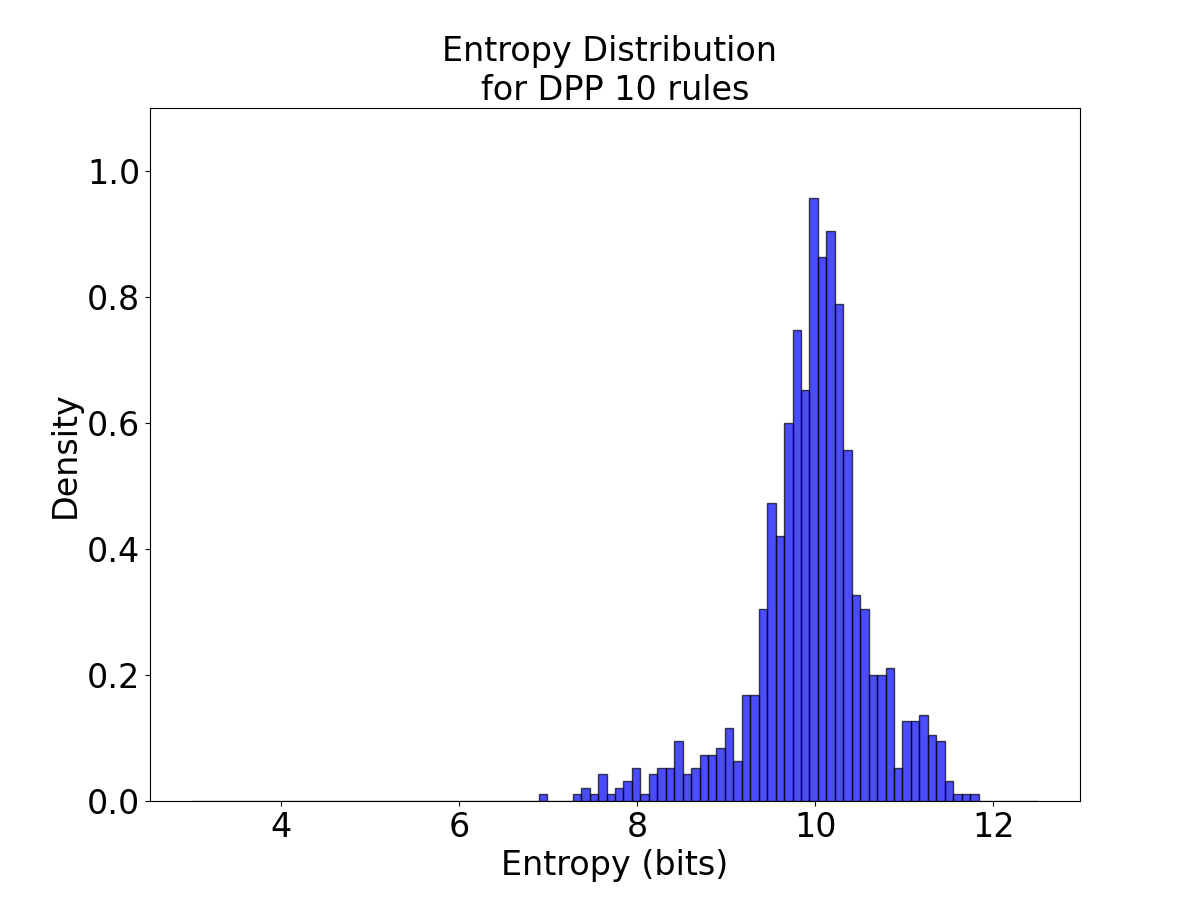}
    \caption{DPP 10 rules}
\end{subfigure}
\begin{subfigure}{0.31\textwidth}
    \includegraphics[width=\linewidth]{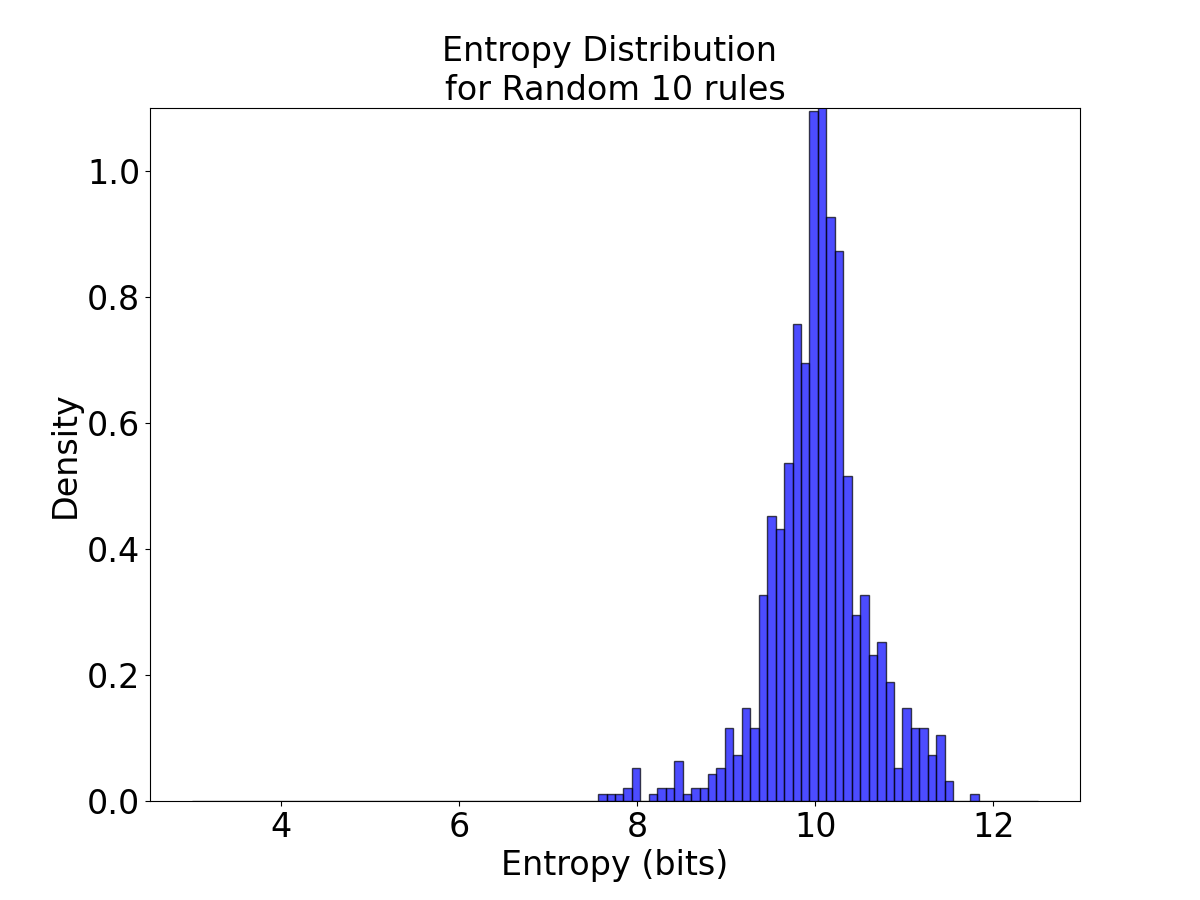}
    \caption{Random 10 rules}
\end{subfigure}
\begin{subfigure}{0.31\textwidth}
    \includegraphics[width=\linewidth]{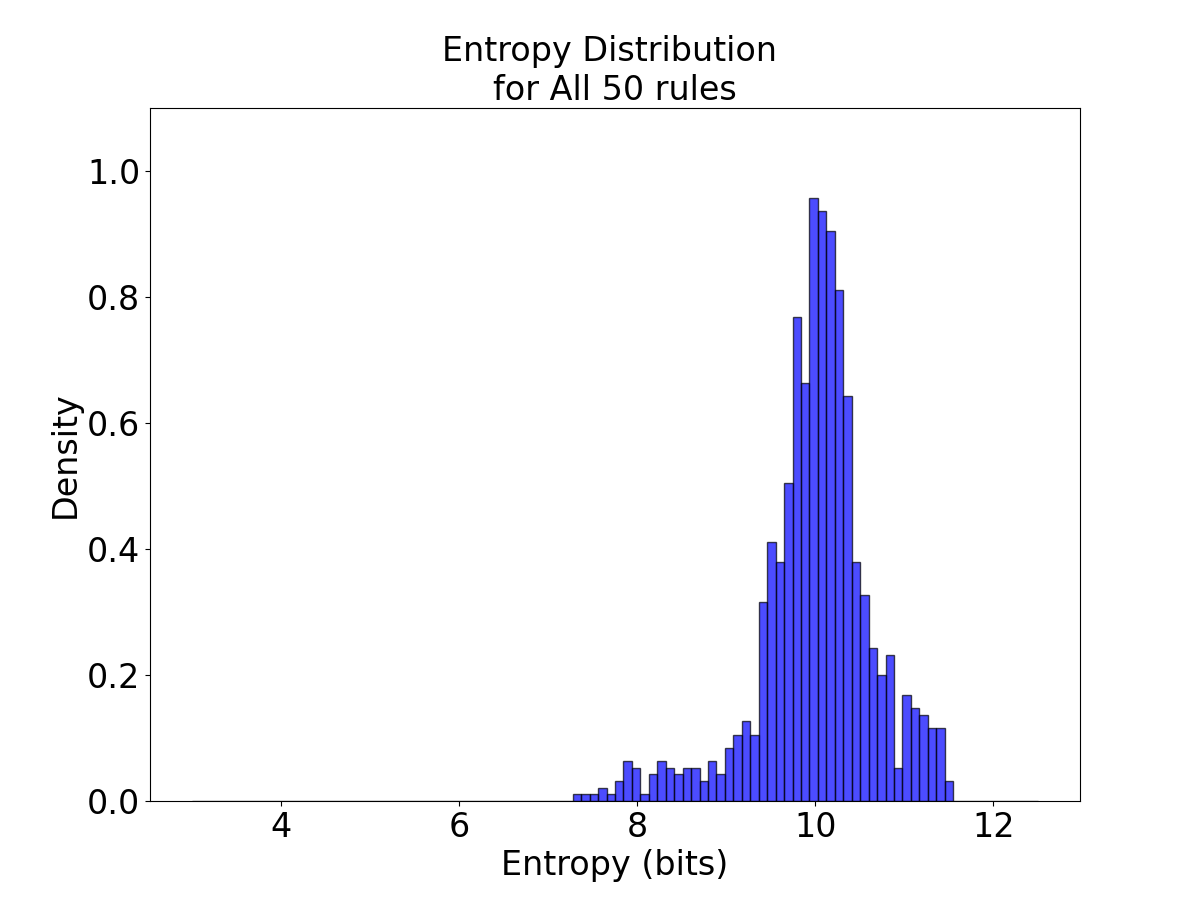}
    \caption{All 50 rules}
\end{subfigure}
\begin{subfigure}{0.31\textwidth}
    \includegraphics[width=\linewidth]{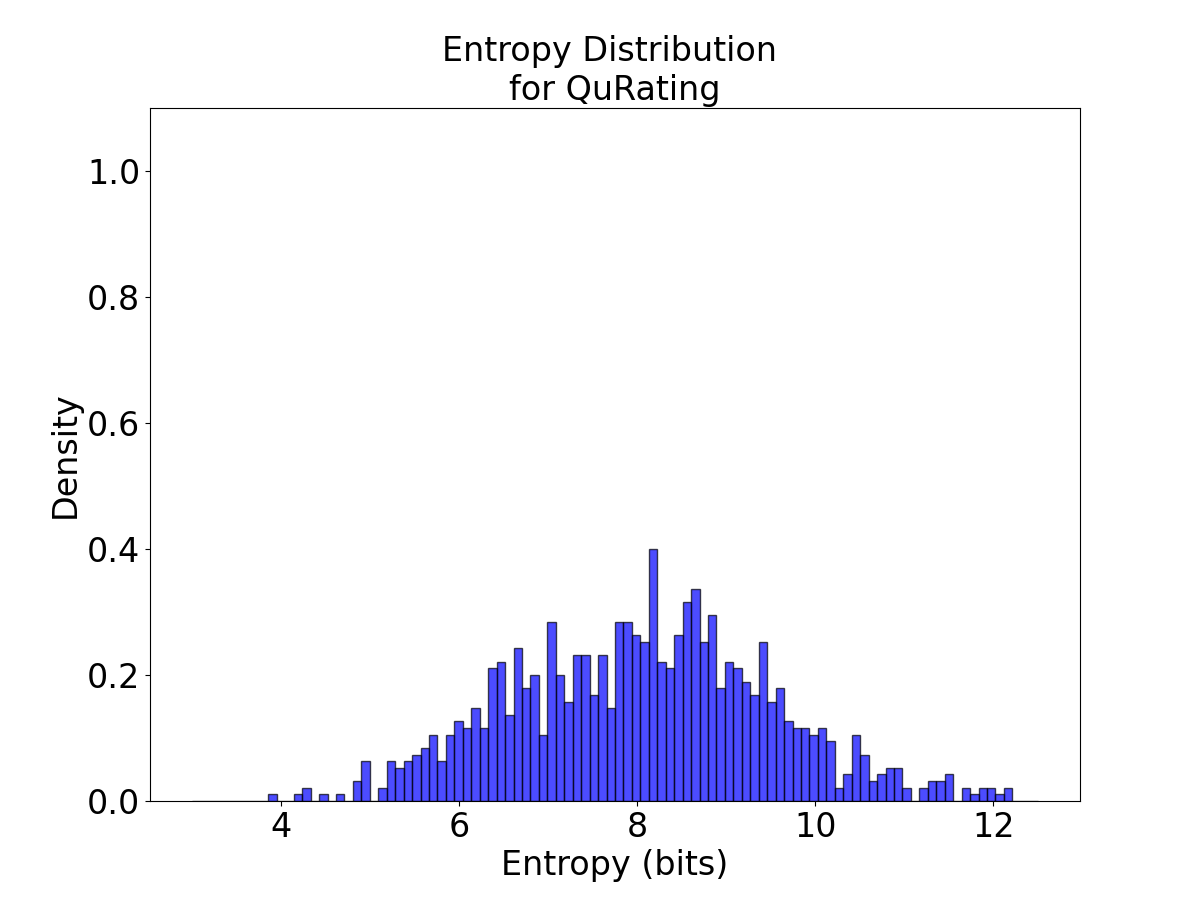}
    \caption{QuRating}
\end{subfigure}
\begin{subfigure}{0.31\textwidth}
    \includegraphics[width=\linewidth]{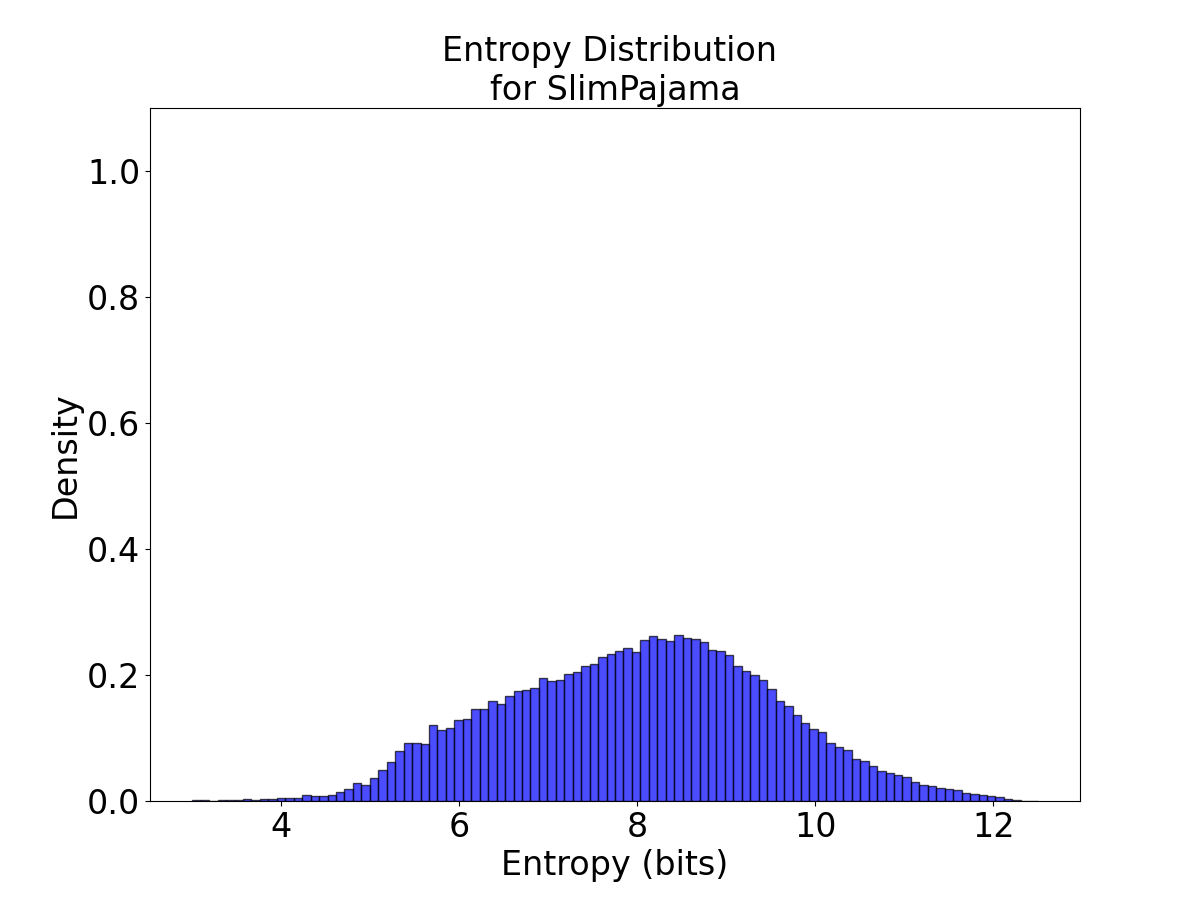}
    \caption{SlimPajama1M}
\end{subfigure}

\caption{Comparison of text diversity distribution across different methods. The last is the original distribution of our source data.} 
\label{fig:data_distribution_IMDB_entropy}
\end{figure*}

Moreover for IMDB domain, in Figure~\ref{fig:data_distribution_IMDB_length} we investigated the text length distribution. We see that the QuRating is very close to the original SlimPajama distribution, where we conjecture that in this case the data distribution is very close to uniformly sampled data. The methods within our framework have a tendency toward longer texts. Additionally, in Figure~\ref{fig:data_distribution_IMDB_entropy} we use bigram entropy (the Shannon entropy of the distribution over the unique bigrams) as an indicator of the text diversity. We again see that the entropy distribution of QuRating is very close to the original SlimPajama, where our methods generally select data with higher entropy/diversity and the entropy distributions are more concentrated.

\begin{figure*}[t!]
\centering

\begin{subfigure}{0.25\textwidth}
    \includegraphics[width=1.0\linewidth]{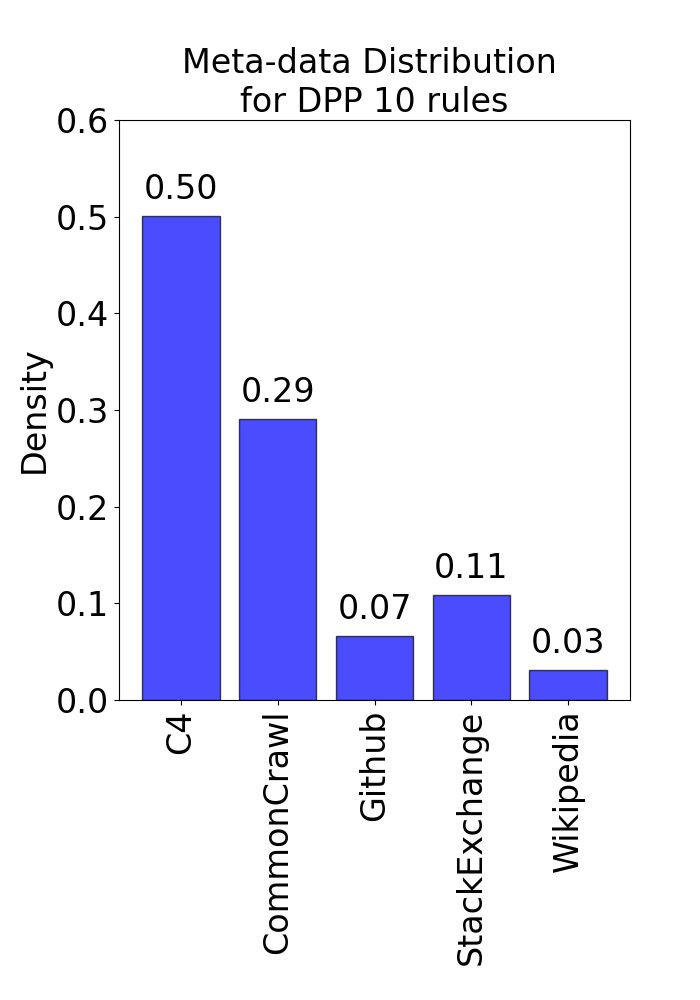}
    \caption{DPP 10 rules}
\end{subfigure}
\begin{subfigure}{0.25\textwidth}
    \includegraphics[width=1.0\linewidth]{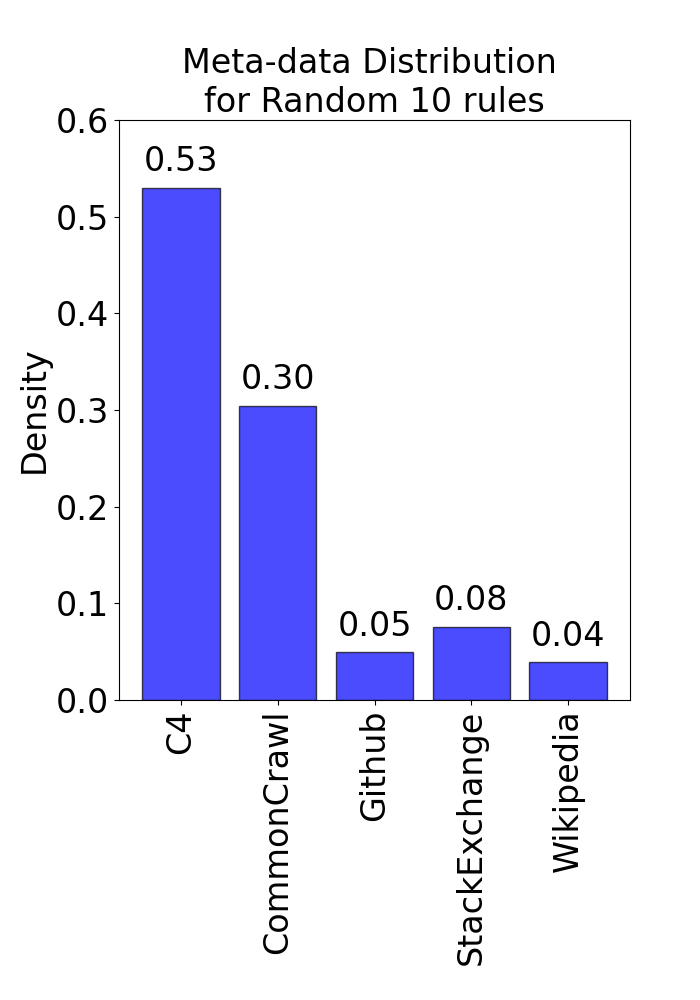}
    \caption{Random 10 rules}
\end{subfigure}
\begin{subfigure}{0.25\textwidth}
    \includegraphics[width=1.0\linewidth]{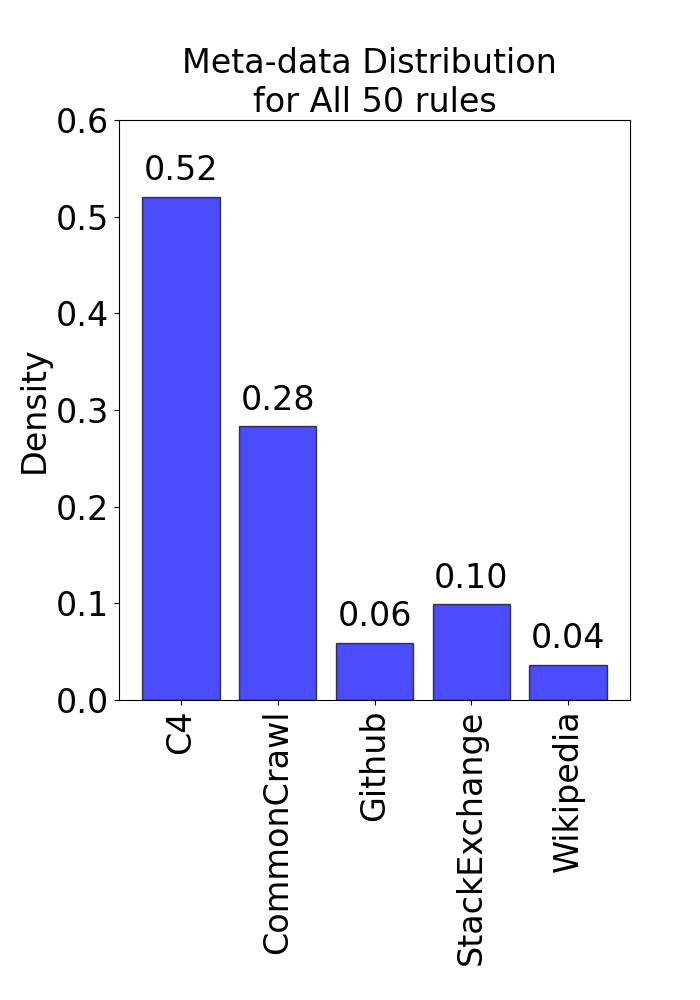}
    \caption{All 50 rules}
\end{subfigure}
\begin{subfigure}{0.25\textwidth}
    \includegraphics[width=1.0\linewidth]{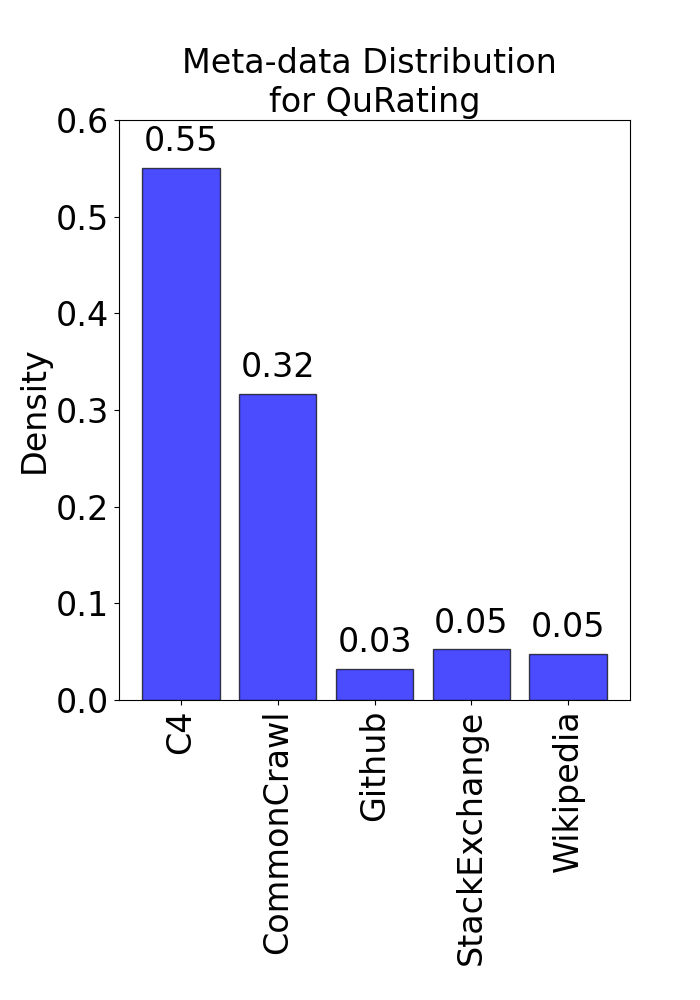}
    \caption{QuRating}
\end{subfigure}
\begin{subfigure}{0.25\textwidth}
    \includegraphics[width=1.0\linewidth]{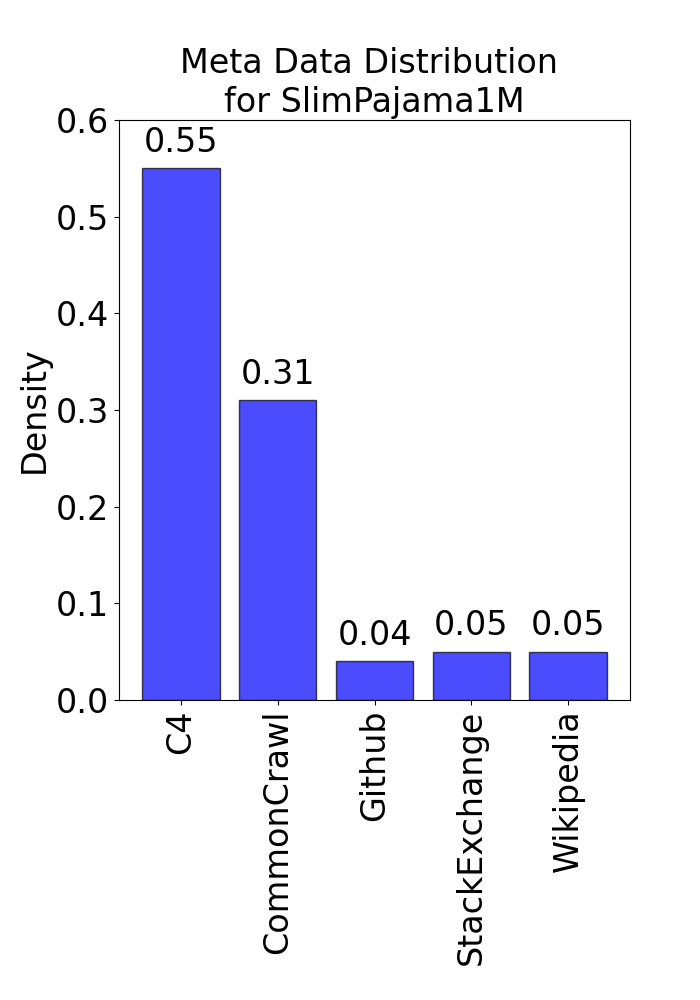}
    \caption{SlimPajama1M}
\end{subfigure}

\caption{Comparison of meta-data distribution across different methods. The last is the original distribution of our source data.} 
\label{fig:data_distribution_Code_meta}
\end{figure*}

\subsection{Prompts and Generated Rules}\label{subsec:Appendix-EvalB-Prompts-Rules}
\label{subsec:Appendix-EvalB-Prompt-Rules}
For brevity, we provide the templates for both the rule generation and rating prompts for the Math domain. To adapt these templates for other domains, replace terms specific to Math (such as ``mathematical tasks'' and ``mathematical reasoning and analysis'') with relevant terminology from the desired domain. We use GPT-4 to help us generate these task description and data descriptions.

\textbf{Rule Generation Prompts:} 
\begin{figure}[H]
\centering
\small
\begin{tcolorbox}[colback=cyan!10!white, 
                  colframe=cyan!30!white, 
                  width=0.50\textwidth, 
                  arc=4mm, 
                  auto outer arc,
                  ]
Generate 50 specific rules for rating data from the training dataset (SlimPajama), in order to select a high-quality subset to train large language models that will improve their performance on mathematical tasks. The descriptions of the training data and the downstream task are provided below. The rules should focus on various aspects such as data quality, relevance, diversity, and other characteristics that would be beneficial for mathematical reasoning and analysis.\\
\\
Description of training data:\\
\texttt{\color{red}<DATA\_DESCRIPTION>}\\
\\
Description of downstream task:\\
\texttt{\color{red}<TASK\_DESCRIPTION>}\\
\\
Requirements for the Rules:\\
Each rule should be concise and specific.\\
The rules could be basic text quality rules or task-related quality rules.\\
The rules should be written in clear, natural language and be easy to understand.\\
\\
Now, please generate the 50 rules.
\end{tcolorbox}
\caption{Example of a rule-generation prompt used to create 50 data rating rules for the Math domain.}
\label{tab:rule_generation_prompt}
\end{figure}

\textbf{Rating Prompts:}
\begin{figure}[H]
\centering
\small
\begin{tcolorbox}[colback=cyan!10!white, 
                  colframe=cyan!30!white, 
                  width=0.50\textwidth, 
                  arc=4mm, 
                  auto outer arc,
                  ]
We are training a language model using the SlimPajama dataset to improve performance on mathematical tasks. Evaluate the following example from SlimPajama dataset and assign a quality score between 0 and 1 (0 indicates the worst quality, and 1 indicates perfect quality) according to the provided rule:\\
\texttt{\color{red}<RULE>}\\
\\
Example:\\
\texttt{\color{red}<DATA\_SAMPLE>}\\\\
Respond only with a single float number.
\end{tcolorbox}
\caption{Example of rule-rating prompt. Here we query the LLM to rate a single data sample based on a specific Math-related rule.}
\label{tab:rule_rating_prompt}
\end{figure}

\textbf{Task and data descriptions:}
\begingroup
\mytiny

\begin{table*}[t!]
\centering
\mytiny
\begin{tabular}{c|p{11cm}}
\hline
\textbf{Type} & \multicolumn{1}{c}{\textbf{Description}} \\ \hline
SlimPajama data description & The SlimPajama dataset is a large-scale dataset. It is designed to be a compact, high-quality dataset curated for pre-training large language models. The dataset includes a diverse range of texts, sourced from various domains such as web pages, books, and academic articles, providing a rich and varied training corpus for developing robust and versatile language models. \\
\hline
IMDB task description & The IMDB review dataset, created by StanfordNLP, is a widely used dataset for sentiment analysis. It contains 50,000 highly polar movie reviews. Each review is labeled as either positive or negative, making it an ideal dataset for binary sentiment classification tasks. The dataset provides a challenging benchmark for evaluating the performance of sentiment analysis models. \\
\hline
Medical task description & The MMLU (Massive Multitask Language Understanding) includes three medical-related subsets: mmlu\_college\_medicine, mmlu\_medical\_genetics, and mmlu\_professional\_medicine. These subsets test a language model's understanding of general medical knowledge, genetic concepts, and advanced professional medical practices, respectively, through multiple-choice questions tailored to assess both foundational and specialized medical expertise. \\
\hline
Math task description & The MMLU (Massive Multitask Language Understanding) includes a range of subsets designed to evaluate language models across various academic subjects, including mathematics. The Math subsets specifically assess a model's capability to understand and solve mathematical problems. These are categorized into multiple difficulty levels—from elementary mathematics to college-level topics like abstract algebra. Each subset consists of multiple-choice questions that test different areas of mathematical knowledge, aiming to measure both basic arithmetic skills and more complex mathematical reasoning. This structure allows researchers to gauge a model's proficiency in mathematical logic and its application to solve real-world problems. \\
\hline
Code task description & The Code Generation LM Evaluation Harness, part of the BigCode project, is a framework designed to evaluate large language models (LLMs) on their ability to generate code. It provides a structured environment to assess the performance of these models across various programming tasks and languages. The harness supports automated evaluation metrics and facilitates benchmark comparisons, making it a valuable tool for researchers and developers aiming to enhance the code generation capabilities of LLMs. \\
\hline
\end{tabular}
\caption{Data descriptions of SlimPajama and task descriptions of four domains.}
\label{tab: task_description}
\end{table*}
\endgroup

\textbf{Generated 50 rules for each of four domains:}
Note that the IMDB rules here are used to select data for LLM training, whereas the IMDB rules in \ref{subsec:Appendix-EvalA-Prompts-Rules} are used for data comparison in order to eventually calculate quality scores for the $50$ IMDB reviews. Although similar, they are not the same set of rules.
\begin{table*}[t!]
\small 
\caption{Generated rules for the IMDB domain.}\label{tab:rules_for_IMDB}
\setlength{\tabcolsep}{6pt}
\renewcommand{\arraystretch}{1.15}
\begin{tabularx}{\textwidth}{c|X}
\toprule
\textbf{Index} & \textbf{Rule Description} \\
\midrule
0  & Text Length: Be between 100 and 1000 words to match the typical length of IMDB reviews. \\
\hline
1  & Sentiment Clarity: Clearly express either positive or negative sentiments. \\
\hline
2  & Language Quality: Have fewer than 2 spelling or grammatical errors per 100 words. \\
\hline
3  & Language Focus: Be in English to maintain focus on the language of the target dataset. \\
\hline
4  & Source Diversity: Be sourced evenly from web pages, books, and academic articles. \\
\hline
5  & Tone Appropriateness: Minimize neutral tones as they are less useful for binary sentiment analysis. \\
\hline
6  & Cultural Relevance: Discuss culturally significant topics relevant to a global English-speaking audience. \\
\hline
7  & Language Style: Use informal, conversational language. \\
\hline
8  & Sarcasm Avoidance: Avoid sarcasm to prevent misinterpretation by sentiment analysis models. \\
\hline
9  & Subjectivity: Express opinions rather than just stating facts. \\
\hline
10 & Emotional Expression: Express emotions to aid in sentiment understanding. \\
\hline
11 & Redundancy Avoidance: Avoid redundancy and excessive similarity to other texts in the dataset. \\
\hline
12 & Contemporary Relevance: Be from the past decade to ensure relevance. \\
\hline
13 & Industry Relevance: Include mentions of movies, actors, or film industry terms. \\
\hline
14 & Sentiment Indicators: Contain explicit sentiment indicators. \\
\hline
15 & Sentence Complexity: Feature complex sentence structures. \\
\hline
16 & Figurative Language: Use metaphors and similes. \\
\hline
17 & Contextual Richness: Provide enough context to understand the sentiment on their own. \\
\hline
18 & Jargon Avoidance: Avoid heavy use of irrelevant technical jargon. \\
\hline
19 & Format Appropriateness: Avoid non-continuous formats like lists and tables. \\
\hline
20 & Persuasiveness: Be persuasive, reflecting the tone often found in positive or negative reviews. \\
\hline
21 & Genre Balance: Represent a balanced variety of genres (e.g., fiction, non-fiction, journalism). \\
\hline
22 & Citation Minimization: Avoid being predominantly composed of citations or quotes. \\
\hline
23 & Interactive Media Handling: Exclude interactive media texts unless they provide narrative value. \\
\hline
24 & Structural Cohesion: Be cohesive and well-structured. \\
\hline
25 & Offensive Content Avoidance: Avoid containing hate speech, excessive violence, or other offensive content. \\
\hline
26 & Demographic Inclusivity: Discuss or be relevant to a variety of demographic groups. \\
\hline
27 & Sentiment Extremity: Express strong sentiments, either positive or negative. \\
\hline
28 & Colloquial Language: Mimic spoken language, as often found in movie reviews. \\
\hline
29 & Descriptive Nature: Avoid being purely descriptive and lack subjective opinions. \\
\hline
30 & Historical Context: Include historical references only if they enhance the sentiment or narrative. \\
\hline
31 & Plagiarism Avoidance: Be free from plagiarism. \\
\hline
32 & Domain-Specific Language: Contain relevant film and media terms. \\
\hline
33 & User-Generated Content: Include user-generated content such as blogs and user reviews. \\
\hline
34 & Narrative Emphasis: Be narrative-driven, resembling the storytelling found in reviews. \\
\hline
35 & Error Avoidance: Avoid formatting or data errors. \\
\hline
36 & Topical Relevance: Discuss topics commonly found in movie reviews such as plot, acting, and direction. \\
\hline
37 & Satire Handling: Avoid satire unless it is clearly marked or well-known. \\
\hline
38 & Subject Line Clarity: Have moderate and descriptive subject lines. \\
\hline
39 & Outdated Content Avoidance: Avoid containing outdated societal views or terminologies. \\
\hline
40 & Regional Representation: Represent various English dialects and regional variations. \\
\hline
41 & Emotional Variability: Exhibit a range of emotions from joy to sadness, to anger. \\
\hline
42 & Controversial Topic Inclusion: Include discussions on controversial topics if they enhance sentiment understanding. \\
\hline
43 & Generalization Avoidance: Avoid making broad generalizations without substantiation. \\
\hline
44 & Source Reliability: Be from reliable and reputable sources. \\
\hline
45 & Uniqueness: Be unique with no duplicates in the dataset. \\
\hline
46 & Formality Variance: Include a variety of formality levels, particularly matching the informal style of many movie reviews. \\
\hline
47 & Impactful Sentences: Contain emotionally resonant sentences critical for sentiment analysis. \\
\hline
48 & Engagement: Be engaging and likely to provoke reader reactions. \\
\hline
49 & Visual Storytelling: Include vivid descriptions akin to visual storytelling in movies. \\
\bottomrule
\end{tabularx}
\end{table*}

\begin{table*}[t!]
\small 
\caption{Generated rules for the Medical domain.}\label{tab:rules_for_Medical}
\setlength{\tabcolsep}{6pt}
\renewcommand{\arraystretch}{1.15}
\begin{tabularx}{\textwidth}{c|X}
\toprule
0  & Relevance to Medical Topics: Include texts that contain medical terminology or discuss medical topics. \\
\hline
1  & Exclusion of Non-Medical Content: Exclude texts that do not pertain to health, medicine, or biological sciences. \\
\hline
2  & Clarity of Medical Information: Select texts where medical information is clearly explained and easy to understand. \\
\hline
3  & Accuracy of Medical Content: Ensure texts contain medically accurate information, verified against reputable medical sources. \\
\hline
4  & Diversity of Medical Subfields: Include texts covering a range of medical fields such as genetics, anatomy, pharmacology, and pathology. \\
\hline
5  & Contemporary Relevance: Prefer texts discussing current medical practices and technologies over outdated treatments. \\
\hline
6  & Technical Depth: Include texts with a deep, technical discussion of medical topics suitable for professional medicine. \\
\hline
7  & Exclusion of Ambiguous Content: Avoid texts with ambiguous or unclear medical claims or data. \\
\hline
8  & Citation of Sources: Select texts that cite reputable medical journals or textbooks. \\
\hline
9  & Grammar and Spelling: Ensure texts are free from grammatical errors and spelling mistakes. \\
\hline
10 & Use of Professional Language: Prefer texts that utilize professional medical jargon correctly. \\
\hline
11 & Inclusion of Case Studies: Include texts that discuss medical case studies or clinical trials. \\
\hline
12 & Representation of Rare Diseases: Ensure inclusion of texts discussing rare or less common diseases. \\
\hline
13 & Coverage of Ethical Considerations: Include texts discussing ethical considerations in medical practice and research. \\
\hline
14 & Language Diversity: Include texts in multiple languages relevant to global medical practice. \\
\hline
15 & Patient Education Focus: Include texts aimed at patient education that explain medical conditions and treatments clearly. \\
\hline
16 & Statistical Data Presentation: Prefer texts that present medical data and statistics clearly. \\
\hline
17 & Illustration of Medical Procedures: Include texts with detailed descriptions or illustrations of medical procedures. \\
\hline
18 & Pharmacological Content: Include texts discussing drug mechanisms, interactions, side effects, and benefits. \\
\hline
19 & Genetic Concepts Coverage: Ensure texts covering genetic concepts are detailed and accurate. \\
\hline
20 & Medical Research Updates: Include texts with the latest research findings in the medical field. \\
\hline
21 & Interdisciplinary Approach: Select texts that integrate medical knowledge with other sciences like biochemistry or physics. \\
\hline
22 & Historical Medical Milestones: Include texts discussing historical advancements in medicine. \\
\hline
23 & Medical Guidelines and Protocols: Include texts that detail medical guidelines, protocols, or standard operating procedures. \\
\hline
24 & Interviews with Medical Professionals: Include interviews or discussions with recognized experts in the medical field. \\
\hline
25 & Patient Case Confidentiality: Exclude texts that potentially breach patient confidentiality or privacy. \\
\hline
26 & Texts from Medical Conferences: Include content from recent medical conferences or symposiums. \\
\hline
27 & Exclusion of Pseudoscience: Strictly exclude texts promoting unverified or pseudoscientific claims. \\
\hline
28 & Clinical Pathway Discussions: Include texts discussing clinical decision-making processes and pathways. \\
\hline
29 & Medical Device Descriptions: Include texts that describe the use and innovation of medical devices. \\
\hline
30 & Nutritional and Lifestyle Medicine: Include texts discussing the impact of nutrition and lifestyle on health. \\
\hline
31 & Pediatric Medicine Coverage: Ensure texts covering pediatric medicine are included. \\
\hline
32 & Mental Health Discussions: Include texts that address various aspects of mental health care. \\
\hline
33 & Healthcare Policy Analysis: Include texts analyzing healthcare policies and their implications. \\
\hline
34 & Disease Prevention Focus: Include texts focused on disease prevention strategies and methods. \\
\hline
35 & Surgical Techniques Description: Prefer texts that detail surgical procedures and techniques. \\
\hline
36 & Medical Training and Education: Include texts related to medical training and education methods. \\
\hline
37 & Veterinary Medicine: Include texts on veterinary medicine where relevant to comparative medicine. \\
\hline
38 & Environmental Health Issues: Include texts discussing the impact of environmental factors on health. \\
\hline
39 & Bioinformatics Data Handling: Include texts discussing the handling and analysis of bioinformatics data. \\
\hline
40 & Medical Imaging Techniques: Include texts discussing modern medical imaging techniques and their applications. \\
\hline
41 & Cultural Competence in Healthcare: Include texts that discuss cultural considerations in healthcare provision. \\
\hline
42 & Global Health Challenges: Include texts discussing global health issues and strategies. \\
\hline
43 & Emergency Medicine Protocols: Include texts detailing protocols and procedures in emergency medicine. \\
\hline
44 & Health Insurance Systems: Include texts discussing different health insurance systems and policies. \\
\hline
45 & Medical Ethics Case Studies: Include case studies discussing medical ethics dilemmas and resolutions. \\
\hline
46 & Integrative Medicine Approaches: Include texts on integrative approaches combining traditional and modern medicine. \\
\hline
47 & AI and Machine Learning in Medicine: Include discussions on the application of AI and machine learning in medical contexts. \\
\hline
48 & Telemedicine and Remote Care: Include texts on the advancements and challenges in telemedicine. \\
\hline
49 & Healthcare Accessibility and Equity: Include texts discussing issues of accessibility and equity in healthcare. \\
\bottomrule
\end{tabularx}
\end{table*}

\begin{table*}[t]
\small 
\caption{Generated rules for the Math domain.}\label{tab:rules_for_Math}
\setlength{\tabcolsep}{6pt}
\renewcommand{\arraystretch}{1.15}
\begin{tabularx}{\textwidth}{c|X}
\toprule
0 & Mathematical Keywords: Prioritize texts containing keywords related to mathematics such as `algebra', `calculus', `geometry', `equations', `theorems', etc. \\
\hline
1 & Problem Statements: Include examples that present mathematical problems or puzzles. \\
\hline
2 & Solution Explanations: Select texts that not only present problems but also explain solutions step-by-step. \\
\hline
3 & High-Quality Sources: Favor texts sourced from academic articles, educational websites, and textbooks over general web pages. \\
\hline
4 & Symbolic Representation: Ensure the presence of mathematical symbols and expressions formatted in LaTeX or similar markup languages. \\
\hline
5 & Advanced Topics Coverage: Include texts that cover advanced mathematical topics such as differential equations, statistics, and abstract algebra. \\
\hline
6 & Logical Structuring: Texts should demonstrate clear logical structuring, particularly in argumentation and problem-solving. \\
\hline
7 & Historical Context: Include content that provides historical context or development of mathematical theories and applications. \\
\hline
8 & Data Sets and Examples: Prioritize texts that include real-world data sets or examples where mathematical principles are applied. \\
\hline
9 & No Misconceptions: Exclude texts containing mathematical misconceptions or common errors unless they are being corrected. \\
\hline
10 & Illustrations and Diagrams: Include texts with clear diagrams, graphs, and illustrations that aid mathematical understanding. \\
\hline
11 & Proofs and Theorems: Include detailed explanations of proofs and discussions of theorems. \\
\hline
12 & Mathematics in Technology: Include examples that link mathematics with its applications in technology and engineering. \\
\hline
13 & Interdisciplinary Links: Select texts that illustrate the application of mathematics in other scientific disciplines like physics and chemistry. \\
\hline
14 & Question and Answer Format: Include texts that follow a question and answer format, especially for complex mathematical concepts. \\
\hline
15 & Exclusion of Irrelevant Content: Exclude texts that are primarily non-mathematical in nature, such as pure narrative or opinion pieces. \\
\hline
16 & Mathematical Definitions: Include texts that provide clear definitions of mathematical terms and concepts. \\
\hline
17 & Tutorial Style: Select tutorial-style texts that are aimed at teaching or explaining mathematical concepts. \\
\hline
18 & Accuracy of Content: Exclude any text with factual inaccuracies related to mathematics. \\
\hline
19 & Age-Appropriate Content: Select content that is appropriate for the educational level, from elementary to college-level mathematics. \\
\hline
20 & Challenge Level: Include texts with varying levels of difficulty to ensure a range of challenges in problem-solving. \\
\hline
21 & Language Clarity: Ensure the text uses clear and precise language appropriate for teaching or explaining mathematics. \\
\hline
22 & Cultural Diversity: Include mathematical content from diverse cultural backgrounds to promote inclusivity. \\
\hline
23 & Recency of Content: Prioritize recent texts that reflect the current state of mathematical education and theory. \\
\hline
24 & Real-World Applications: Select texts that discuss the application of mathematical concepts in real-world scenarios. \\
\hline
25 & Peer-Reviewed Sources: Favor texts extracted from peer-reviewed academic journals and conferences. \\
\hline
26 & Multiple Perspectives: Include texts that present multiple perspectives or methods for solving a single mathematical problem. \\
\hline
27 & Step-by-Step Guides: Prioritize texts that provide step-by-step guides to solving mathematical problems. \\
\hline
28 & Integration of Tools: Include texts that discuss or utilize mathematical tools and software. \\
\hline
29 & Variety of Formats: Include a variety of text formats such as articles, essays, and problem sets. \\
\hline
30 & Consistency in Terminology: Ensure consistency in mathematical terminology across the selected texts. \\
\hline
31 & Explanatory Footnotes: Include texts that make use of footnotes or side-notes to explain complex terms or provide additional context. \\
\hline
32 & Interactive Elements: Select texts that include or suggest interactive elements like quizzes or interactive diagrams. \\
\hline
33 & Avoid Redundancy: Avoid texts that are redundant in content, especially if they do not add new information or perspective. \\
\hline
34 & Mathematical Puzzles: Include texts that feature mathematical puzzles and games to enhance problem-solving skills. \\
\hline
35 & Comparative Analyses: Select texts that involve comparative analyses of different mathematical methods or theories. \\
\hline
36 & Language Models and Mathematics: Include texts discussing the intersection of language processing models and mathematics. \\
\hline
37 & Excerpts from Lectures: Include transcribed excerpts from academic lectures on mathematics. \\
\hline
38 & Mathematical Narratives: Include narratives that weave mathematical concepts into broader storylines or real-life applications. \\
\hline
39 & Authoritative Authors: Prioritize texts authored by well-regarded mathematicians or educators. \\
\hline
40 & Exclusion of Vague Language: Avoid texts that use vague or ambiguous language when explaining mathematical concepts. \\
\hline
41 & Feedback Loops: Include texts that describe the importance of feedback loops in mathematical learning. \\
\hline
42 & Error Analysis: Include texts that focus on error analysis in mathematical calculations or theories. \\
\hline
43 & Cross-Referencing: Favor texts that cross-reference other works or theories effectively. \\
\hline
44 & Mathematical Software Tutorials: Include tutorials or guides on using mathematical software. \\
\hline
45 & Engagement Metrics: Favor texts that have historically engaged readers or viewers, indicating quality and interest. \\
\hline
46 & Student Contributions: Include texts written by students, which can provide fresh perspectives and innovative approaches. \\
\hline
47 & Reviews and Critiques: Select texts that review or critique mathematical theories or textbooks. \\
\hline
48 & Accessibility Features: Include texts that are accessible to people with disabilities, such as those formatted for screen readers. \\
\hline
49 & Alignment with Curriculum: Ensure that the content aligns well with standard mathematical curriculums at various educational levels. \\
\bottomrule
\end{tabularx}
\end{table*}

\begin{table*}[t]
\small 
\caption{Generated rules for the Code domain.}\label{tab:rules_for_Code}
\setlength{\tabcolsep}{6pt}
\renewcommand{\arraystretch}{1.15}
\begin{tabularx}{\textwidth}{c|X}
\toprule
0 & Syntax Highlighting: Include texts that contain syntax highlighting or structured code comments. \\
\hline
1 & Grammar Quality: Exclude texts with excessive spelling and grammatical errors. \\
\hline
2 & Programming Keywords: Prioritize samples containing programming language keywords and constructs. \\
\hline
3 & Language Focus: Exclude texts that are predominantly non-English unless they are code snippets. \\
\hline
4 & Concept Explanation: Select texts with clear, concise explanations of programming concepts. \\
\hline
5 & Reputable Sources: Prioritize texts from reputable sources like well-known programming blogs and documentation sites. \\
\hline
6 & Minimum Length: Exclude texts that contain less than 50 words as they may not provide sufficient context. \\
\hline
7 & Best Practices: Include examples that demonstrate best coding practices. \\
\hline
8 & Language Diversity: Prioritize texts that include diverse programming languages covered in the BigCode project. \\
\hline
9 & Error Solutions: Select texts that provide examples of common programming errors and their solutions. \\
\hline
10 & Technique Comparison: Include texts with comparative discussions of different coding techniques or tools. \\
\hline
11 & Current Practices: Exclude texts with outdated or deprecated coding practices. \\
\hline
12 & Algorithm Explanation: Prioritize texts that include algorithm explanations with code snippets. \\
\hline
13 & Duplication Check: Exclude samples that are heavily duplicated within the dataset. \\
\hline
14 & API Usage: Select samples that demonstrate use of APIs from well-known software libraries. \\
\hline
15 & Multi-Language Code: Include texts with embedded code in multiple programming languages. \\
\hline
16 & Development Paradigms: Prioritize texts that discuss software development paradigms (e.g., object-oriented programming). \\
\hline
17 & Relevance Check: Exclude non-relevant texts like purely historical accounts of programming without technical details. \\
\hline
18 & Decision Context: Include texts that provide context on why certain coding decisions are made. \\
\hline
19 & Annotated Code: Prioritize texts that contain code with annotations explaining each part of the code. \\
\hline
20 & Step-by-Step Code: Include samples where code is broken down into step-by-step explanations. \\
\hline
21 & Non-Promotional: Exclude texts that are purely promotional or sales-focused. \\
\hline
22 & Debugging Techniques: Select texts that discuss debugging techniques with code examples. \\
\hline
23 & Tool Comparison: Include texts that compare different programming tools or environments. \\
\hline
24 & Technical Focus: Prioritize articles or excerpts from technical books that focus on programming. \\
\hline
25 & Academic Pseudo-Code: Include texts from academic papers that contain pseudo-code or algorithms. \\
\hline
26 & Content Density: Exclude texts that are excessively verbose without substantive content. \\
\hline
27 & Optimization Tips: Prioritize texts that provide insights into code optimization. \\
\hline
28 & Advanced Topics: Include texts that cover advanced programming topics like concurrency or security. \\
\hline
29 & Architecture Patterns: Select texts that discuss architectural patterns with code examples. \\
\hline
30 & Programming Paradigms: Prioritize examples that demonstrate functional or logic-based programming. \\
\hline
31 & Technical Emphasis: Exclude samples that focus solely on non-technical aspects of IT projects. \\
\hline
32 & Quality Solutions: Include forum and Q\&A entries with high-quality code solutions. \\
\hline
33 & Documentation Inclusion: Select project documentation and readme files that include example usage of code. \\
\hline
34 & Commented Code: Prioritize texts with code that includes comprehensive inline comments. \\
\hline
35 & Jargon Balance: Exclude texts with a high density of technical jargon unless accompanied by clear explanations or code. \\
\hline
36 & Executable Snippets: Include code snippets that are functional and can be executed without modifications. \\
\hline
37 & Complexity Discussion: Prioritize texts that explain the computational complexity of algorithms with examples. \\
\hline
38 & Integration Showcase: Include texts that showcase the integration of different technologies or languages. \\
\hline
39 & Version Control: Select samples that explain version control practices with code snippets. \\
\hline
40 & Cross-Platform Coding: Prioritize texts that discuss cross-platform coding challenges and solutions. \\
\hline
41 & Interactive Tutorials: Include interactive coding tutorials or walkthroughs. \\
\hline
42 & Proprietary Code: Exclude any samples containing proprietary code without proper authorization. \\
\hline
43 & Scalability Focus: Select examples that discuss the scalability of code or systems. \\
\hline
44 & Accessibility Coding: Prioritize samples that address coding for accessibility or internationalization. \\
\hline
45 & Performance Analysis: Include texts that analyze the performance of different coding approaches. \\
\hline
46 & Content Relevance: Exclude texts that mix code with irrelevant images or multimedia that don’t add educational value. \\
\hline
47 & Ethical Coding: Prioritize texts that discuss ethical considerations in programming. \\
\hline
48 & Tool Usage: Include examples of how to use popular development tools and environments through coding tutorials. \\
\hline
49 & Project Scope: Select texts that clearly define the scope and objectives of programming projects. \\
\bottomrule
\end{tabularx}
\end{table*}

\section*{Appendix H — Extended Discussion}

We include here additional commentary on practical considerations, modeling decisions, and possible extensions of our method, with the aim of providing readers with a broader perspective on its applicability and trade-offs.

\paragraph{On the minimal manual filtering step.} 
While our pipeline is largely automated, we retain a light manual filtering step during rule generation to remove duplicate or malformed rules. This intervention is minimal: typically under 15 minutes per domain, and serves only to ensure syntactic clarity and diversity of candidate rules. In future applications, simple heuristics could easily replace this step. Importantly, once rule scoring and orthogonality-aware sampling begin, the selection process is entirely model-driven.

\paragraph{On compute requirements.} 
The cost of rule generation and rating is modest compared to full-scale fine-tuning. Rule generation uses a single GPT-4 call per domain, and the pairwise rule correlation matrix is computed efficiently over a single batch of $n = 10{,}000$ examples. The DPP selection step is fast due to the small size of the kernel ($R \leq 50$), and the entire pipeline runs comfortably on a single GPU. Overall, our method introduces only a minor overhead relative to the total training cost, while enabling more targeted, higher-quality supervision.

\paragraph{On task coverage.} 
We chose four representative domains: IMDB, Medical, Math, and Code to demonstrate the method's generality across classification, reasoning, and program synthesis. These already cover a wide range of input modalities and supervision structures. Extending to more open-ended tasks (e.g. dialog safety or creative writing) is an exciting direction but would require new evaluation methodologies, which we view as orthogonal to the core contributions of this paper.

\paragraph{On rating model overlap.} 
Although we use the same backbone family (e.g., Llama-3) for scoring and fine-tuning, we mitigate circularity in two ways: (i) using both 8B and 70B models for rule-based scoring to assess robustness, and (ii) validating that downstream improvements hold across architectures and model sizes. We also include experiments using Claude-Sonnet as a rule generator to verify that our pipeline is not overly sensitive to the initial LLM used. Evaluating with independent human or hybrid judges is left for future work.

\paragraph{On parameter choices.} 
We fix $r = 10$ rules per task and use $n = 10K$ examples for scoring based on empirical pilot studies. These values work reliably across all domains. Our ablation study shows the method is stable across a broad range of $r$, and performance does not degrade sharply unless $r$ is extremely low or high. While further tuning may yield marginal improvements, our results suggest that high-quality rule-based selection does not require careful hyperparameter search.

\end{document}